%% file: moeve-framework-paper.tex
\documentclass[pdflatex]{sn-jnl}

\usepackage[utf8]{inputenc}
\usepackage[T1]{fontenc}

\usepackage[square,numbers]{natbib}
\usepackage{graphicx}%
\usepackage{multirow}%
\usepackage{amsmath,amssymb,amsfonts}%
\usepackage{amsthm}%
\usepackage{mathrsfs}%
\usepackage[title]{appendix}%
\usepackage{xcolor}%
\usepackage{textcomp}%
\usepackage{manyfoot}%
\usepackage{booktabs}%
\usepackage{algorithm}%
\usepackage{algorithmicx}%
\usepackage{algpseudocode}%
\usepackage{listings}%
\usepackage{tabularx}%
\usepackage{makecell}
\usepackage{rotating}
\usepackage{csquotes}  
\usepackage{subcaption}
\usepackage{tikz}
\usetikzlibrary{positioning, fit, backgrounds, calc, arrows.meta, shapes.geometric}
\usepackage{fontawesome5}
\usepackage{etoolbox}
\makeatletter
\patchcmd{\@maketitle}
  {Contributing authors:\ \authemail}
  {E-mail(s): \authemail}{}{}
\makeatother
\definecolor{moeve-blue}{HTML}{2B5C8A}
\definecolor{moeve-perf}{HTML}{4A90C4}
\definecolor{moeve-gov}{HTML}{C4784A}
\definecolor{moeve-light-perf}{HTML}{E8F0F8}
\definecolor{moeve-light-gov}{HTML}{F8EDE8}
\definecolor{moeve-gray}{HTML}{F5F5F5}
\definecolor{moeve-darkgray}{HTML}{555555}

\theoremstyle{thmstyleone}%
\theoremstyle{thmstyletwo}%

\theoremstyle{thmstylethree}%

\begin{document}

\title[MÖVE]{\raisebox{-0.3\height}{\includegraphics[height=1.25\baselineskip]{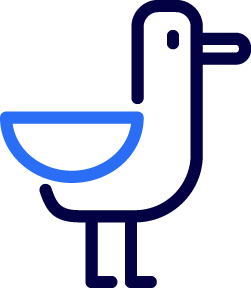}}MÖVE: A Holistic LLM Benchmark for the German Public Sector}


\author{\fnm{Camilla} \sur{Dalerci}}\email{camilla.dalerci@bdr.de}
\equalcont{These authors contributed equally to this work.}

\author{\fnm{Thilo} \sur{Michael}}\email{thilo.michael@bdr.de}
\equalcont{These authors contributed equally to this work.}

\author{\fnm{Robin} \sur{Schaefer}}\email{robin.schaefer@bdr.de}
\equalcont{These authors contributed equally to this work.}

\author{\fnm{Daniel} \sur{Weinland}}\email{daniel.weinland@bdr.de}
\equalcont{These authors contributed equally to this work.}

\affil{\orgname{Innovations Department, Bundesdruckerei GmbH}, \city{Berlin}, \country{Germany}}


\input{sections/00_abstract}

\keywords{Large Language Models, Benchmark, German Public Sector, NLP, AI Governance, Sustainability, LLM Evaluation}



\maketitle

\tableofcontents
\newpage

\input{sections/01_introduction}
\input{sections/02_related_work}
\input{sections/03_framework_overview}
\input{sections/04_methodology}
\input{sections/05_model_evaluation}
\input{sections/06_benchmark_evaluation}
\input{sections/08_future_work_conclusion}


\bibliographystyle{unsrtnat}
\bibliography{bibliography}

\input{sections/09_appendix}

\end{document}

%% file: sections/00_abstract.tex
\abstract{We present MÖVE (\textit{\textbf{M}odelle für die \textbf{\"O}ffentliche \textbf{V}erwaltung \textbf{E}valuieren}), a holistic benchmark for evaluating large language models (LLMs) in the context of the German public sector.
While LLMs are increasingly adopted in public administration, model selection remains largely ad hoc, and existing benchmarks offer limited guidance: they are predominantly English-centric, US-centric in content, and focus exclusively on task performance.
MÖVE addresses these gaps by evaluating 39 models across two complementary dimensions.
\textit{Performance criteria} cover summarization, question answering, and topic extraction.
\textit{Governance criteria} assess hallucination tendencies, energy consumption, provider transparency, and alignment with German constitutional values and knowledge about positions by German political parties.
In total, we utilize ten German-language datasets, including gold- and silverstandard datasets that we constructed to reflect public-administration domains.
We employ a multi-metric evaluation strategy combining classical NLP metrics, embedding-based methods, and LLM-as-a-judge approaches.
Our results show that no single model dominates across all criteria: top performers differ between tasks, and model size alone is a poor predictor of quality.
We further evaluate the benchmark itself, analyzing its statistical precision, LLM judge reliability, the impact of our private datasets on model rankings, the sensitivity of our results to prompt formulation, and the validity of our energy consumption estimates.
MÖVE is designed as a living benchmark under active development; results are publicly available at \url{https://moeve.bundesdruckerei.de/}.}

%% file: sections/01_introduction.tex
\section{Introduction}
\label{ch:introduction}

Large language models (LLMs) are increasingly being adopted in public administration, where they are applied to tasks such as summarizing policy documents, generating responses to questions about legal texts, extracting key topics from lengthy reports, and automating service workflows~\cite{fraunhofer_llm_publicadmin_2024}.
In Germany, this adoption coincides with mounting demographic and organizational pressure on the public sector: of the roughly 5.4 million people employed in public service in 2024~\cite{destatis_publicservice_2025}, more than 1 million are expected to retire by 2030~\cite{pwc_publicsector_retention}.
Given this projected workforce gap, LLMs are widely discussed as a means to help sustain service delivery and administrative capacity.
As governments explore these opportunities, the question of which model to deploy becomes consequential: model choice affects output quality, operational cost, energy consumption, regulatory compliance, and ultimately the trust that citizens and civil servants place in AI-based processes.

Yet in practice, model selection in the German public sector is rarely informed by systematic evaluation.
Based on our extensive experience working with public-sector institutions, we observe that decisions are often driven by unbalanced media coverage surrounding the release of new LLMs or by country-of-origin preferences, for instance, favoring domestic providers on the assumption that a German model is inherently better suited for German-language tasks.
While such heuristics are understandable in the absence of better information, they do not constitute a reliable basis for procurement and deployment decisions that affect administrative processes at scale.

Existing LLM benchmarks offer limited guidance for this context.
The most prominent general-purpose benchmarks, including MMLU~\cite{hendrycks_measuring_2021}, BIG-Bench~\cite{srivastava_beyond_2023}, and GLUE/SuperGLUE~\cite{wang-etal-2018-glue,wang-etal-2019-superglue}, are English-only and focus on knowledge retrieval or linguistic competence rather than domain-specific tasks.
Multilingual efforts such as Global MMLU~\cite{singh-etal-2025-global} and the European benchmark by \citet{thellmann-etal-2024-towards} include German, but rely predominantly on translated data.
However, translation addresses only the language gap, not the domain gap: a benchmark based on US civics or legal precedents does not automatically become relevant for German civil servants just because it has been translated into German.
Most existing benchmarks fail to address either gap for the German public sector.
In addition, the vast majority of benchmarks focus exclusively on task performance.
While HELM~\cite{liang_holistic_2023} represents a ``holistic'' evaluation, its notion of holism refers to breadth of NLP tasks rather than to evaluating models across fundamentally different dimensions; governance criteria such as transparency, sustainability, or value alignment remain underrepresented in the framework.
The widespread use of publicly available test data further raises concerns about data contamination~\cite{sainz-etal-2023-nlp,dong-etal-2024-generalization}, as models may have been inadvertently trained on the very datasets used to evaluate them.

Recently, a small but growing number of benchmarks target government domains specifically.
PubHealthBench~\cite{harris-etal-2025-healthy} evaluates LLMs on UK public health guidance, CitizenQuery-UK~\cite{majithia-etal-2026-citizenquery} focuses on citizen-facing QA in the UK, and MSGABench~\cite{liu-etal-2025-msgabench} addresses Chinese government affairs.
However, these benchmarks are again tied to their respective national contexts and languages, evaluate only task performance, and do not include governance criteria.
A recent meta-analysis by \citet{rystrom-etal-2026-agentbenchmarks} corroborates this assessment, finding that no existing benchmark meets public-sector requirements.

For public-sector institutions, governance is not an optional add-on but a core requirement.
Article 53 of the EU AI Act~\cite{euaiact2024} introduces concrete obligations regarding transparency and documentation for providers of general-purpose AI models, increasing the relevance of documented regulatory compliance throughout a model's life cycle.
Similarly, sustainability considerations are becoming increasingly relevant for public procurement, where energy consumption and environmental impact may factor into purchasing decisions.
Hallucination, or the generation of plausible yet unsupported content, poses particular risks when models are used to process official documents, legal texts or policy briefs, where factual reliability is paramount.
Finally, while notable European efforts exist, the LLM landscape remains dominated by models developed in the United States and China.
For public administrations bound by the principles of the German constitution (\textit{Grundgesetz}), this raises the question of whether widely available models align with German and European values.

\begin{figure}[t]
    \centering
    \resizebox{0.85\linewidth}{!}{%
    \begin{tikzpicture}[
        every node/.style={font=\small},
        criterion/.style={
            rounded corners=3pt,
            minimum width=3.8cm,
            minimum height=0.7cm,
            align=center,
            font=\small,
        },
        perf-criterion/.style={criterion, fill=moeve-light-perf, draw=moeve-perf!60, text=moeve-blue},
        gov-criterion/.style={criterion, fill=moeve-light-gov, draw=moeve-gov!60, text=moeve-gov!80!black},
        pillar-header/.style={
            rounded corners=4pt,
            minimum width=4.2cm,
            minimum height=0.85cm,
            align=center,
            font=\small\bfseries,
            text=white,
        },
        detail/.style={font=\scriptsize, text=moeve-darkgray, align=center},
    ]
    \node[font=\large\bfseries, text=moeve-blue] (title) {MÖVE};
    \node[font=\small, text=moeve-darkgray, below=0.05cm of title] (subtitle) {Modelle f\"ur die \"Offentliche Verwaltung Evaluieren};
    \node[pillar-header, fill=moeve-perf, below left=0.5cm and -2.4cm of subtitle] (perf-header) {Performance};
    \node[pillar-header, fill=moeve-gov, below right=0.5cm and -2.4cm of subtitle] (gov-header) {Governance};
    \node[perf-criterion, below=0.4cm of perf-header] (sum) {Summarization};
    \node[perf-criterion, below=0.15cm of sum] (qa) {Question Answering};
    \node[perf-criterion, below=0.15cm of qa] (te) {Topic Extraction};
    \node[gov-criterion, below=0.4cm of gov-header] (hall) {Hallucination};
    \node[gov-criterion, below=0.15cm of hall] (sus) {Sustainability};
    \node[gov-criterion, below=0.15cm of sus] (trans) {Transparency};
    \node[gov-criterion, below=0.15cm of trans] (pol) {Politics \& Values};
    \begin{scope}[on background layer]
        \node[fit=(perf-header)(sum)(qa)(te), rounded corners=6pt, fill=moeve-perf!5, draw=moeve-perf!25, inner xsep=10pt, inner ysep=10pt] (perf-box) {};
        \node[fit=(gov-header)(hall)(sus)(trans)(pol), rounded corners=6pt, fill=moeve-gov!5, draw=moeve-gov!25, inner xsep=10pt, inner ysep=10pt] (gov-box) {};
    \end{scope}
    \node[rounded corners=4pt, fill=moeve-gray, draw=moeve-blue!30, minimum width=10.5cm, minimum height=0.9cm, align=center, font=\small, text=moeve-darkgray, below=0.7cm of $(perf-box.south)!0.5!(gov-box.south)$] (bottom) {39 models\quad$\cdot$\quad 10 datasets\quad$\cdot$\quad German language\quad$\cdot$\quad Private gold/silver-standard datasets\quad$\cdot$\quad Living benchmark};
    \end{tikzpicture}%
    }
    \caption{Overview of the MÖVE framework. Models are evaluated along two complementary dimensions: performance criteria assess output quality on administrative tasks, while governance criteria address the broader conditions for responsible deployment.}
    \label{fig:framework_overview}
\end{figure}
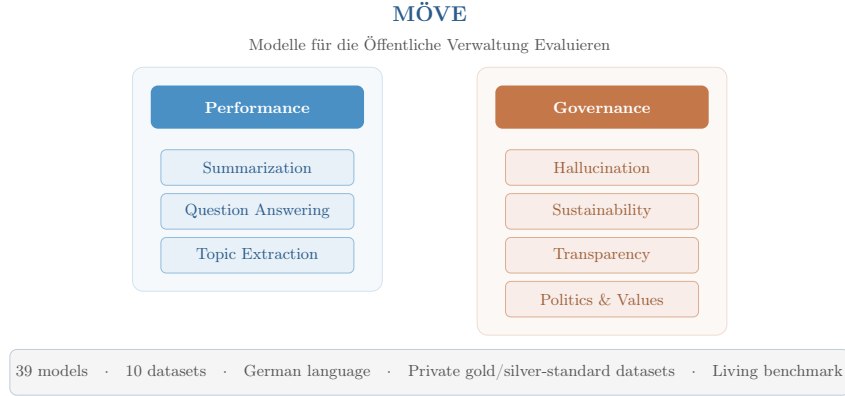

In this paper, we present \textbf{MÖVE} (\textit{Modelle für die öffentliche Verwaltung evaluieren}), a holistic LLM benchmark designed for the German public sector.
MÖVE evaluates models along two complementary dimensions.
\textit{Performance criteria} assess output quality on tasks directly relevant to administrative work: summarization, question answering, and topic extraction.
\textit{Governance criteria} address the broader conditions for responsible deployment: hallucination tendencies, sustainability, transparency, and alignment with political values.
We evaluate 39 models on ten datasets, including gold- and silverstandard datasets that we constructed specifically to reflect the domain of German public administration.
All tasks and prompts are in German, and we employ a multi-metric evaluation strategy combining classical NLP metrics, embedding-based methods, and LLM-as-a-judge approaches.
Beyond reporting model rankings, we conduct a methodological self-evaluation of the benchmark itself, analyzing its statistical precision, the reliability of LLM-based judge metrics, the impact of our internal datasets on model rankings, the sensitivity of results to prompt formulation, and the validity of our energy consumption estimates.
MÖVE is designed as a living benchmark under active development.
This paper presents its first comprehensive evaluation; results are continuously updated and publicly available.\footnote{\url{https://moeve.bundesdruckerei.de/}}
Figure~\ref{fig:framework_overview} provides a high-level overview of the framework.

The remainder of this paper is structured as follows.
Section~\ref{ch:related_work} reviews related work on LLM benchmarking for both performance and governance criteria.
Section~\ref{ch:framework_overview} introduces the MÖVE framework, defining the evaluation criteria and the general evaluation setup.
Section~\ref{ch:methodology} describes the methodology in detail, including datasets, models, metrics, and the evaluation design for each criterion.
Section~\ref{ch:model_evaluation} presents the results of the model evaluation across all performance and governance criteria.
Section~\ref{ch:benchmark_evaluation} evaluates the benchmark itself through analyses of statistical precision, judge reliability, dataset impact, prompt sensitivity, and estimate validity of energy consumption.
Section~\ref{ch:future_work_conclusion} concludes and discusses directions for future work.

%% file: sections/02_related_work.tex
\section{Related Work}
\label{ch:related_work}

In this section, we present the relevant related literature on LLM benchmarking in general, and with respect to both performance and governance tasks.
As our framework focuses on the evaluation of LLMs, we refrain from including research on the evaluation of agentic or end-to-end AI systems.

\subsection{LLM Benchmarking}

While benchmarking as such is a traditional concept, modern approaches to the benchmarking of language models are strongly associated with the emergence of \textit{transformer} models~\cite{vaswani-etal-2017-attention,DBLP:journals/corr/abs-1810-04805,brown-etal-2020-language}.
Benchmarks can be classified according to the different dimensions of \textit{competence}\footnote{We acknowledge that LLMs should be evaluated in terms of functionality instead of \textit{competence}, given their design as next-token predictors. However, as the relevant research tends to frame benchmarking as \textit{assessing competence}, we use the term in the \textit{related work} section.} they try to assess, including linguistic competence, domain-specific competence, and language-specific competence.
\paragraph{Linguistic Competence}
Different evaluation suites were developed to assess linguistic competence, usually with a rather broad conception of the linguistic phenomenon at hand.
GLUE~\cite{wang-etal-2018-glue} and SuperGLUE~\cite{wang-etal-2019-superglue} are early examples, covering diverse linguistic tasks (textual entailment, sentiment analysis, coreference resolution) as proxies for \textit{language understanding}.
Similarly, Swag \cite{zellers-etal-2018-swag} and HellaSwag \cite{zellers-etal-2019-hellaswag} form iterations of datasets used for evaluating model functionalities in \textit{commonsense natural language inference}.
Another influential benchmark is WinoGrande \cite{sakaguchi-etal-2021-winogrande}, designed to evaluate commonsense reasoning by reducing the confounding effect on models of statistical biases in the data.
While these benchmarks were designed for English, it is important to note that some work on linguistic benchmarking for other languages exists, e.g., CLUE \cite{xu-etal-2020-clue}, which has been developed for LLM evaluation in Chinese.

\paragraph{Domain Competence}
Researchers have focused on the development of evaluation suites for domain competence, including domain knowledge, both from a broad and narrow perspective.
One prominent example is MMLU~\cite{hendrycks_measuring_2021}.
This test includes 57 tasks across various domains, e.g. philosophy, US history, law and mathematics, and levels of expertise, ranging from elementary school to professional knowledge.
More recently, MMLU-Pro~\cite{wang-etal-2024-mmlupro} was introduced as a more challenging extension to MMLU, which focuses more intensely on \textit{reasoning} instead of the former \textit{knowledge}-driven task design.
Concentrating on the natural sciences, GPQA \cite{rein2024gpqa} tests expert-level knowledge in the respective field through questions created by domain experts.
Conceptualized as a multi-turn dialogue setting, HealthBench \cite{arora2025healthbenchevaluatinglargelanguage} contains \textit{conversations} between a model and a user or healthcare professional.
This benchmark differs from other evaluation suits by employing a sophisticated rubric developed by physicians.
HiST-LLM \cite{hauser-etal-2024-large} represents a humanities benchmark, which focuses on global history knowledge.
With history being an inherently biased field, the researchers include expert-level references covering over 600 historical societies to provide more geographically balanced coverage.
Despite being of interest for the general public, benchmarking LLMs for the public sector domain remains understudied, with some notable exceptions.
PubHealthBench \cite{harris-etal-2025-healthy}, for instance, assesses LLM output accuracy about UK public health guidance in QA and open response settings.
Data has been collected from official UK government websites.
CitizenQuery-UK \cite{majithia-etal-2026-citizenquery} contains 22,000 citizen queries on government-related aspects, like policies and services, and synthetically generated responses, usable for LLM evaluation.
MSGABench \cite{liu-etal-2025-msgabench} is a Chinese-language dataset employed for the benchmarking of LLMs, revealing, for instance, substantial performance differences and concerns with respect to privacy vulnerabilities and observed biases.
DGDB \cite{deswart-etal-2025-detecting}  evaluates how well LLMs detect social bias in Dutch government documents.

\paragraph{Language-Specific Competence}
While many of the previously mentioned benchmarks cover a relatively diverse set of tasks, they primarily focus on the English language.
This substantially hinders the interpretability of said benchmarks and their applicability to other linguistic contexts.
In recent years, researchers have begun to develop alternatives, either by designing multilingual benchmarks \cite{adelani-etal-2025-irokobench,singh-etal-2025-global,susanto-etal-2025-sea} or by concentrating on individual languages \cite{yin-etal-2024-respect,yuksel-etal-2024-turkishmmlu,son-etal-2025-kmmlu}.
These benchmarks often draw inspiration from previously existing datasets like MMLU, but to account for potential translation errors and biases in the source data, many opt to create original content. 
For instance, the Korean KMMLU \cite{son-etal-2025-kmmlu} and TurkishMMLU \cite{yuksel-etal-2024-turkishmmlu} are fully based on own data, drawn from license tests and high school curricula, respectively. The Japanese JMMLU \cite{yin-etal-2024-respect} takes a hybrid approach, complementing machine-translated MMLU data with original Japanese data.
This accounts for potential errors during translation and biases existing in the source data.
A recent example of a multilingual benchmark is Global MMLU \cite{singh-etal-2025-global}, which explores cultural and linguistic biases prevalent in LLM evaluation.
The benchmark consists of data in 42 languages, obtained through a mixture of machine translation and professional human translation.
Other work has been conducted on the development of LLM benchmarks for languages that are geographically clustered.
IrokoBench \cite{adelani-etal-2025-irokobench}, for instance, is a benchmark for 17 African languages covering three tasks, e.g., natural language reasoning and QA.
Another example is SEA-HELM \cite{susanto-etal-2025-sea}, a benchmarking suite for five south-east Asian languages.
Tasks fall into five diverse areas, broadly covering language-, culture- and safety-related evaluations.
Both benchmarks have in common that they cover a set of typologically diverse languages, which, thus, represent a challenge for a given language model.
Finally, \citet{thellmann-etal-2024-towards} suggest a benchmark for 21 European languages, including German.
The authors evaluate 40 LLMs on five prominent datasets, e.g. MMLU and HellaSwag, which have been machine-translated into the respective target languages.

\paragraph{Holistic and General Intelligence Benchmarks}
Some benchmarks follow an holistic approach, i.e., try to evaluate LLMs across various dimensions.
This includes linguistic task performance, factual output accuracy with respect to general knowledge and different performance or governance tasks.
For instance, HELM \cite{liang_holistic_2023} covers 42 scenarios, half of which were not part of previous benchmarks, evaluated with up to seven metrics.
While performance evaluations form the majority, the benchmark also includes some governance criteria, e.g. fairness.
Another vast benchmark is BIG-Bench \cite{srivastava_beyond_2023}, which covers 204 tasks drawn from a diverse set of areas, including linguistics, common-sense reasoning, (natural) sciences and coding.
Human expert ratings are used as baselines.
Finally, some benchmarks try to assess a model's so-called \textit{general intelligence}.
For instance, ARC \cite{chollet-2019-measure} and ARC-AGI-2 \cite{chollet-etal-2025-arcagi2} use puzzle solving to assess general reasoning and logical thinking.
This contrasts with the usual focus on skill performance applied in the majority of benchmarks.

\paragraph{Challenges and Criticisms in LLM Benchmarking}
Despite the substantial reliance on benchmarking suites for LLM evaluation, a growing body of criticism questions whether these benchmarks meaningfully capture what they are meant to, as well as whether the current method is a sustainable and fair approach to measure progress \cite{gevers-etal-2025-benchmarks}.
A central concept in this debate is \textit{construct validity}: the degree to which a benchmark actually measures the functionality or dimension it claims to assess~\cite{raji-etal-2021-ai}.
\citet{raji-etal-2021-ai} criticize the tendency to use a set of somewhat arbitrarily defined tasks as evidence for improvement in general dimensions such as \textit{language understanding}, resulting in low construct validity.
Having also observed an overall low construct validity in a sample of 445 LLM benchmarks, \citet{bean-etal-2025-measuring} recently suggested a list of recommendations to incorporate during benchmark development.
The concept is directly relevant to the present work: a benchmark intended to inform model selection for the German public sector must ensure that its tasks, datasets, and metrics reflect the actual requirements of that domain. This concern motivates both the design choices described in Sections~\ref{ch:framework_overview}--\ref{ch:methodology} and the methodological self-evaluation presented in Section~\ref{ch:benchmark_evaluation}.
Other issues identified are \textit{performance saturation} \cite{wang-etal-2019-superglue,bowman-dahl-2021-will,wang-etal-2024-mmlupro} and \textit{data contamination} \cite{sainz-etal-2023-nlp,dong-etal-2024-generalization,li-etal-2024-open-source,oren--etal-2024-proving}.
To account for the latter, PeerBench \cite{cheng-etal-2025-benchmarking} has been suggested as a community-governed approach to benchmarking, which includes 1) the usage of secret continuously updated test data, 2) the execution of evaluation runs in an identical environment for each model, 3) setting up a community-based data curation and review workflow, and 4) reporting multiple metrics per task.
Focusing on the full benchmark lifecycle, BetterBench \cite{reuel-etal-2024-betterbench} was developed as an assessment framework of 46 best practices that can be used as an evaluation suit for benchmarks.
Evaluating 24 benchmarks again provided evidence for substantial issues in contemporary benchmarks.

\subsection{Performance Criteria}

\paragraph{Summarization}
Automatic text summarization is a long-standing NLP task, traditionally divided into \textit{extractive} approaches, which select salient sentences from the source~\cite{kupiec1995trainable,gong_generic_2001,cheng_neural_2016}, and \textit{abstractive} approaches, which generate novel text that paraphrases and condenses the input~\cite{knight2002summarization,rush_neural_2015,see_get_2017}.
With the advent of large pre-trained language models, abstractive summarization has seen substantial progress, making it the predominant paradigm in recent work; for a comprehensive overview, we refer the reader to the surveys by \citet{koh_empiricalsurveylong_2023} and \citet{zhang2025comprehensive}.

Much of the benchmarking literature on summarization centers on a small number of English-language datasets.
CNN/DailyMail~\cite{nallapati_abstractivetextsummarization_2016} and XSum~\cite{narayan_dontgiveme_2018} are among the most widely used, targeting news article summarization at different levels of abstraction.
For longer documents, GovReport~\cite{huang_efficientattentionslong_2021} provides summaries of US government reports, and BillSum~\cite{kornilova_billsumcorpusautomatic_2019} focuses on US congressional bills, both demonstrating the relevance of summarization in public-sector contexts.
However, these resources are overwhelmingly English-centric.
Notable exceptions in multilingual legal summarization include Eur-Lex-Sum~\cite{aumiller2022eur} and the Swiss Leading Decision Summarization dataset~\cite{rasiah2023scale}, which provide German-language document–summary pairs in the legal domain and which we adopt in the \textit{MÖVE framework}.
Beyond these, resources for German-language summarization, particularly in the context of public administration, remain scarce.

The evaluation of summarization quality has evolved alongside the models.
Traditional metrics such as \textit{ROUGE}~\cite{lin_rouge_2004} and \textit{BLEU}~\cite{papineni_bleumethodautomatic_2001} measure lexical overlap but are known to correlate poorly with human judgments of summary quality, particularly for abstractive outputs~\cite{fabbri_summevalreevaluatingsummarization_2021a}.
This has motivated the adoption of model-based metrics, including \textit{BERTScore}~\cite{zhang_bertscore_2020} and \textit{SemScore}~\cite{aynetdinov_semscore_2024}, which compare candidate and reference texts via contextualized embeddings, and more recently LLM-as-a-judge approaches~\cite{zheng_judgingllmasajudgemtbench_2023,liu_gevalnlgevaluation_2023,li2024llms}, which leverage LLMs to assess summary quality along multiple dimensions.
However, LLM-based judges are not without limitations, as studies have identified biases related to position, verbosity, and self-enhancement~\cite{zheng_judgingllmasajudgemtbench_2023}, motivating the use of complementary metric types.

Notably, the summarization datasets described above were primarily designed for training and evaluating pre-LLM summarization models rather than for benchmarking LLMs.
Only recently have studies begun to specifically assess LLM summarization functionalities, notably \citet{goyal_newssummarizationevaluation_2022} and \citet{zhang_benchmarkinglargelanguage_2024a}, both of which focus on English news summarization.
Systematic LLM benchmarks that target non-English languages or domain-specific settings such as public administration remain scarce.

\paragraph{Question Answering}
Question answering (QA) is a core NLP task with a long research history; for comprehensive overviews of QA tasks and datasets, we refer the reader to \citet{rogers_qadatasetexplosion_2023}; for a survey of retrieval-augmented generation methods applicable to QA, see \citet{zhu_retrievalaugmentedgeneration_2024}.

QA systems can be distinguished along several dimensions.
With respect to the knowledge source, a key distinction is between systems that are based on a model's \textit{parametric knowledge} and those that generate responses to questions related to a \textit{provided context}~\cite{hermann_teachingmachinesread_2015,rajpurkar-etal-2016-squad}.
In the latter case, the context may be retrieved from a large corpus (\textit{open-domain} QA)~\cite{chen_readingwikipediaanswer_2017,lee_latentretrievalweakly_2019} or provided directly (\textit{reading comprehension})~\cite{hermann_teachingmachinesread_2015,rajpurkar-etal-2016-squad}.
Orthogonally, and analogous to the distinction in summarization, answers can be \textit{extractive}, i.e. directly taken from the source text, or \textit{abstractive}, i.e. generated as continuous text.
Many of the currently used QA benchmarks actually focus on probing a model's \textit{parametric knowledge} through multiple-choice formats~\cite{hendrycks_measuring_2021,lin_truthfulqa_2022}.
However, such settings are far removed from real-world applications in which users need to obtain answers from specific documents.
For public-sector applications, context-grounded abstractive QA is therefore the most relevant setting.

Much of the benchmarking literature on QA is centered on English-language datasets.
SQuAD~\cite{rajpurkar-etal-2016-squad} and its successor SQuAD~2.0~\cite{rajpurkar_knowwhatyou_2018} are among the most influential reading comprehension benchmarks, requiring models to extract answer spans from Wikipedia paragraphs.
Natural Questions~\cite{kwiatkowski_naturalquestionsbenchmark_2019} extends this paradigm by using real user queries from a search engine, while TriviaQA~\cite{joshi_triviaqalargescale_2017} tests factual knowledge retrieval over longer documents.
For abstractive and long-form QA, ELI5~\cite{fan_eli5longform_2019} requires paragraph-length explanatory answers.

Regarding non-English QA resources, several multilingual benchmarks have been developed.
XQuAD~\cite{artetxe_crosslingualTransferability_2020} and MLQA~\cite{lewis_mlqaevaluatingcrosslingual_2020} provide cross-lingual reading comprehension data, including German.
For German specifically, German-QuAD~\cite{moller-etal-2021-germanquad} adapts the SQuAD paradigm to German Wikipedia articles and represents one of the few dedicated German-language QA resources, which we adopt in the \textit{MÖVE framework}.
Beyond these, German-language QA resources for domain-specific settings such as public administration are largely absent.

Evaluation of QA systems has traditionally relied on lexical-overlap metrics such as \textit{Exact Match} and token-level \textit{F1}~\cite{rajpurkar-etal-2016-squad}.
However, these metrics are ill-suited for abstractive QA, where semantically equivalent answers may differ substantially in surface form.
This has motivated the adoption of embedding-based metrics such as \textit{BERTScore}~\cite{zhang_bertscore_2020} and \textit{SemScore}~\cite{aynetdinov_semscore_2024}, as well as LLM-driven evaluation frameworks such as RAGAS~\cite{es_ragasautomatedevaluation_2025}, which decomposes QA quality into dimensions including \textit{faithfulness} and \textit{noise sensitivity}.
Systematic QA benchmarks that combine non-English data with domain-specific contexts and LLM-oriented evaluation remain scarce.

\paragraph{Topic Extraction}
Topic extraction, i.e., the task of extracting a concise set of keywords or phrases that capture the central themes of a document, is closely related to, yet distinct from, document summarization.
While summarization produces coherent prose, topic extraction yields compact, structured labels and is often used as a first orientation aid in information management workflows.
The task is commonly studied under the label of \textit{keyphrase extraction}; for a comprehensive overview, we refer the reader to the surveys by \citet{papagiannopoulou_reviewkeyphraseextraction_2020} and \citet{song_surveyrecentadvances_2023}.

As in summarization and question answering, a fundamental distinction exists between \textit{extractive} methods, which select phrases that appear verbatim in the document, and \textit{generative} approaches, which produce keyphrases that may not occur in the source text~\cite{meng_deepkeyphrasegeneration_2017}.
A variety of unsupervised extractive approaches have been proposed, including graph-based methods such as TextRank~\cite{mihalcea_textrankbringingorder_2004} and statistical methods such as YAKE~\cite{campos_yake_2020}.
More recently, embedding-based approaches like KeyBERT~\cite{grootendorst_maartengrkeybertv0_2021} leverage pre-trained language model representations to rank candidate phrases by semantic similarity to the document.
Early studies on LLMs for keyphrase generation report mixed results: some find that prompted models can outperform traditional methods~\cite{martinezcruz_chatgptstateoftheart_2023}, while others highlight remaining challenges, particularly for generating keyphrases not present in the source text~\cite{song_ischatgptgood_2023}.
Traditional methods remain competitive and offer clear advantages in terms of cost and efficiency; however, evaluating LLM-based topic extraction is relevant for applications that are already centered around LLMs, where keyphrase generation can be integrated as part of a broader workflow.
Topic extraction should also be distinguished from \textit{topic modeling}, where methods such as LDA~\cite{blei_latentdirichletallocation_2003} or BERTopic~\cite{grootendorst_bertopicneuraltopic_2022} detect latent thematic structures across a corpus.
In contrast, the task we address operates at the document level, generating keywords for individual texts, i.e., a setting more directly aligned with the needs of public-sector employees processing specific documents.

Standard benchmarks for keyphrase extraction are predominantly English and drawn from the scientific domain, including Inspec~\cite{hulth_improvedautomatickeyword_2003}, SemEval-2010 Task~5~\cite{kim_automatickeyphraseextraction_2013}, and KP20k~\cite{meng_deepkeyphrasegeneration_2017}; KPTimes~\cite{gallina_kptimeslargescaledataset_2019} extends this to the news domain.
Multilingual resources have only recently begun to emerge; notably, EUROPA~\cite{salaun_europalegalmultilingual_2024} provides keyphrase annotations for EU legal documents across 24 official languages including German.
Beyond this, dedicated German-language resources for keyphrase or topic extraction, particularly in public-administration contexts, are largely absent.

Evaluation of keyphrase extraction has traditionally relied on exact-match F1 between predicted and reference keyphrases.
However, this approach is overly sensitive to lexical variation and penalizes valid paraphrases or synonyms.
Recent work has proposed semantic evaluation alternatives based on embedding similarity, such as BERTScore~\cite{zhang_bertscore_2020} and keyphrase-specific metrics~\cite{wu_kpevalfinegrainedsemanticbased_2024}, which better capture the semantic equivalence of predicted and reference topics.
Systematic benchmarks that combine document-level topic extraction with non-English data, domain-specific contexts, and LLM-oriented evaluation remain scarce.

\subsection{Governance Criteria}

\paragraph{Hallucination}
LLMs are well known for occasionally producing incorrect or unsupported output, a phenomenon often referred to as \textit{hallucination}~\cite{maynez_faithfulnessfactualityabstractive_2020,kryscinski_evaluatingfactualconsistency_2019}.
Given its potential downstream risks, systematic evaluation of hallucination has become a central component of LLM assessment.
Several benchmarks have been proposed to evaluate hallucination tendencies, including TruthfulQA~\cite{lin_truthfulqa_2022} and FActScore~\cite{min_factscorefinegrainedatomic_2023}; for a comprehensive overview, we refer the reader to the surveys by \citet{ji_surveyhallucinationnatural_2023,huang_surveyhallucinationlarge_2025}.
There is not a single type of hallucination, and various taxonomies have been introduced to classify the phenomenon~\cite{huang_surveyhallucinationlarge_2025, bang_hallulensllmhallucination_2025}.
A key distinction is between hallucinations that conflict with \textit{world knowledge} and those that are not properly grounded in the \textit{provided context}, for instance when a model generates details or conclusions that cannot be inferred from the supplied documents.
For public-sector applications, the latter is often the more pressing concern, as staff typically work with specific provided materials, making faithfulness to the input context critical.

\paragraph{Sustainability}
The environmental impact of LLMs has become an increasingly important governance criterion, as both training and inference contribute substantially to energy consumption and carbon emissions~\cite{strubell-etal-2019-energy,patterson-etal-2021-carbon}.
However, research has shown that the relevant factors tend to be complex.
For instance, \cite{wu-etal-2025-unveiling} argued that larger LLMs have a less severe impact than small models if request rates are kept low and a high output quality is achieved, while smaller models outperform larger ones if the request rate increases.
Furthermore, while quantization yields lower carbon emissions, new hardware tends to contain embodied carbon, which may increase its ecological footprint compared to older hardware.
Drawing from their own ecological LLM benchmark, \cite{jegham-etal-2025-hungry} emphasized that while LLM usage per token becomes cheaper, absolute usage increases, which accelerates the negative environmental consequences of LLMs.
Different tools have been suggested to measure the ecological impact of models.   
While direct energy measurement using hardware sensors or tools like CodeCarbon~\cite{codecarbon} is feasible for locally deployed models, this approach is not applicable when evaluating LLMs accessed through commercial APIs, where infrastructure details remain opaque.
To address this limitation, estimation-based approaches have been developed.
EcoLogits~\cite{rince-banse-2025-ecologits} is an open-source Python library that estimates the environmental footprint of generative AI inference by combining model-specific parameters with request metadata.
The framework takes a bottom-up approach, estimating GPU energy consumption and latency based on model architecture and number of output tokens.
Crucially, EcoLogits supports parameter estimates for major LLM providers including OpenAI and Anthropic, making it particularly suited for benchmarking scenarios that involve both open-weight and closed-source models accessed via APIs.
The framework reports multiple environmental impact metrics, including Global Warming Potential (GWP), primary energy consumption, and embodied impacts from hardware manufacturing.

\paragraph{Transparency}
Transparency in the context of AI systems refers broadly to the provision of information about the entire life cycle of an AI system and its ecosystem~\cite{_transparenzkisystemen_}, enabling stakeholders to make informed decisions about its use.
The concept has gained traction through proposals for standardized documentation practices, most notably \textit{Model Cards}~\cite{mitchell-etal-2019-model}, which provide structured summaries of model characteristics and intended uses, and \textit{Datasheets for Datasets}~\cite{gebru-etal-2021-datasheets}, which apply similar principles to training data.
Although these frameworks have been widely discussed as best practice in principle, the actual adoption and quality of documentation vary considerably between providers.
To systematically assess this gap, the Stanford Center for Research on Foundation Models introduced the \textit{Foundation Model Transparency Index} (FMTI)~\cite{bommasani-etal-2023-fmti}, which evaluates major foundation model providers across 100 indicators spanning data, compute, deployment, and downstream use.
The index has been updated yearly since 2023, with recent iterations reporting a notable decline in average transparency scores~\cite{wan20252025}, suggesting that the rapid pace of model releases may outstrip documentation efforts.

From a regulatory perspective, the EU AI Act~\cite{euaiact2024} has established transparency as a core obligation for providers of general-purpose AI models, codified in Article~53.
The accompanying \textit{GPAI Code of Practice}~\cite{gpaicontents} operationalizes these requirements by specifying documentation obligations across domains such as model identification, training data, and computational resources.
These regulatory developments provide a concrete, enforceable framework against which model documentation can be assessed, an approach we adopt in the \textit{MÖVE framework} through a structured \textit{Transparency Matrix} derived from the Code of Practice requirements.
For public-sector organizations, transparency is particularly consequential: public-sector employees tasked with procuring or deploying LLMs need reliable information about training data provenance, licensing conditions, and known limitations to ensure compliance with legal obligations and to justify technology decisions in a context of public accountability.
Without adequate documentation, informed model selection becomes difficult, effectively shifting the burden of risk assessment from the provider to the adopting institution.

\paragraph{Political Bias and Alignment}

In recent years, the question of politically biased LLM output has attracted increasing attention in the literature~\cite{hagendorff-2025-inevitability,rettenberger-etal-2025-assessing,rottger-etal-2024-political}. 
While being applied in different contexts, these studies often share an interest in \textit{model positioning} across the political spectrum, which is reflected in their adoption of popular tests used for assessing human political preferences. 
For instance, in a common evaluation approach models are prompted to complete the \textit{Political Compass Test}\footnote{\url{https://www.politicalcompass.org/}}, which consists of 62 propositions according to which participants are positioned along the two dimensions \textit{economic} and \textit{social}.
Based on the resulting scores, studies frequently argue that LLMs tend to generate content exhibiting a bias towards left-liberal perspectives, a pattern which tends to be framed as harmful \cite{hartmann-etal-2023-political,helwe-etal-2025-navigating,rozado-2023-political}.
Similar work has been conducted using the German Wahl-O-Mat\footnote{\url{https://www.bpb.de/themen/wahl-o-mat/} (in German)}, an online tool to compare positions of voters with those of political parties \cite{batzner-etal-2025-germanpartiesqa,rettenberger-etal-2025-assessing}. 
However, research has also shown that model output is highly dependable on context, i.e., prompt phrasings \cite{lunardi-etal-2024-elusiveness,rottger-etal-2024-political}, and language choice \cite{helwe-etal-2025-navigating}, which affects the observed bias in the data.
Other criticism includes \cite{hagendorff-2025-inevitability}, who argues that political left-wing bias is associated with a model alignment paradigm focused on producing harmless, helpful, and honest models \cite{bai-etal-2022-training}, and thus does not require mitigation.

We account for these findings by refraining from attributing model outputs to concrete partisan politics. 
Instead, we evaluate general model functionality to correctly classify party positions.
We further assess stance in model output with respect to values of the German constitution, i.e., values that are not associated with a specific party or political leaning.

%% file: sections/03_framework_overview.tex
\section{MÖVE Framework Overview}
\label{ch:framework_overview}

This section introduces the \textit{MÖVE framework} at a conceptual level: \textit{what} it evaluates, \textit{for whom}, and \textit{how} the evaluation is structured.
We first identify the target groups whose needs shaped the design of the framework (Section~\ref{sec:target_groups}), then define the performance and governance criteria that form the two pillars of the framework (Section~\ref{sec:criteria}), and describe the general evaluation setup that applies across criteria (Section~\ref{sec:evaluation_setup}).
The detailed methodology for each criterion, including specific datasets, metrics, and experimental designs, follows in Section~\ref{ch:methodology}.

\subsection{Target Groups}
\label{sec:target_groups}

Benchmark design should be grounded in stakeholder objectives and downstream use contexts, not assume a one-size-fits-all evaluation~\cite{reuel-etal-2024-betterbench}.
The design of the \textit{MÖVE framework} was accordingly guided by the needs of four primary stakeholder groups in the public-sector ecosystem as described in Table~\ref{tab:target_groups}.
In this sense, target groups are not merely an audience for benchmark results; they shape which criteria are relevant, how results should be presented, and what kinds of decisions the benchmark is ultimately meant to support. 
A detailed description of each group's requirements and typical interaction patterns is provided in Appendix~\ref{app:target_groups}.

These target groups were identified and validated through a stakeholder-informed design process.
We conducted semi-structured feedback sessions with representatives from ministries and other public-sector institutions, and engaged with a broader practitioner community through more than 20 sector-specific events and conferences, including the IT-Planungsrat\footnote{The IT-Planungsrat (IT Planning Council) is the central body for coordinating IT policy between the German federal and state governments.} Fachkongress in March 2025 and the AI Community of Practice Group of the BMDS\@.\footnote{Bundesministerium für Digitales und Staatsmodernisierung (Federal Ministry for Digital and State Modernisation).}
The resulting feedback informed iterative refinements of the benchmark's criteria, structure, and reporting logic.

\begin{table}[t]
    \centering
    \small
    \caption{Target groups of the MÖVE framework and their primary focus.}
    \label{tab:target_groups}
    \begin{tabular}{@{}p{0.30\linewidth}p{0.64\linewidth}@{}}
        \toprule
        \textbf{Stakeholder Group} & \textbf{Primary focus} \\
        \midrule
        AI decision-makers &
        Government decision-makers and public-sector leadership require a strategic basis for assessing LLMs, from a holistic, 360-degree perspective. Their focus lies on informed decision-making, responsible adoption and public trust.\\
        \addlinespace
        Domain experts &
       Public administration domain experts require models that produce accurate, understandable and practical outputs for day-to-day administrative tasks. Their focus is operational.\\
        \addlinespace
        IT departments and security-critical institutions &
        Technical stakeholders require risk-aware model evaluation that aligns with legal, organisational, and security requirements. Their focus lies on preventive measures, compliance, and safe deployment.\\
        \addlinespace
        Broader civil society &
        External stakeholders seek transparency, accountability, and the ability to critically assess the societal implications of LLM deployment. Their focus is democratic oversight.\\
        \bottomrule
    \end{tabular}
\end{table}

\subsection{Criteria}
\label{sec:criteria}

We structure evaluation in the \textit{MÖVE framework} around two groups of criteria that reflect the specific needs of the target groups described above: \textit{Performance Criteria} and \textit{Governance Criteria}.
Together, they provide a holistic perspective on a model's suitability for public-sector applications.

\subsubsection{Performance Criteria}
Performance criteria assess the direct quality of a language model’s output when applied to administrative tasks.
They include at the time of writing:

\begin{itemize}
    \item \textbf{Summarization:} Evaluation of whether a summary produced by the language model contains the essential points of a text and aligns with a human-preferred summary.
    \item \textbf{Question Answering (QA):} Assessment of whether the model generates a correct answer to a user-provided question based on a given text.
    \item \textbf{Topic Extraction:} Evaluation of whether a model generates the central topics within a text.
\end{itemize}

\subsubsection{Governance Criteria}
Governance criteria evaluate aspects of the model that go beyond task-specific performance. They include:

\begin{itemize}
    \item \textbf{Hallucination:} Examination of the extent to which the model generates false or misleading information.
    \item \textbf{Sustainability:} Evaluation of the model’s energy consumption and environmental impact.
    \item \textbf{Transparency:} Evaluation of information disclosure by model providers, in line with the requirements of the European AI Act and specifically with the General-Purpose AI (GPAI) code of practice and Model Documentation Form of the European commission.
    \item \textbf{Politics and Values:} Evaluation of how well the model’s generated output reflects the positions of the German political parties, as well as the alignment of the model's output with the fundamental principles of the German constitution.
\end{itemize}

\subsection{Evaluation Setup}
\label{sec:evaluation_setup}

As we describe below, we evaluate performance criteria using a unified, fully automated setup.
For governance criteria, we use a broader and more heterogeneous set of methodological approaches.
Figure~\ref{fig:evaluation_setup} summarizes the resulting evaluation modes.

\input{figures/evaluation_setup}

\subsubsection{Performance Evaluation}

For the evaluation of performance criteria we follow a standardized, fully automated procedure, in line with established methodologies in LLM benchmarking, e.g.~\cite{liang_holistic_2023,wang_superglue_2020,lintang_sutawika_eleutherailm-evaluation-harness_2024,srivastava_beyond_2023,hendrycks_measuring_2021,chang_survey_2024,cobbe_training_2021}.

Each evaluation setup consists of four components:

\begin{enumerate}
    \item \textbf{Task Definition:}
    a clearly specified task, such as summarization or question answering, aligned with the performance criteria described in Section~\ref{sec:criteria}.

    \item \textbf{Dataset:}
    a dataset providing input texts and reference responses.

    \item \textbf{Prompt:}
    a task-specific instruction (e.g., \verb|Summarize the following text:|).

    \item \textbf{Evaluation Metrics:}
    a set of metrics to quantify model output quality, including classical NLP metrics, embedding-based methods, and modern LLM-as-a-judge approaches.
\end{enumerate}

We apply each language model to all setups, compare outputs against reference responses, and aggregate results at two levels:

\begin{enumerate}
    \item \textbf{Setup Level:}
    across all instances within a dataset.

    \item \textbf{Task Level:}
    across all setups belonging to the same task category (e.g., summarization or question answering).
\end{enumerate}

\subsubsection{Governance Evaluation}

For the evaluation of governance criteria we use a more heterogeneous methodological structure.
Some governance criteria can be assessed using evaluation setups analogous to those used for performance evaluation.
For example, whether a model correctly produces party positions can be evaluated using a dataset of questions and reference answers together with a suitable classification metric.

Other governance aspects can be assessed by extending existing performance setups.
For instance, sustainability can be measured by tracking the energy consumption of a model while it executes one of the performance evaluation setups.

However, not all governance criteria can be meaningfully assessed through model interactions alone.
In particular, transparency, which captures alignment with the EU AI Act through the information disclosed in public model documentation, is instead evaluated via a structured analysis of model cards, technical reports, and other publicly available materials.

Regardless of whether the evaluation is automated or manual, every governance criterion is quantified through one or more metrics, allowing integration into the overall model evaluation framework.

A more detailed description of each evaluation is provided in Section~\ref{ch:methodology}.

%% file: figures/evaluation_setup.tex
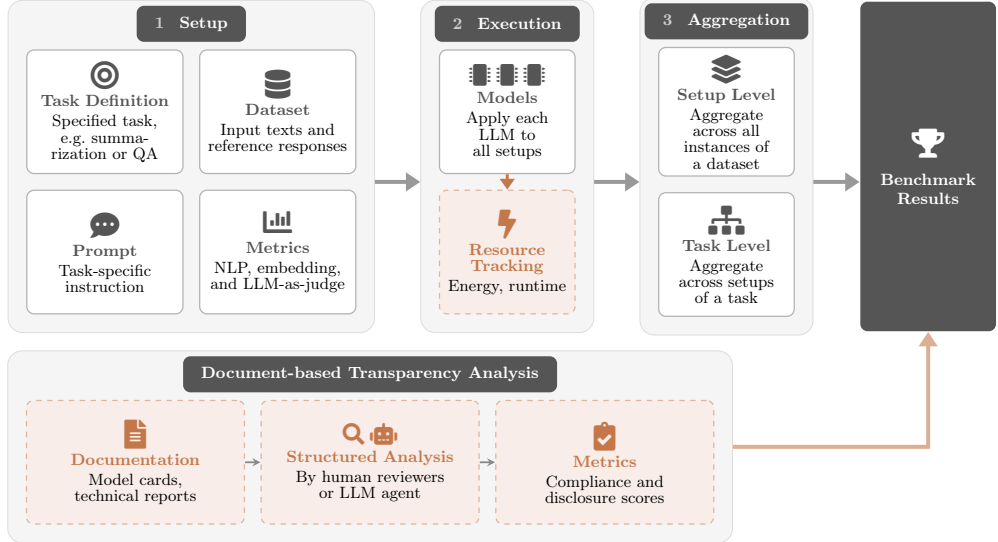
\begin{figure}[t]
    \centering
    \resizebox{\linewidth}{!}{%
    \begin{tikzpicture}[
        every node/.style={font=\small},
        comp/.style={
            rounded corners=4pt,
            minimum width=3.0cm,
            minimum height=2.4cm,
            text width=2.75cm,
            align=center,
            font=\scriptsize,
            inner sep=4pt,
            line width=0.7pt,
        },
        comp-auto/.style={comp, fill=white, draw=moeve-darkgray!45},
        comp-gov/.style={comp, fill=moeve-light-gov, draw=moeve-gov!55, dashed, minimum width=4.2cm, text width=3.95cm},
        comp-auto-narrow/.style={comp-auto, minimum width=2.6cm, text width=2.35cm},
        comp-gov-narrow/.style={comp-gov, minimum width=2.6cm, text width=2.35cm},
        group-header/.style={
            rounded corners=4pt,
            minimum height=0.7cm,
            align=center,
            font=\small\bfseries,
            text=white,
            inner xsep=10pt,
            inner ysep=3pt,
        },
        result-box/.style={
            rounded corners=4pt,
            minimum width=2.6cm,
            minimum height=6.4cm,
            text width=2.4cm,
            align=center,
            font=\small\bfseries,
            text=white,
            fill=moeve-darkgray,
            line width=0.7pt,
        },
        flow-arr/.style={
            -{Triangle[length=8pt, width=10pt]},
            line width=2.2pt,
            draw=moeve-darkgray!60,
        },
        small-arr/.style={->, >=stealth, semithick, draw=moeve-darkgray!70},
        side-arr/.style={
            -{Triangle[length=6pt, width=8pt]},
            line width=1.6pt,
            draw=moeve-gov,
            dash pattern=on 4pt off 2.5pt,
        },
        note/.style={font=\scriptsize\itshape, text=moeve-darkgray, align=center},
    ]
    \node[comp-auto] (task) {%
        {\Large\textcolor{moeve-darkgray}{\faIcon{bullseye}}}\\[2pt]
        \textcolor{moeve-darkgray}{\textbf{Task Definition}}\\[2pt]
        Specified task, e.g.\ summarization or QA};
    \node[comp-auto, right=0.3cm of task] (data) {%
        {\Large\textcolor{moeve-darkgray}{\faIcon{database}}}\\[2pt]
        \textcolor{moeve-darkgray}{\textbf{Dataset}}\\[2pt]
        Input texts and reference responses};
    \node[comp-auto, below=0.3cm of task] (prompt) {%
        {\Large\textcolor{moeve-darkgray}{\faIcon{comment-dots}}}\\[2pt]
        \textcolor{moeve-darkgray}{\textbf{Prompt}}\\[2pt]
        Task-specific instruction};
    \node[comp-auto, below=0.3cm of data] (metr) {%
        {\Large\textcolor{moeve-darkgray}{\faIcon{chart-bar}}}\\[2pt]
        \textcolor{moeve-darkgray}{\textbf{Metrics}}\\[2pt]
        NLP, embedding, and LLM-as-judge};
    \coordinate (setup-top) at ([yshift=0.6cm]task.north);
    \begin{scope}[on background layer]
        \node[fit=(task)(data)(prompt)(metr)(setup-top), rounded corners=8pt, fill=moeve-gray, draw=moeve-darkgray!20, inner xsep=10pt, inner ysep=10pt] (setup-box) {};
    \end{scope}
    \node[group-header, fill=moeve-darkgray, anchor=north] at ([yshift=-3pt]setup-box.north) (setup-h) {\textcolor{moeve-darkgray!30!white}{\textbf{1}}\;\, Setup};
    \coordinate (flow) at ($(task.south)!0.5!(prompt.north)$);
    \node[comp-auto-narrow, right=1.6cm of data, anchor=west] (model) {%
        {\large\textcolor{moeve-darkgray}{\faIcon{microchip}\hspace{3pt}\faIcon{microchip}\hspace{3pt}\faIcon{microchip}}}\\[2pt]
        \textcolor{moeve-darkgray}{\textbf{Models}}\\[2pt]
        Apply each LLM to all setups};
    \node[comp-gov-narrow, below=0.3cm of model] (energy) {%
        {\Large\textcolor{moeve-gov}{\faIcon{bolt}}}\\[2pt]
        \textcolor{moeve-gov}{\textbf{Resource Tracking}}\\[2pt]
        Energy, runtime};
    \draw[side-arr] (model.south) -- (energy.north);
    \coordinate (exec-top) at ([yshift=0.6cm]model.north);
    \begin{scope}[on background layer]
        \node[fit=(model)(energy)(exec-top), rounded corners=8pt, fill=moeve-gray, draw=moeve-darkgray!20, inner xsep=10pt, inner ysep=10pt] (exec-box) {};
    \end{scope}
    \node[group-header, fill=moeve-darkgray, anchor=north] at ([yshift=-3pt]exec-box.north) (exec-h) {\textcolor{moeve-darkgray!30!white}{\textbf{2}}\;\, Execution};
    \node[comp-auto-narrow, right=1.6cm of model, yshift=0cm] (setup-agg) {%
        {\Large\textcolor{moeve-darkgray}{\faIcon{layer-group}}}\\[2pt]
        \textcolor{moeve-darkgray}{\textbf{Setup Level}}\\[2pt]
        Aggregate across all instances of a dataset};
    \node[comp-auto-narrow, below=0.3cm of setup-agg] (task-agg) {%
        {\Large\textcolor{moeve-darkgray}{\faIcon{sitemap}}}\\[2pt]
        \textcolor{moeve-darkgray}{\textbf{Task Level}}\\[2pt]
        Aggregate across setups of a task};
    \coordinate (agg-top) at ([yshift=0.6cm]setup-agg.north);
    \begin{scope}[on background layer]
        \node[fit=(setup-agg)(task-agg)(agg-top), rounded corners=8pt, fill=moeve-gray, draw=moeve-darkgray!20, inner xsep=10pt, inner ysep=10pt] (agg-box) {};
    \end{scope}
    \node[group-header, fill=moeve-darkgray, anchor=north] at ([yshift=-3pt]agg-box.north) (agg-h) {\textcolor{moeve-darkgray!30!white}{\textbf{3}}\;\, Aggregation};
    \node[result-box, right=0.9cm of agg-box.east, anchor=west] (results) {%
        {\Large\faIcon{trophy}}\\[6pt]
        Benchmark\\Results};
    \draw[flow-arr] (setup-box.east |- flow) -- (exec-box.west |- flow);
    \draw[flow-arr] (exec-box.east |- flow) -- (agg-box.west |- flow);
    \draw[flow-arr] (agg-box.east |- flow) -- (results.west |- flow);
    \node[comp-gov, anchor=north west] at ([xshift=10pt, yshift=-1.3cm]setup-box.south west) (docs) {%
        {\Large\textcolor{moeve-gov}{\faIcon{file-alt}}}\\[2pt]
        \textcolor{moeve-gov}{\textbf{Documentation}}\\[2pt]
        Model cards,\\technical reports};
    \node[comp-gov, right=0.3cm of docs] (review) {%
        {\large\textcolor{moeve-gov}{\faIcon{search}\hspace{3pt}\faIcon{robot}}}\\[2pt]
        \textcolor{moeve-gov}{\textbf{Structured Analysis}}\\[2pt]
        By human reviewers\\or LLM agent};
    \node[comp-gov, right=0.3cm of review] (mmet) {%
        {\Large\textcolor{moeve-gov}{\faIcon{clipboard-check}}}\\[2pt]
        \textcolor{moeve-gov}{\textbf{Metrics}}\\[2pt]
        Compliance and disclosure scores};
    \draw[small-arr, line width=1pt] (docs.east) -- (review.west);
    \draw[small-arr, line width=1pt] (review.east) -- (mmet.west);
    \coordinate (man-top) at ([yshift=0.6cm]docs.north);
    \begin{scope}[on background layer]
        \node[fit=(docs)(review)(mmet)(man-top), rounded corners=8pt, fill=moeve-gray, draw=moeve-darkgray!20, inner xsep=10pt, inner ysep=10pt] (man-box) {};
    \end{scope}
    \node[group-header, fill=moeve-darkgray, anchor=north] at ([yshift=-3pt]man-box.north) (man-h) {Document-based Transparency Analysis};
    \draw[flow-arr, draw=moeve-gov!60] (man-box.east) -| ([xshift=0pt]results.south);
    \end{tikzpicture}%
    }
    \caption{Evaluation modes in the MÖVE framework. Most criteria (all performance criteria and most governance criteria) follow the shared automated pipeline (\emph{Setup} → \emph{Execution} → \emph{Aggregation} → \emph{Benchmark Results}). Two governance criteria are exceptions, shown in \textcolor{moeve-gov}{\textbf{orange}}. Sustainability is evaluated via the \emph{Resource Tracking} branch, which captures runtime measurements alongside model execution. Transparency is evaluated separately through a \emph{Document-based Transparency Analysis}, in which model cards and technical reports are scored by human reviewers and an LLM-based agent in alignment with the EU AI Act.}
    \label{fig:evaluation_setup}
\end{figure}

%% file: sections/04_methodology.tex
\section{Methodology}
\label{ch:methodology}

Having defined the evaluation criteria and the general setup in the previous section, we now describe the methodology in detail.
We begin with the cross-cutting elements that apply to all evaluations: the datasets we use and how they were constructed (Section~\ref{sec:datasets}), the models included in the benchmark (Section~\ref{sec:models}), the context window handling strategy (Section~\ref{sec:context-handling}), and the metrics employed (Section~\ref{sec:metrics}).
We then present the evaluation design for each individual criterion, covering performance criteria (Section~\ref{sec:performance_criteria}) and governance criteria (Section~\ref{sec:governance_criteria}).

\subsection{Datasets}
\label{sec:datasets}

Evaluating LLMs requires collecting or creating datasets, which can be used for the systematic assessment of model functionalities.
Given that we focus on the usability of language models in specific application areas of public administration, datasets need to fulfill the following conditions:
1) Data must come from the public administration domain.
Since different texts exhibit distinct linguistic characteristics that may affect LLM processing, domain-specific data enables more reliable evaluation of models for public-sector use cases.
2) Texts must adhere to task-specific formats, for example high-quality document-summary pairs for summarization.
To address the scarcity of German public administration datasets, which are often out of scope, low-quality, or unavailable, we employ two main approaches: a) employing existing datasets (e.g., from HuggingFace), and b) compiling our own datasets, which we create at different quality levels: high-quality (\textit{goldstandard}) or curated from public sources (\textit{silverstandard}). 
All were selected based on our performance and governance criteria.
An overview of the datasets is given in Table~\ref{tab:dataset_overview}.
In addition to these ten datasets, our governance evaluation draws on two further datasets that deviate from the conventional document/groundtruth format: the \textit{MÖVE Transparency Matrix}, comprising 21 questions on provider documentation (Section~\ref{sec:transparency}), and a set of eight \textit{constitutional values} used to probe model outputs (Section~\ref{sec:politics_values}).

\begin{table}[ht]
  \centering
  \caption{Overview of datasets. \emph{Type} indicates whether a dataset is pre-existing or specifically constructed for the benchmark. For the latter category, we distinguish between goldstandard and silverstandard datasets. Dataset size is reported with respect to the evaluation unit used in each task, for example, \emph{summaries} in summarization tasks, \emph{unique topics} in topic extraction tasks, or \emph{political statements} in the Wahl-O-Mat evaluation.}
  \label{tab:dataset_overview}
  \begin{tabular}{lccc}
    \toprule
    Dataset & Criterion & Type & Size \\
    \midrule
    Eur-Lex-Sum \cite{aumiller2022eur} & Summarization & Existing & 850 \\
    Swiss Leading Decision Summarization \cite{rasiah2023scale} & Summarization & Existing &  1,530 \\
    KIKC Summary & Summarization & Gold & 40 \\
    German Ministry Publications (Summaries) & Summarization & Silver & 1,530 \\
    German-QuAD \cite{moller-etal-2021-germanquad} & QA & Existing & 4,710 \\
    KIKC QA & QA & Gold & 72 \\
    FAQ Law & QA & Silver &  129 \\
    KIKC Topics & Topic Extraction & Gold & 205\\
    German Ministry Publications (Topics) & Topic Extraction & Silver & 4,199 \\
    Wahl-O-Mat & Politics \& Values & Existing & 4,788 \\
    \bottomrule
  \end{tabular}
\end{table}

\paragraph{Already Existing Datasets}

For every criterium, we begin with researching potentially usable datasets, that have previously been published by other researchers.
In particular, we concentrate on data that is thematically in scope, i.e. data that is related to public administration.
It further needs to be German data, which is permissively licensed.
Preferably but optionally, the data should be of substantial size and readily accessible in well-formatted form.
In total, we include 4 existing datasets in our benchmark.

\paragraph{Our Own Datasets}

Given that only a small amount of publicly existing datasets is available, we complement our data collection with our own datasets.
Depending on different factors such as size, complexity and feasibility of the task, we either manually constructed datasets, thereby ensuring high data quality, or collected suitable data from other sources.
In the case of the former, which we refer to as our \textit{goldstandard} datasets, we compiled a set of documents representative of the context of public administration and produced, for instance, summaries that can be used for model evaluation.
We proceeded similarly for the performance tasks \textit{Question Answering} and \textit{Topic Extraction}.
While having the benefit of a high level of control over the result, this procedure of dataset creation has the caveat of being costly both with respect to the time needed to accomplish the task and the required financial resources.
A suitable alternative are so-called \textit{silverstandard} datasets, which are not yet available as evaluation datasets but are already complete in terms of content.
For example, abstracts of text can be used as summaries.
Because such data sources are often available in large quantities, meticulous quality assurance is not feasible.
Instead, the datasets were reviewed on a sample basis to ensure a basic level of quality.
In total, we construct 6 gold- and silverstandard datasets (in addition to the \textit{Transparency Matrix} and the \textit{constitutional values} taxonomy).

\subsection{Models}
\label{sec:models}

We selected models in the MÖVE framework according to several criteria to ensure broad coverage of relevant application scenarios.
We focused primarily on \emph{open-weight} models, i.e. models whose parameters are publicly available, so that they support private deployment on on-premise data centers or local computing hardware. To enable comparison with leading proprietary systems, we additionally included selected models from OpenAI.
An overview of all evaluated models, including their parameters, context length, \textit{reasoning} functionality, and licensing, is provided in Table~\ref{tab:models}.

We consider the following aspects when selecting models for our benchmark:

\paragraph{Model Providers}
We included models covering both open-weight systems and prominent commercial models.
The selection ensures representation of the key organisations and systems relevant for public- and private-sector deployment, as detailed in Table~\ref{tab:models}.\footnote{The list of models reflects the state at the time of writing; the MÖVE framework is designed to support continuous integration of newly released models.}

\paragraph{Models Relevant for Public-Sector Deployment}
We selected models that are operationally used in German public sector applications.
At the time of the study (April 2026), this included models deployed in the ITZBund application \emph{KIPITZ}, which served as a representative reference case.\footnote{ITZBund (Informationstechnikzentrum Bund) is the central IT service provider of the German federal government. KIPITZ is an internal ITZBund application employing selected LLMs for public administration workflows.}

\paragraph{Small Language Models}
We separately considered models with fewer than approximately 20 billion parameters because of their lower computational requirements, improved energy efficiency, and suitability for local execution on high-end laptops or workstations.

\include{models_table}

\paragraph{German Fine-Tuned Models}
We included models fine-tuned for German language tasks due to their potential applicability in public administration environments, where German is the primary working language.

\paragraph{Licensing Considerations}
Licensing conditions for all evaluated models are summarized in Table~\ref{tab:models}.
We note that proprietary models (e.g., from OpenAI) generally require commercial licensing, while open-weight models are often freely deployable but may impose restrictions under certain circumstances.
This overview ensures transparency regarding operational and legal constraints when deploying the models. \\

\noindent \textbf{Technical Deployment Setup:} We deploy open-weight models on cloud-based virtual instances equipped with up to four NVIDIA A100 GPUs, using standard inference backends such as \texttt{vLLM}\footnote{\url{https://vllm.ai/}} and \texttt{ollama}\footnote{\url{https://ollama.com/}}.
Models whose resource requirements exceeded this configuration, as well as proprietary systems, are accessed via commercial APIs.
We use the models' default decoding parameters (e.g. temperature and top-\emph{p}), but, where configurable, we \emph{disable} additional safety or content-moderation settings in order to elicit the models’ unconstrained output; unless stated otherwise for a specific experiment, we retain the providers' default system prompts.

\subsection{Context Windows and Truncation Policy}
\label{sec:context-handling}

Across all evaluations, we apply a consistent policy for handling the limited context windows of large language models.
Depending on a model’s context window, some documents or prompts may be too long to process in full.
In these cases, we truncate the document from the end to fit the maximum available context length.
Importantly, we always reserve part of the context window for the model’s output tokens, so the effective capacity for the input text is smaller than the nominal context size.
We never truncate the task instructions or system prompts; truncation is applied only to the document content that is inserted into the prompt.

Table~\ref{tab:output_tokens} summarizes how many output tokens we reserve as a function of the model’s context window.
For standard models, the reserved output capacity increases stepwise with the context window, while for reasoning-oriented models we allocate a larger, fixed output capacity to account for extended chain-of-thought generations.

We chose not to implement multi-step processing algorithms in cases where the full document does not fit into the context window, such as chunking or map–reduce–style approaches, in order to focus the evaluation on the models’ inherent performance rather than on algorithmic strategies.
This design choice implicitly penalizes models with shorter context windows.
However, most modern models now support sufficiently long contexts to handle very lengthy documents, as shown in Table~\ref{tab:models}.

\begin{table}[ht]
  \centering
  \caption{Reserved output tokens as a function of the model’s context window.
  For reasoning-oriented models, we reserve a larger fixed number of output tokens to account for longer chain-of-thought outputs.}
  \label{tab:output_tokens}
  \begin{tabular}{lc}
    \toprule
    Context window (tokens) & Reserved output tokens \\
    \midrule
    $0$ -- $4{,}095$        & $512$ \\
    $4{,}096$ -- $31{,}999$ & $1{,}024$ \\
    $\geq 32{,}000$         & $2{,}048$ \\
    \midrule
    Reasoning models        & $8{,}192$ \\
    \bottomrule
  \end{tabular}
\end{table}

\subsection{Metrics}
\label{sec:metrics}

We evaluate the outputs of language models using a variety of metrics, which we group into three categories:

\begin{itemize}
    \item \textbf{Classical metrics:} These established metrics were used in natural language processing even before the development of large language models, for example in machine translation.
    They are often based on basic statistical methods, such as word distribution analysis, and can capture simple textual properties, but are limited in assessing complex meanings and nuances.

    \item \textbf{Embedding-based metrics:} These metrics use modern machine learning models, typically embeddings, to represent and compare texts.
    Unlike classical metrics, they are less sensitive to exact wording and can measure semantic similarity, enabling a more detailed and flexible analysis of text content.

    \item \textbf{LLM-as-a-judge metrics:} These metrics leverage large language models to compare and evaluate text content.
    Recent work suggests that they are particularly effective at assessing both the linguistic quality and the semantic depth of texts~\cite{zheng_judgingllmasajudgemtbench_2023,li2024llms}.
\end{itemize}

Since each metric category captures different aspects of text quality and none is sufficient on its own, the MÖVE framework draws primarily on embedding-based and LLM-as-a-judge metrics for its final scores, while classical metrics serve as complementary diagnostic indicators.
Across both performance and governance criteria, we combine metrics from these categories to obtain a more robust evaluation.
In addition, we implemented \textbf{task-specific metrics} for tasks not fully covered by the categories described above.
Table~\ref{tab:metrics_overview} provides an overview of all metrics included in the MÖVE framework.

\begin{table}[htbp]
    \footnotesize
    \centering
    \caption{Overview of metrics implemented in the MÖVE framework for evaluating performance and governance criteria.}
    \label{tab:metrics_overview}
    \begin{tabularx}{\textwidth}{l>{\raggedright\arraybackslash}p{3cm}X}
        \toprule
        \textbf{Metric} & \textbf{Task / Criterion} & \textbf{Description} \\
        \midrule
        \multicolumn{3}{l}{\textbf{Classical Metrics}} \\
        \midrule
        ROUGE & Summarization, Q\&A & N-gram based comparison against reference texts with focus on completeness (recall). \\
        BLEU & Summarization, Q\&A & N-gram based comparison against reference texts with focus on precision. \\
        Exact Match & Summarization, Q\&A, Topic Extraction & Case-insensitive exact-string comparison ignoring punctuation and numbers. \\
        \midrule
        \multicolumn{3}{l}{\textbf{Embedding-based Metrics}} \\
        \midrule
        BERTScore & Summarization, Q\&A & Contextual semantic similarity using word embeddings. \\
        SemScore & Summarization, Q\&A & Semantic similarity using embeddings of longer phrases (sentences or paragraphs). \\
        Semantic Topic Match & Topic Extraction & Embedding-based comparison of predicted and reference topic sets. \\
        Value Alignment (BERT) & Alignment with German constitution & Classifier-based assessment of outputs as \textit{pro}, \textit{neutral}, or \textit{contra} relative to the German constitution. \\
        \midrule
        \multicolumn{3}{l}{\textbf{LLM-as-a-Judge Metrics}} \\
        \midrule
        Factual Correctness & Summarization, Q\&A & LLM-based assessment of the factual accuracy of model outputs given a reference text. \\
        Faithfulness & Q\&A & LLM-based assessment of whether all information in the output is supported by the input text. \\
        Noise Sensitivity & Q\&A & Evaluation of robustness to irrelevant information in input. \\
        Topic Adherence & Topic Extraction & LLM-based comparison of topic sets, augmented with topic/word length factors to penalize overly short or overly long topic lists. \\
        Value Alignment (LLM) & Alignment with German constitution & LLM-based assessment of outputs as \textit{pro}, \textit{neutral}, or \textit{contra} relative to the German constitution. \\
        \midrule
        \multicolumn{3}{l}{\textbf{Task-specific}} \\
        \midrule
        Political Parties & Party Stance Classification & Accuracy of model outputs compared to reference party positions. \\
        German Summary Proportion & Summarization & Proportion of model outputs written in German. \\
        Sustainability & All tasks & Quantitative metrics including energy usage and estimated carbon footprint derived from model and token statistics. \\
        \bottomrule
    \end{tabularx}
\end{table}

A more detailed description of the specific metrics used for each evaluation criterion is provided in the sections that follow.

\subsection{Performance Criteria}
\label{sec:performance_criteria}

\subsubsection{Summarization}
\label{sec:summarization}
Public-sector employees regularly work with extensive documents such as policy briefs, legal documents, regulatory reports, or technical assessments.
Automatic summarization can help by generating short, decision-ready formats, reducing workload and allowing staff to concentrate on the subsequent tasks supported by those summaries.

In the following, we first describe our evaluation setup for abstractive summarization in the public sector context, then introduce the metrics used to quantify summarization quality, and finally present the datasets on which the evaluations are carried out.

\paragraph{Evaluation Setup}
To evaluate LLMs' applicability to summarization scenarios, we developed several evaluation setups.
Each setup contains real German language documents from public administration, covering a range of themes, structures, and lengths, together with manually created reference summaries in an abstractive form, i.e. summaries that paraphrase and condense the original text.

For each dataset, we provide a simple German prompt that yields a summary of the given document from the model, along with brief guidelines regarding the expected output format, such as whether it should include bullet points or free text and the approximate length.

For all summarization experiments, we follow the general context handling policy described in Section~\ref{sec:context-handling}, i.e., inputs are truncated to the model’s effective context capacity and a fixed number of tokens is reserved for the generated summary.

\paragraph{Evaluation Metrics}
As primary summarization score we compute an average of four metrics: \textit{BERTScore}, \textit{Semscore}, \textit{Factual Correctness}, and the proportion of the generated summaries that are in German.

\subparagraph{BERTScore}
To assess the semantic similarity between a model-generated summary and a human reference, we use \textit{BERTScore}~\cite{zhang_bertscore_2020}.
In contrast to lexical-overlap metrics (e.g. \textit{ROUGE}~\cite{lin_rouge_2004}), \textit{BERTScore} compares contextualized embeddings of candidate and reference tokens and derives precision, recall, and F1 based on pairwise cosine similarities.
This makes it well suited for evaluating abstractive summaries that rely on paraphrasing rather than verbatim reuse of the source text.
In the \textit{MÖVE framework}, we employ the \texttt{bert-base-multilingual-cased} encoder~\cite{DBLP:journals/corr/abs-1810-04805} and compute the F1 variant of the score, averaged at the document level.
However, since \textit{BERTScore} may reward semantically related but factually incorrect content, we complement it with metrics that capture semantic coverage and factual grounding.

\subparagraph{Semscore}
\textit{Semscore}~\citep{aynetdinov_semscore_2024} provides a measure of semantic alignment between a generated summary and a human reference by embedding each summary as a single unit.
\textit{Semscore} is thus more robust to reordering, stylistic variation, and structural differences commonly found in abstractive summarization compared to token-based metrics such as \textit{BERTScore}.
This is particularly relevant for long administrative documents, where detecting and condensing major themes is more important than reproducing specific phrasings.
In the \textit{MÖVE framework}, candidate and reference summaries are embedded using the \texttt{jina-embeddings-v2-base-de} model~\citep{mohr2024multi}, and similarity is computed at the summary level to yield a single score per document.
\textit{Semscore} thereby offers a complementary perspective to \textit{BERTScore}, focusing on high-level content representation rather than fine-grained lexical correspondence.

\subparagraph{Factual Correctness}
\textit{Factual Correctness} is an LLM-as-a-judge–based metric implemented through the \texttt{Ragas} framework~\citep{es_ragasautomatedevaluation_2025}.
In contrast to embedding-based similarity metrics, it evaluates the degree to which the information contained in a generated summary is factually supported by the reference text.
The metric proceeds in two LLM-driven steps: first, the model extracts all relevant statements (“claims”) from both the candidate summary and the reference text; second, each claim in the summary is checked for support by any claim in the reference set, and vice versa, yielding precision, recall, and an overall F1 score.
In the \textit{MÖVE framework}, both steps are performed using \texttt{GPT-4o Mini}.
To reduce stochastic variance in the judge model, we set the sampling temperature to $0$ and fix a random seed; nevertheless, small fluctuations in the resulting scores can still occur.
In principle, this variance could be further reduced by querying the judge model multiple times per example and averaging the scores, but we deliberately refrain from doing so in order to limit the computational and monetary cost of the evaluation.
The final \textit{Factual Correctness} score used in our evaluation is the resulting F1 measure.
By operating at the level of explicit claims, the metric was designed to approximate the human-like evaluation paradigm of decomposing text into atomic factual statements~\cite{min_factscorefinegrainedatomic_2023}, thus complementing embedding-based measures by targeting a distinct and critical aspect of summary quality.

\subparagraph{German Summary Proportion}
The fourth primary metric integrated into the overall summarization score is the proportion of generated summaries that are classified as being written in German.
In our experiments, we observe that some models occasionally produce output that is fully or partially in other languages, even when both the instructions and the source documents are in German.
Since our three other primary metrics exhibit a degree of invariance to language, capturing semantic or factual alignment even when parts of the text are in a different language, we explicitly include this metric to reflect that the principal aim of our benchmark is to evaluate German language generation.

To compute the German proportion, we employ a FastText-based language identification model~\cite{joulin2016bag}.
For each generated summary, the model outputs a probability distribution over languages; we use the predicted probability for German as a document-level indicator of how likely the entire summary is to be in German, distinguishing coherent German text that may legitimately contain foreign vocabulary from summaries that exhibit spurious artefacts of foreign language or are predominantly written in a different language.
We classify a summary as German if this probability exceeds a fixed threshold, which we calibrated on a manually annotated dataset of 200 LLM-generated summaries labeled as German or non-German.
The German proportion is then defined as the fraction of summaries in an evaluation run that are classified as German, and this fraction is incorporated into the aggregated summarization score so that systems producing substantial non-German output receive lower overall scores.

\bigskip

In addition to the four primary metrics described above, the \textit{MÖVE framework} also computes traditional surface-level metrics, namely \textit{ROUGE}~\citep{lin_rouge_2004}, \textit{BLEU}~\citep{papineni_bleumethodautomatic_2001}, and \textit{exact match} scores.
These metrics are reported to provide additional context regarding lexical overlap and n-gram precision, which can be informative for certain use cases.
However, we did not include them in the overall summarization score because they primarily capture surface similarity rather than deeper semantic alignment or factual correctness.
Including them in the overall score could overemphasize exact wording at the expense of meaning and factual reliability, which are critical for long-form administrative documents.

\paragraph{Datasets}
Having defined the evaluation setup and core metrics for summarization, we now turn to the underlying datasets on which these evaluations are performed.
In particular, our benchmark combines existing large-scale resources with high-quality, manually curated gold- and silverstandard data from the public sector.

\subparagraph{Eur-Lex-Sum}
The dataset~\cite{aumiller2022eur} is a public multilingual summarization corpus situated in the legal domain.
It contains summaries of legislative acts of the European Union that were originally written in English and subsequently translated by professional translators into the official EU languages.
In total, the German subset comprises 1{,}490 documents with corresponding abstractive summaries formatted as continuous text, from which we have sampled 850 documents for use in the \textit{MÖVE framework}.

To determine an adequate sample size, we conducted a stability analysis of our evaluation metrics.
Starting from a small subset, we incrementally increased the number of documents and monitored the evolution of the summarization metrics described above, averaging results over multiple model runs.
We selected 850 examples as the smallest sample size at which, for 20 consecutive increments in sample size, successive changes in the estimated mean metric were smaller than $0.0005$ (in metric units) and the width of the $95\%$ confidence interval around the mean was smaller than $0.05$ (in metric units).

Source documents in the resulting dataset contain on average 13{,}689 lexical tokens, while the reference summaries contain on average 818 lexical tokens, resulting in a mean token compression ratio of approximately 17.1.
Models are prompted in German with the following instruction (see Appendix~\ref{app:user_prompts} for the original prompt):
\enquote{You are a consultant specializing in summarizing legal texts. Provide a summary of the following European law that gives the reader a comprehensive overview of the law.}

\subparagraph{Swiss Leading Decision Summarization}
The dataset~\cite{rasiah2023scale} is likewise a multilingual legal summarization corpus, but based on court decisions rather than legislative acts.
It contains summaries of decisions of the Swiss Federal Supreme Court in German, French, and Italian.
The dataset includes 12{,}415 German documents with corresponding abstractive summaries that are again provided as continuous text.
For our benchmark, we sample 1{,}530 German document–summary pairs using the same procedure as described above for the \textit{Eur-Lex-Sum} dataset, incrementally increasing the sample size until the mean summarization metrics and their confidence intervals stabilized.
On average, source documents in this sampled subset contain 1{,}951 lexical tokens and their reference summaries 75 lexical tokens, corresponding to a mean document-to-summary compression rate of approximately 30.8.
Models are prompted in German to produce a concise \enquote{Regeste} (a brief headnote-style case summary typical for Swiss Federal Supreme Court decisions) tailored to the legal context of Swiss court decisions using the following instruction (see Appendix~\ref{app:user_prompts} for the original prompt): \enquote{Generate a brief summary (regeste) for the following case details from Swiss court rulings:}.

\subparagraph{KIKC Summary}
The dataset is an example of a goldstandard corpus specifically constructed for use in the \textit{MÖVE framework}.
It consists of 29 publicly available documents (pdf and docx) that are representative of materials used by staff of German ministries and other public sector institutions.
The raw text for these documents was extracted from the original pdf and docx files using Azure Document Intelligence.\footnote{\url{https://azure.microsoft.com/en-us/products/ai-services/ai-document-intelligence}}
The documents cover a broad range of topics relevant to public administration, including digital transformation and data policy, artificial intelligence, global economic relations, and European/German industrial and innovation policy.
Several of the source documents are very long (spanning several hundred pages), so we split them into chapters and treat these chapters as individual documents for the evaluation.
This results in a final dataset of 40 documents, each paired with a high-quality bullet point summary prepared by subject matter experts.
The preference for bullet point summaries reflects frequent feedback from our collaboration with administrative staff, who often request concise, list-based formats to support rapid orientation and decision-making.
Expert annotators included members of our team with extensive experience working with public administration, complemented by an external provider whose summaries were subject to a second expert review and correction.
On average, documents in the dataset contain approximately 2{,}991 lexical tokens, while the corresponding summaries contain around 333 lexical tokens, yielding a mean document-to-summary compression rate of about 11.7.

Models are prompted in German to generate concise numbered bullet point summaries of the given article using the following instruction (see Appendix~\ref{app:user_prompts} for the original prompt):
\enquote{You are a consultant specializing in summarizing articles. Your task is to write a summary in numbered bullet points of the following text. Do not include any information that is not in the article. Please summarize the article in five to a maximum of eight bullet points.}

\subparagraph{German Ministry Publications}
The dataset consists of German-language publications collected from the public web portal of a federal ministry.\footnote{We deliberately omit the explicit source to reduce the risk that this curated collection is incorporated into future model training data.}
The portal provides, for each publication, both the original document (in pdf format) and an associated summary formatted as continuous text that is not part of the document itself but appears on the publication’s landing page.
The documents span a broad range of policy areas relevant to public administration, including environment and climate policy, energy and transport, economic and consumer policy, digitalisation, and sustainability.
In total, the source collection comprises more than 2{,}000 pdf documents.

For use in the \textit{MÖVE framework}, we sampled 1{,}530 document–summary pairs from this collection.
The pdf files are converted to markdown using the \texttt{docling}\footnote{\url{https://github.com/docling-project/docling}} library, which executes document layout analysis and text extraction.
We retain only those cases for which both the document and the corresponding landing-page summary satisfy minimum length requirements of 300 and 50 tokens, respectively.
In the resulting dataset, the source documents contain on average 23{,}092 lexical tokens, while the reference summaries contain on average 91 lexical tokens, yielding a mean document-to-summary compression rate of approximately 273.8.
This extreme compression highlights the challenge of condensing long policy documents into very short, decision-oriented abstracts.

Models are prompted in German as follows (see Appendix~\ref{app:user_prompts} for the original prompt):
\enquote{Write a factual and concise summary that gives readers an overview of the following document:}

\bigskip

To provide a concise overview, Table~\ref{tab:summarization_datasets} summarizes the main properties of the four summarization datasets used in the \textit{MÖVE framework}.

\begin{table}[t]
  \centering
  \small
  \caption{Main properties of the four summarization datasets used in the \textit{MÖVE framework}.
  For each dataset, we report the number of document--summary pairs used in our benchmark, the average document and reference-summary length in lexical tokens, and the resulting mean document-to-summary compression ratio.}
  \label{tab:summarization_datasets}
  \begin{tabular}{lrrrr}
    \toprule
    Dataset & \# Docs &
    \makecell[r]{Doc len\\(tokens)} &
    \makecell[r]{Sum.\ len\\(tokens)} &
    Compression \\
    \midrule
    Eur-Lex-Sum & 850   & 13{,}689 & 818 & 17.1 \\
    Swiss Leading Decision Summarization & 1{,}530 & 1{,}951 & 75  & 30.8 \\
    KIKC Summary & 40    & 2{,}991 & 333 & 11.7 \\
    German Ministry Publications & 1{,}530 & 23{,}092 & 91  & 273.8 \\
    \bottomrule
  \end{tabular}
\end{table}

\subsubsection{Question Answering}
\label{sec:question_answering}

Administrative staff frequently need to retrieve specific information from lengthy documents rather than survey their full content.
\textit{Question Answering} (QA) supports this by allowing users to pose targeted questions and obtain focused answers, complementing summarization with greater flexibility, since the desired information can be steered through the question itself.

\paragraph{Evaluation Setup}
Our setup combines elements of \textit{reading comprehension}~\citep{rajpurkar-etal-2016-squad}, where models answer questions against a provided passage, with \textit{abstractive QA}~\citep{fan_eli5longform_2019}, in which answers are generated rather than extracted as verbatim spans.
The context is provided directly as input rather than incorporating a retrieval component, as the evaluation targets LLM output quality on QA tasks rather than the performance of complete QA pipelines.

For each dataset, we provide a German prompt that instructs the model to answer a given question based on the provided context, together with brief guidelines regarding the expected output format and length.

For all QA experiments, we follow the general context handling policy described in Section~\ref{sec:context-handling}, i.e., inputs are truncated to the model's effective context capacity and a fixed number of tokens is reserved for the generated answer.

\paragraph{Evaluation Metrics}

The overall QA score is defined as the average of the four metrics: \textit{Semscore}, \textit{Factual Correctness}, \textit{Faithfulness}, and \textit{Noise Sensitivity}.
Since \textit{Semscore} and \textit{Factual Correctness} were already introduced in Section~\ref{sec:summarization}, we describe here only the QA-specific metrics, \textit{Faithfulness} and \textit{Noise Sensitivity}.

\subparagraph{Faithfulness}
\textit{Faithfulness} captures the degree to which a generated answer is grounded in the provided context rather than in the model's parametric knowledge, an aspect that is critical in document-grounded administrative QA where users expect answers anchored in a specific regulation, FAQ, or policy text.
Like \textit{Factual Correctness}, it is implemented as an LLM-as-a-judge metric through the \texttt{Ragas} framework~\citep{es_ragasautomatedevaluation_2025}.
The metric proceeds in two LLM-driven steps: first, all claims are extracted from the generated answer; second, each claim is checked for whether it can be inferred from the provided context.
The final \textit{Faithfulness} score is then defined as the ratio of supported claims to the total number of claims in the answer.
In the \textit{MÖVE framework}, both steps are performed using \texttt{GPT-4o Mini}, with the sampling temperature set to $0$ and a fixed random seed to reduce stochastic variance in the judge model.

\subparagraph{Noise Sensitivity}
\textit{Noise Sensitivity} captures how susceptible a model is to irrelevant or noisy information in the provided context, that is, the extent to which extraneous passages induce incorrect claims that the model would not have generated otherwise.
Such noise can arise either from unrelated material within a single context document or from the retrieval of multiple contexts of which only some are relevant to the query.
Like \textit{Faithfulness}, it is implemented as an LLM-as-a-judge metric through the \texttt{Ragas} framework~\citep{es_ragasautomatedevaluation_2025}.
The model's output is first decomposed into individual claims, which are evaluated for correctness with respect to a reference answer; incorrect claims are then checked for entailment by the provided context.
The final \textit{Noise Sensitivity} score is defined as the ratio of incorrect claims attributable to the context to the total number of claims in the answer.
As for the other Ragas metrics, we use \texttt{GPT-4o Mini} with the sampling temperature set to $0$ and a fixed random seed.

\bigskip

As for summarization, the \textit{MÖVE framework} also computes \textit{BERTScore}~\citep{zhang_bertscore_2020}, \textit{ROUGE}~\citep{lin_rouge_2004}, \textit{BLEU}~\citep{papineni_bleumethodautomatic_2001}, and \textit{exact match} scores for QA.
These additional metrics are reported for transparency but are not included in the overall QA score, for the same reasons outlined in the summarization section.

\paragraph{Datasets}

We compiled three QA datasets for our benchmark, combining one widely used existing resource for German QA with a manually curated gold-standard set from the public-administration domain and a silver-standard collection derived from official ministerial FAQ pages.
All datasets share the same structure of question-answer-context triplets; corpus-size statistics are summarized in Table~\ref{tab:qa_datasets}.

\subparagraph{German-QuAD} 
Collected from Wikipedia, German-QuAD~\citep{moller-etal-2021-germanquad} consists of German question-answer pairs together with the corresponding contexts from which the answers can be extracted.
It is based on the German versions of the Wikipedia articles used for SQuAD~\citep{rajpurkar-etal-2016-squad}, an English QA dataset.
Although German-QuAD is strictly speaking out-of-domain with respect to public administration and was originally constructed for \textit{extractive} rather than \textit{abstractive} QA, we still include it as an influential reference dataset for German QA: our evaluation relies on embedding- and LLM-based metrics, that is, metrics that do not require exact token matching, which makes the data suitable for our setting.
Our subset consists of 4,710 questions with their answers and contexts.
On average, contexts contain 214 lexical tokens and answers 9 lexical tokens.
Models are prompted in German to produce short, extractive-style answers from the provided context, using the following instruction (see Appendix~\ref{app:user_prompts} for the original prompt):
\enquote{You are an assistant for processing question-answer tasks. Use the following context to answer the question. If the answer cannot be derived from the given information, say \enquote*{No answer}. Keep your answers as brief as possible, using no full sentences where possible. Context: \texttt{\{context\}} Question: \texttt{\{question\}}}

\subparagraph{KIKC QA}
We constructed the KIKC QA dataset to allow for a high degree of control with respect to data quality.
It includes 72 questions, answers, and contexts derived from nine publicly available documents in the context of public administration.
The documents represent a subset of those used in the KIKC Summary dataset.
Questions have been designed such that they include different complexities ranging from questions requiring a sophisticated answer to questions whose answer could directly be extracted.
On average, contexts contain 394 lexical tokens and answers 32 lexical tokens.
Models are prompted in German to generate a precise answer grounded in the provided context, using the following instruction (see Appendix~\ref{app:user_prompts} for the original prompt):
\enquote{You are an assistant for processing question-answer tasks. Use the following extracted context to answer the question. If the answer cannot be derived from the given information, sage \enquote*{The answer is not found in the context.}. Use a maximum of five sentences and keep the answer precise. Context: \texttt{\{context\}} Question: \texttt{\{question\}}}

\subparagraph{FAQ-LAW}
This dataset consists of 129 questions and their respective answers and contexts derived from official FAQ pages maintained by German government ministries for three German laws.
The laws focus on social as well as research-related issues.   
On average, contexts contain 5{,}573 lexical tokens and answers 78 lexical tokens.
Unlike the KIKC QA dataset which uses document chunks, FAQ-LAW provides the full law text as context, which explains the substantially larger context length.
Models are prompted in German to generate an answer grounded in the provided law text, using the following instruction (see Appendix~\ref{app:user_prompts} for the original prompt):
\enquote{You are an assistant that answers questions about legal texts and makes them understandable for citizens. Always refer to the following legal text to answer the question. Use a maximum of five sentences in your answer and keep the answer generally comprehensible. Law: \texttt{\{context\}} Question: \texttt{\{question\}}}

\begin{table}[t]
  \centering
  \small
  \caption{Main properties of the three QA datasets used in the \textit{MÖVE framework}.
  For each dataset, we report the number of questions used in our benchmark and the average context and answer length in lexical tokens.}
  \label{tab:qa_datasets}
  \begin{tabular}{lrrr}
    \toprule
    Dataset & \# Questions &
    \makecell[r]{Context len\\(tokens)} &
    \makecell[r]{Answer len\\(tokens)} \\
    \midrule
    German-QuAD & 4{,}710 & 214 & 9\\
    KIKC QA & 72 & 394 & 32\\
    FAQ-LAW & 129 & 5{,}573 & 78\\
    \bottomrule
  \end{tabular}
\end{table}

\subsubsection{Topic Extraction}
\label{sec:topic_extraction}
Topic extraction, i.e.\ generating short lists of keywords that represent the central themes of a document, is another key application to reduce the workload of administrative staff in handling extensive documents.
Compared to the summarization tasks described above, topic extraction provides an even more compact, high-level overview and is often used as a first orientation aid that facilitates the staff's decision whether and how they should engage further with a document.

Our evaluation setup for topic extraction closely mirrors the one used for summarization, with the main difference that models are prompted to output short lists of topic keywords instead of summaries for each document.
We nevertheless rely on distinct metrics and datasets tailored specifically to this task and describe them in detail in the following.

\paragraph{Evaluation Metrics}

We define our overall Topic Extraction score as the average of two metrics, \textit{Semantic Topic Match} and \textit{Topic Adherence}, both of which we design specifically for the \textit{MÖVE framework} to address the particular requirements of topic extraction in long administrative documents.

\subparagraph{Semantic Topic Match}
This metric uses semantic similarity in an embedding space to measure how well a model’s extracted topics match a reference set of topics and hence allows for fuzzy matching of topics that are semantically similar but not lexically identical.
For each predicted topic, it computes an embedding and determines the maximum cosine similarity (mapped to $[0,1]$) to any reference-topic embedding; the average of these values yields a precision score.
Analogously, it computes the average over the maximum similarities of each reference topic to any predicted topic to obtain recall.
As final score it computes the F1 measure as the harmonic mean of precision and recall.
As embedding model, we use \texttt{jina-embeddings-v2-base-de}~\citep{mohr2024multi}, which we also employ for the \textit{Semscore} metric described above.

\subparagraph{Topic Adherence}
\textit{Topic Adherence} is an LLM-driven metric that we introduce in this work to measure how closely a model’s extracted topics align with a reference set of topics.
Similar to \textit{Semantic Topic Match}, it computes precision, recall, and F1 scores over topic pairs; however, instead of relying on embedding-based semantic similarities, an LLM is prompted to generate a binary judgment on whether a predicted topic matches a given reference topic.
The prompt configures the LLM to normalize case, exclude trivial punctuation and pluralization, accept synonyms and close paraphrases, resolve acronyms, and incorporate domain knowledge in match evaluation.

To compute the final \textit{Topic Adherence} score, we augment the F1 score with two structural penalties that down-weight overly long or structurally mismatched topic predictions.
We designed this extension because some models, despite explicit instructions, produced very long lists of topics (sometimes several hundred entries) or generated full-sentence or paragraph-style outputs instead of short keyword-like topics.
Such behavior is not adequately penalized by the F1 score alone, as extensive lists may yield high recall despite many incorrect matches, and the LLM-based judge may still assign matches to long, sentence-like topics that overlap semantically with the reference topics even though they violate the intended short, keyword-style format.

First, we compute a \emph{word factor} that measures how many more words the predicted topics contain compared to the reference topics; this factor is defined as the ratio between the total number of words in the predicted and reference topic lists, clamped to be at least 1, so that only excessively verbose predictions are penalized.
Second, we compute a \emph{topic factor} that captures the relative difference in the number of topics between prediction and reference; this factor equals 1 when the counts match and grows linearly with the absolute deviation in topic count.
The final \textit{Topic Adherence} score is then obtained by multiplying the F1 score by the inverse of both factors, thereby down-weighting high F1 scores for predictions that are unnecessarily wordy or that introduce too many topics.

\bigskip

In addition to these two primary metrics, the \textit{MÖVE framework} also reports an exact-match topic overlap F1 score based on string-identical matches between predicted and reference topics.
This metric provides a familiar surface-level perspective and can be informative when models produce highly similar labels.
However, we do not include it in the overall Topic Extraction score, as it is overly sensitive to minor lexical variation and thus fails to adequately capture semantic similarity and domain-appropriate paraphrasing, which are central for evaluating topic extraction on long administrative documents.

\paragraph{Datasets}

We compiled two Topic Extraction datasets for our benchmark. See Table~\ref{tab:topic_extractions_datasets} for statistics on corpus size.

\begin{table}[t]
  \centering
  \small
  \caption{Main properties of the two topic-extraction datasets used in the \textit{MÖVE framework}.
  For each dataset, we report the number of documents, the mean document length in lexical tokens, summary statistics of the number of topics per document, and the size of the topic vocabulary.}
  \label{tab:topic_extractions_datasets}
  \setlength{\tabcolsep}{4pt}
  \begin{tabular}{lrrrrrr}
    \toprule
    Dataset & \# Docs &
    \makecell[r]{Doc len\\(tokens)} &
    \makecell[r]{Topics\\per doc\\(mean)} &
    \makecell[r]{Topics\\per doc\\(min)} &
    \makecell[r]{Topics\\per doc\\(max)} &
    \makecell[r]{Unique\\topics} \\
    \midrule
    KIKC Topics & 40 & 2{,}991 & 5.75 & 1 & 13 & 205 \\
    German Ministry Publications & 1{,}530 & 22{,}455 & 5.49 & 2 & 23 & 4{,}199 \\
    \bottomrule
  \end{tabular}
\end{table}

\subparagraph{KIKC Topics}
The dataset is a gold-standard topic-extraction corpus constructed specifically for use in the \textit{MÖVE framework}.
Its documents are identical to those in the \textit{KIKC Summary} dataset described above and were selected to be representative of materials used in projects with staff from German ministries and other public-sector institutions.
While compiling the summaries, expert annotators were additionally asked to provide a short list of topic keywords per document that capture the central themes, yielding on average 5.75 topics per document and a total of 205 distinct topics across the corpus (see Table~\ref{tab:topic_extractions_datasets}).
Models are prompted in German to generate single-word keywords, using the following instruction (see Appendix~\ref{app:user_prompts} for the original prompt):
\enquote{You are a German-speaking expert in political and administrative texts and create summarizing keywords for documents. Summarize the content of the document in single-word keywords and return a maximum of five keywords, comma-separated and without any additional text.}

\subparagraph{German Ministry Publications}
The dataset is derived from the same source as the summarization corpus described above, but we sample a distinct set of 1{,}500 documents for the Topic Extraction task.
For each document, the web-portal provides a list of topics that we use as reference labels, with a comparatively wide range in the number of topics per document (see Table~\ref{tab:topic_extractions_datasets}).
On average, documents in this dataset are annotated with 5.49 topics, with the number of topics per document ranging from 2 to 23 and a total of 4{,}199 unique topics across the corpus.
As for summarization, the source documents cover a broad spectrum of policy areas relevant to public administration, including environment and climate policy, energy and transport, economic and consumer policy, digitalisation, and sustainability.
Models are prompted in German to generate keywords, using the following instruction (see Appendix~\ref{app:user_prompts} for the original prompt):
\enquote{You are a German-speaking expert in political and administrative texts. Your task is to create concise, summarizing keywords from the following document. Return only comma-separated keywords — without any additional text — and generate approximately \texttt{\{topic\_count\}} keywords.}

\subsection{Governance Criteria}
\label{sec:governance_criteria}

\subsubsection{Hallucination}
\label{sec:hallucination}

Hallucination, i.e.\ the generation of content that is not grounded in the provided input or that contradicts factual knowledge, is a central concern when deploying LLMs in public administration.
In our interactions with administrative staff, this issue is consistently raised as one of the primary barriers to adoption: when working with official documents, policy briefs, or legal texts, model outputs must be reliable and faithful to the source material.
An answer that sounds plausible but introduces unfounded claims can lead to misinformed decisions, undermine the trustworthiness of AI-augmented workflows, and, in the worst case, cause legal or administrative complications.

From an application perspective, hallucination risks can be mitigated through traceability features that link generated outputs back to their source documents.
For instance, a document-based QA system can mark the relevant passages in the original text that support each claim in the generated answer, allowing users to verify the information at a glance.
Such features are essential for building trustworthy applications in the public sector.
However, traceability alone does not eliminate the underlying problem: if a model frequently produces hallucinations, users face a high verification burden, which diminishes the efficiency gains that LLM-based tools are meant to provide.
Consequently, selecting models that exhibit lower hallucination tendencies from the outset is a complementary strategy that can substantially improve the user experience and reduce the risk of undetected errors.

\paragraph{Evaluation Setup}
Our current approach to measuring hallucination in the \textit{MÖVE framework} focuses on hallucinations with respect to a given context and builds directly on components already used in our QA evaluation.
Rather than introducing a separate hallucination benchmark with additional datasets, the \textit{MÖVE framework} reuses the \textit{Faithfulness} metric computed as part of the QA evaluation described in Section~\ref{sec:question_answering}.
\textit{Faithfulness}, a RAGAS metric~\citep{es_ragasautomatedevaluation_2025}, quantifies the proportion of claims in a generated answer that are supported by the provided context.
By aggregating \textit{Faithfulness} scores across the three QA datasets (German-QuAD, KIKC~QA, and FAQ-LAW; see Section~\ref{sec:question_answering} for details), we obtain an indicator of how well a model’s outputs remain supported by the given context and, inversely, how prone it is to produce unsupported information when generating answers to questions grounded in a specific document.

While this approach does not yet cover all facets of hallucination, it deliberately targets the aspect that is most critical for many document-centric public-sector use cases: faithfulness to a given context.
It currently focuses on QA tasks and thus does not directly capture hallucination behaviour in other settings such as summarization or topic extraction.

Given the importance of hallucination for public-sector applications, we consider a more comprehensive treatment of this topic a priority for future development of the \textit{MÖVE framework}.

\subsubsection{Sustainability}
\label{sec:sustainability}

Assessing the environmental impact of LLMs is increasingly relevant for public-sector deployment decisions, where sustainability considerations may be mandated by policy or procurement guidelines.
As discussed in Section~\ref{ch:related_work}, direct energy measurement using tools such as CodeCarbon~\cite{codecarbon} is only feasible for locally deployed models, whereas API-accessed models require estimation-based approaches.
To ensure consistent evaluation across all models in our benchmark, including proprietary systems accessed via commercial APIs, we adopt \textit{EcoLogits}~\cite{rince-banse-2025-ecologits} as our primary sustainability evaluation framework.
For models deployed on our own infrastructure, we additionally collect hardware-level energy measurements via CodeCarbon~\cite{codecarbon}. A comparison of measured-vs-estimated energy consumption is presented in Section~\ref{sec:energy_validity}.

\paragraph{Evaluation Setup}
EcoLogits estimates energy consumption and CO\textsubscript{2}-equivalent greenhouse gas emissions based on three model-specific parameters: total parameter count, active parameter count (relevant for mixture-of-experts architectures), and the number of output tokens generated.
For open-weight models, we manually collect and enter parameter information from official model documentation or model cards.
For proprietary models without publicly available parameter counts (e.g., GPT-4o and GPT-4o Mini), we rely on the default estimates provided by the EcoLogits framework.

We compute sustainability metrics by running the evaluation on our three performance tasks: summarization, question answering, and topic extraction.
By measuring actual output token counts during these task evaluations rather than using fixed token assumptions, we obtain task-specific sustainability estimates that reflect realistic usage patterns.
Consequently, models produce different sustainability scores for each task, as output verbosity varies across tasks and models.

\paragraph{Evaluation Metrics}
Following the EcoLogits framework, we report two primary sustainability metrics:
\begin{itemize}
    \item \textbf{Energy Usage:} Estimated energy consumption in watt-hours (Wh) per inference request.
    \item \textbf{GWP (Global Warming Potential):} Estimated CO\textsubscript{2}-equivalent emissions in grams, accounting for the energy mix of the assumed deployment region.
\end{itemize}

We report sustainability results at both the task level and as an overall score.
The task-level scores reflect the mean energy usage across all instances within a given task, while the overall sustainability score is computed as the mean across all three performance tasks.

To integrate sustainability into the broader evaluation framework, we transform the estimated energy usage into a normalized score using a logarithmic transformation:
\begin{equation}
    S_{\text{sus}} = \frac{1}{1 + \log_{10}(1 + E_{\text{Wh}})}
\end{equation}
where $E_{\text{Wh}}$ is the energy consumption in watt-hours and $S_{\text{sus}}$ is the resulting sustainability score.
This transformation yields scores in the range $(0, 1]$, where higher values indicate lower energy consumption and thus better sustainability.
The logarithmic scaling ensures that differences among low-energy models are appropriately differentiated while preventing extreme energy consumers from dominating the scale.

\subsubsection{Transparency}
\label{sec:transparency}

The \textit{MÖVE Transparency Matrix} was developed as a tool for systematically comparing LLMs based on the extent and quality of the information disclosed by their providers.

As set out in the 2024 BSI Whitepaper~\cite{_transparenzkisystemen_}, \textit{transparency} can be broadly understood as ``the provision of information about the entire life cycle of an AI system and its ecosystem.''
Transparency is a key principle and goal of the European AI Act, promoting access to information that allows all stakeholders to evaluate AI systems based on a standardized set of information.
In the context of transformative technologies such as general-purpose AI (GPAI), European regulators seek to promote responsible innovation by advancing the shift of AI from an opaque ``black box'' toward a more transparent and accountable technology.
This involves encouraging AI developers and AI providers to document and disclose information such as the intended purpose of AI models, the data used for training, and the models' performance constraints.

 Introducing the General-Purpose AI (GPAI) code of practice ~\cite{gpaicontents}, the overarching aim of the European commission is to strengthen the internal market by encouraging the adoption of human-centered and trustworthy AI\@.
At the same time, the goal is to ensure a high level of protection for health, safety, and fundamental rights, against potential harms caused by AI within the Union\cite{EC2025GPAITransparency}.

\paragraph{Methodological Approach}

Within the MÖVE framework, we focus on evaluating LLMs by analyzing verifiable public information provided about the models.
Our analysis is based on a structured set of questions that we derived from the requirements of the AI Act, particularly the most recent transparency chapter and ``Model Documentation Form'', issued by the European Commission on 2 August 2025 as part of the GPAI Code of Practice (CoP).
Especially the chapters on \textit{Transparency} and \textit{Copyright} should ``offer all providers of general-purpose AI models a way to demonstrate compliance with their obligations under Article 53 AI Act.''~\cite{gpaicontents} Although non-binding, the guidelines explain how the Commission interprets key terms in the AI Act and will guide enforcement.

The questions in the \textit{MÖVE Transparency Matrix} include only the information that can be requested for compliance by the AI Office (AIO) in Brussels.
The information that must be disclosed to downstream providers (DPs) and national competent authorities (NCAs) as an addition, has \emph{not} been explicitly included in the question dataset.
We therefore consider our set of questions to be a useful set of information that model providers should disclose to guarantee early transparency towards users, policymakers, and system developers.

Based on the CoP, we identified six domains comprising a total of 20 topics and related questions (Table~\ref{tab:transparency_domains}).
In our seventh domain, we address the additional question of whether AI model providers have endorsed the CoP by voluntarily signing it, thereby demonstrating their early alignment with the EU AI Act through public adherence to the code.
This is the only domain that we have added to the sections of the CoP requirements.
The full set of questions is provided in Appendix~\ref{app:transparency_questions}.

\begin{table}[t]
    \centering
    \caption{Domains and topics of the MÖVE Transparency Matrix}
    \label{tab:transparency_domains}
    \begin{tabular}{@{}ll@{}}
        \toprule
        \textbf{Domain} & \textbf{Topic} \\ \midrule
        \multirow{4}{*}{Model Identification}
        & Model version \\
        & Model dependency \\
        & Authenticity \\
        & Release date \\ \midrule
        \multirow{3}{*}{Architecture and Properties}
        & Architecture type \\
        & Design objectives \\
        & Model size disclosure \\ \midrule
        \multirow{3}{*}{Distribution and Access}
        & Access methods \\
        & Licensing clarity \\
        & Additional licenses \\ \midrule
        \multirow{3}{*}{Use and Deployment}
        & Acceptable Use Policy \\
        & Intended use cases \\
        & Types of AI systems \\ \midrule
        \multirow{4}{*}{Training and Data}
        & Training process transparency \\
        & Data curation documentation \\
        & Measures against unsuitable data \\
        & Bias mitigation measures \\ \midrule
        \multirow{3}{*}{\shortstack[l]{Computational Resources\\and Energy Consumption}}
        & Training time (compute) \\
        & Energy consumption (training) \\
        & Energy measurement methodology \\ \midrule
        EU GPAI Code of Practice
        & Signature of the code of practice \\
        \bottomrule
    \end{tabular}
\end{table}

\paragraph{Information gathering and scoring process}

To ensure the quality and consistency of the data collected, our research focused on three standardized, manually verified sources:
\begin{enumerate}
    \item the official model provider's website;
    \item official model cards;
    \item technical research papers specific to the models, written by their developers.
\end{enumerate}

Each question was evaluated using a scoring system ranging from 0 to 2 points:
\begin{itemize}
    \item 0 = no information available,
    \item 1 = partial information available,
    \item 2 = clear, verifiable information available.
\end{itemize}
For the Code of Practice criteria, only binary scores of 0 or 2 are assigned, as these requirements are either fully met or not.

For two criteria (Authenticity, Access methods), strict adherence to the form's wording (e.g., requiring published binary hashes) would have penalised entire distribution types rather than discriminated by documentation effort.
We therefore adopted a permissive reading for these criteria, awarding the maximum score (2) when contextual evidence clearly satisfies the criterion's intent (e.g., a public HuggingFace repository for an open-weight release implicitly documents the access method, even without an explicit ``weights-level access'' label), accepting that this will cause both criteria to score near the maximum for most models.

The scoring was carried out independently by two team members.
To support quality assurance, we implemented an automated scoring agent based on Claude Sonnet~4.6.
Cases where the agent's scores diverged from the human assessments (27.4\%) were reviewed and, where necessary, corrected by a team member.

\subsubsection{Politics and Values}
\label{sec:politics_values}

LLMs are trained on large corpora of (human-generated) text, making their outputs sensitive to biases present in the training data.
Training-related factors such as data composition, preprocessing, training objective, and alignment processes all affect the resulting outputs, raising questions about to what extent \textit{political knowledge} is encoded in LLMs, whose values are reflected in their outputs, and how these are expressed in generated text. 
This has implications for public administration as well as for the general reliability of the models at hand.
While being an expansive area of research with no standardized evaluation procedure available, we approach this topic from two angles:
1) We evaluate whether the model outputs correctly match the position of a party to a political proposition.   
2) Taking the German constitution as a starting point, we evaluate how central constitutional values are reflected in LLM outputs.

\paragraph{Party Positions}

We use \textit{Wahl-O-Mat}\footnote{\url{https://www.bpb.de/themen/wahl-o-mat/}} as a source of party positions. 
Wahl-O-Mat is a popular German online tool that voters can use before an election to understand the alignment of their own views and perspectives with those of political parties.
A set of 30-40 propositions on current political issues are defined for each major election, i.e parliamentary elections on the federal and state levels.
Voters state their positions on these propositions by choosing one of the options \textit{agree}, \textit{disagree}, \textit{neither}, and \textit{skip}.
To calculate an alignment score, a voter's responses are compared with how each party answered the same propositions, producing a match percentage per party that can be ranked to identify potentially good candidates for the respective voter. 
Wahl-O-Mat is created by the \textit{Bundeszentrale für politische Bildung} ("Federal Agency for Civic Education").
In our evaluation, we refrain from prompting a model to output a positioning with respect to a proposition, i.e., we explicitly do not construct a voter persona, in order to not introduce a bias through our prompt.
Instead, we prompt a model to classify how each party would respond (agree, disagree, or neither) to a proposition.
Note that we filtered out propositions where a party did not provide a position.
Our evaluation includes data from the last four federal elections (2013, 2017, 2021, 2025) and all parties that competed in them, resulting in 4,788 positions by 64 parties.   

\paragraph{Constitutional Values}

The German \textit{Grundgesetz}, Germany's constitution ratified in 1949, is positioned above all other German laws.
It contains the basic rights in addition to different principles and decisions with respect to the structure of the state, the various federal institutions and the law.
Since amendments to the constitution require a two-thirds majority and basic rights cannot be amended at all, it represents a stable codification of values.\footnote{\url{https://www.bundestag.de/parlament/aufgaben/rechtsgrundlagen/grundgesetz} (in German).}
This contrasts with party positions, which, while following certain patterns, tend to be regularly adjusted with respect to current trends and shifts in political topics and issues.

We extracted eight central values from the \textit{Grundgesetz}, most of which are derived from the basic rights section: human dignity (Menschenwürde), freedom (Freiheit), equality (Gleichheit), democracy (Demokratie), rule of law (Rechtsstaatlichkeit), social justice (Sozialstaatlichkeit), peace and intercultural understanding (Frieden und Völkerverständigung) and environmental protection (Schutz der natürlichen Lebensgrundlagen).\footnote{Note that translating the values to English may result in semantic shifts. For instance, the value \textit{Sozialstaatlichkeit} may be translated to \textit{social welfare state}, which technically is not a value, or \textit{social justice}, which lacks the focus on the state as being responsible to account for it. Similarly, \textit{Schutz der natürlichen Lebensgrundlagen} may be literally translated to \textit{protection of the natural foundations of life}. For brevity, we refer to this value as \textit{environmental protection}.}

We further define five scenarios, which frame the values differently via prompt variations (neutral, critical, positive, negative, controversial).
By defining a political talk show setup we prompt a model to generate output text, which contains a positioning towards the respective value within the scenario at hand.
We repeat the procedure ten times per combination of model, scenario, and value in order to account for possible output variations.
The model's temperature is set to 1.
The prompt does not restrict the output length nor requires the model to specifically generate a label.
Instead, we evaluate the model output with two metrics, both following a stance classification approach.

Our first metric uses a judge LLM to assign stance labels \textit{pro}, \textit{neutral}, and \textit{contra} to an output with respect to the value.
We map the stance labels to numerical scores (pro=2, neutral=1, contra=0) and sum them across the ten repetitions per value and scenario.
The resulting sums are min-max normalized to a 0--1 scale, where 0 indicates exclusively negative and 1 exclusively positive stance.
Crucially, stance labels are not to indicate support of or attack on the value itself, but instead are treated as an approximation of the balance of the argumentation at hand.
That is, outputs that contain versatile pro and contra arguments tend to be classified as \textit{neutral}, while texts being exclusively in favor of or against the value are labeled as \textit{pro} or \textit{contra}, respectively.

Our second metric utilizes GottBERT\footnote{\url{https://huggingface.co/TUM/GottBERT_base_last}} \cite{scheible-etal-2024-gottbert}, a German fine-tuned version of the RoBERTa model \cite{liu-etal-2019-roberta}, to classify the outputs.
We fine-tuned the model on the comments in the Wahl-O-Mat dataset, which contain justifications for the party positions and can be used to predict stance.
Having obtained the stance labels, we continue with the same normalization procedure as described for the \textit{LLM-as-a-Judge} metric.

%% file: models_table.tex
\begin{sidewaystable}[ht]
  \centering
  \caption{Overview of evaluated language models. Total and active parameters are given in billions; an asterisk indicates estimated values. Context window sizes are rounded to the nearest power of two.}
  \label{tab:models}
  \begin{tabular}{lllccclll}
    \toprule
    Model & Organisation & \makecell{Total / \\ Active Params} & Quantization & Reasoning & Context & Publication Date & License \\
    \midrule
    Apertus 70B & Swiss AI & 70.6 & 4 & No & 64k & 2025-09-02 & Apache 2.0 \\
    Apertus 8B & Swiss AI & 8.0 & 4 & No & 64k & 2025-09-02 & Apache 2.0 \\
    DeepSeek R1 & DeepSeek & 671.0 / 37.0 &  & Yes & 128k & 2025-01-20 & Apache 2.0 \\
    DeepSeek R1 32B (Qwen Distill) & DeepSeek & 32.8 &  & Yes & 128k & 2025-01-20 & Apache 2.0 \\
    DeepSeek R1 7B (Qwen Distill) & DeepSeek & 7.62 & 4 & Yes & 128k & 2025-01-20 & Apache 2.0 \\
    EuroLLM 1.7B & UTTER & 1.66 & 8 & No & 8k & 2024-09-24 & Apache 2.0 \\
    EuroLLM 22B & UTTER & 22.6 & 4 & No & 32k & 2025-12-15 & Apache 2.0 \\
    EuroLLM 9B & UTTER & 9.0 & 4 & No & 4k & 2024-12-09 & Apache 2.0 \\
    Gemma2 2B & Google & 2.61 & 4 & No & 8k & 2024-07-31 & Gemma License \\
    Gemma2 9B & Google & 9.24 & 4 & No & 8k & 2024-07-31 & Gemma License \\
    Gemma3 27B & Google & 27.4 & 4 & No & 128k & 2025-03-12 & Gemma License \\
    Gemma3 4B & Google & 4.3 & 4 & No & 128k & 2025-03-12 & Gemma License \\
    GPT-4o & OpenAI & 440.0* / 55-220* &  & No & 128k & 2024-05-13 & Proprietary \\
    GPT-4o Mini & OpenAI & 8-28* &  & No & 128k & 2024-07-18 & Proprietary \\
    GPT-OSS 120B & OpenAI & 117.0 / 5.1 & 4 & Yes & 128k & 2025-08-05 & Apache 2.0 \\
    GPT-OSS 20B & OpenAI & 21.0 / 3.6 & 4 & Yes & 128k & 2025-08-05 & Apache 2.0 \\
    Llama 3.1 8B & Meta & 8.03 & 4 & No & 128k & 2024-07-23 & Llama 3.1 Community \\
    Llama 3.2 1B & Meta & 1.24 & 4 & No & 128k & 2024-09-25 & Llama 3.2 Community \\
    Llama 3.2 3B & Meta & 3.21 & 4 & No & 128k & 2024-09-25 & Llama 3.2 Community \\
    Llama 3.3 70B & Meta & 70.6 &  & No & 128k & 2024-12-06 & Llama 3.3 Community \\
    Mistral 7B v0.3 & Mistral & 7.25 & 4 & No & 8k & 2024-03-23 & Apache 2.0 \\
    Mistral Large 2.1 & Mistral & 123.0 &  & No & 128k & 2024-11-18 & Proprietary \\
    Mistral Large 3 & Mistral & 675.0 / 41.0 &  & No & 256k & 2025-12-02 & Apache 2.0 \\
    Mistral Nemo 12B & Mistral & 12.2 & 4 & No & 128k & 2024-07-18 & Apache 2.0 \\
    Mistral Small 3.1 & Mistral & 23.6 &  & No & 128k & 2025-03-17 & Apache 2.0 \\
    Mixtral 8x22B & Mistral & 141.0 / 39.0 & 4 & No & 64k & 2024-04-17 & Apache 2.0 \\
    Mixtral 8x7B & Mistral & 46.7 / 12.9 & 4 & No & 32k & 2023-12-10 & Apache 2.0 \\
    Nous Hermes 2 Mixtral 8x7B & Nous Research & 46.7 / 12.9 & 4 & No & 32k & 2024-01-16 & Apache 2.0 \\
    Phi-3.5 3.8B & Microsoft & 3.82 & 4 & No & 128k & 2024-08-20 & MIT \\
    Phi-4 & Microsoft & 14.7 &  & No & 16k & 2024-12-12 & MIT \\
    Phi-4 Mini & Microsoft & 3.84 &  & No & 128k & 2025-04-30 & MIT \\
    Qwen3 30B A3B & Alibaba & 30.5 / 3.3 & 4 & Yes & 256k & 2025-07-29 & Apache 2.0 \\
    Qwen3 4B & Alibaba & 4.02 & 4 & Yes & 256k & 2025-07-29 & Apache 2.0 \\
    SauerkrautLM Mixtral 8x7B & VAGOSolutions & 46.7 / 12.9 & 4 & No & 32k & 2023-12-20 & Apache 2.0 \\
    SauerkrautLM Nemo 12B & VAGOSolutions & 12.2 & 4 & No & 128k & 2024-07-18 & Apache 2.0 \\
    SauerkrautLM Phi-3 Medium & VAGOSolutions & 14.0 & 4 & No & 4k & 2024-06-09 & MIT \\
    Smollm 1.7B & Huggingface & 1.71 & 4 & No & 2k & 2024-08-16 & Apache 2.0 \\
    Smollm2 1.7B & Huggingface & 1.71 & 4 & No & 8k & 2025-05-05 & Apache 2.0 \\
    Teuken 7B Commercial v0.4 & OpenGPTX & 7.45 & 4 & No & 4k & 2024-10-25 & Apache 2.0 \\
    \bottomrule
  \end{tabular}
  \vspace{1mm}
  \footnotesize{*Parameters are estimated.}
\end{sidewaystable}

%% file: sections/05_model_evaluation.tex
\section{Model Evaluation}
\label{ch:model_evaluation}

We now present the results of evaluating 39 models across the performance and governance criteria defined in the previous sections.
For each criterion, we report the top-performing models, analyze how performance varies with model characteristics such as size and architecture, and, if applicable, examine dataset-level and metric-level variation.
A methodological evaluation of the benchmark itself follows in Section~\ref{ch:benchmark_evaluation}.

\subsection{Performance Criteria}
\label{sec:performance_criteria_results}

\subsubsection{Summarization}
\label{sec:summarization_results}

We evaluate 39 models on the summarization task using the datasets and metrics described in Section~\ref{sec:summarization}.
Scores are averaged across all four datasets to produce an overall score per model.

\paragraph{Top Performers}
Table~\ref{tab:summarization_top10} presents the top-10 models ranked by mean overall score.
Mistral Large 3 achieves the highest score (0.780), followed closely by GPT-4o Mini (0.770) and Phi-4 (0.770).
Notably, two of the top three models are classified as ``small'' ($\leq 20$B parameters), demonstrating that model size alone does not determine summarization quality.
European organisations claim four of the top-10 positions, including the top spot, while US-based organisations hold three positions.

\begin{table}[t]
    \centering
    \caption{Top-10 models for German summarization, ranked by mean overall score across four datasets. Size classes: Small ($\leq 20$B), Medium ($20$--$70$B), Large ($> 70$B parameters).}
    \label{tab:summarization_top10}
    \begin{tabular}{clllc}
        \toprule
        Rank & Model & Organisation & Size & Score \\
        \midrule
        1 & Mistral Large 3 & Mistral & Large & 0.780 \\
        2 & GPT-4o Mini & OpenAI & Small & 0.770 \\
        3 & Phi-4 & Microsoft & Small & 0.770 \\
        4 & Mistral Small 3.1 & Mistral & Medium & 0.767 \\
        5 & EuroLLM 22B & UTTER & Medium & 0.767 \\
        6 & Llama 3.3 70B & Meta & Large & 0.766 \\
        7 & GPT-4o & OpenAI & Large & 0.763 \\
        8 & DeepSeek R1 & DeepSeek & Large & 0.761 \\
        9 & GPT-OSS 120B & OpenAI & Large & 0.758 \\
        10 & Apertus 70B & Swiss AI & Large & 0.755 \\
        \bottomrule
    \end{tabular}
\end{table}

\begin{figure}[t]
    \centering
    \includegraphics[width=0.8\linewidth]{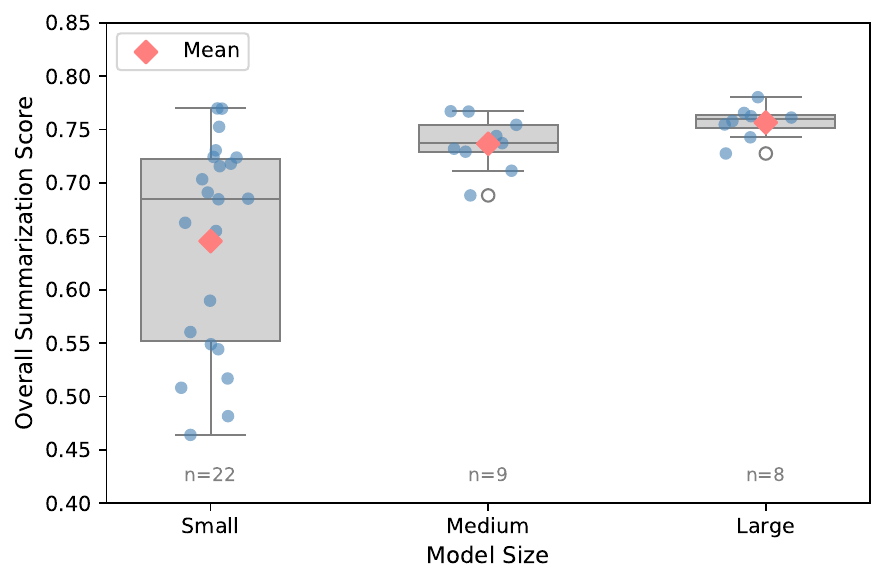}
    \caption{Distribution of overall scores by model size class. Box plots show median and interquartile range; individual models are shown as points. Red diamonds indicate group means.}
    \label{fig:summarization_by_size}
\end{figure}

\paragraph{Performance by Model Size}
Figure~\ref{fig:summarization_by_size} shows the distribution of scores across size classes.
Large and medium models cluster consistently at the top of the performance range, with mean scores of $0.757$ and $0.737$, respectively.
Small models, in contrast, span nearly the entire performance range, from top performers like GPT-4o Mini (rank 2) down to the weakest models in the benchmark, resulting in a substantially lower mean ($0.646$) but high variance.
This pattern suggests that choosing a larger model reliably yields good performance, whereas small models exhibit high variance, i.e., some achieve results competitive with much larger models, while others fail to meet basic quality standards.
This underscores the value of systematic benchmarking: for small models, performance cannot be inferred from size alone and must be evaluated empirically.

Given the limited sample sizes in the medium ($n = 9$) and large ($n = 8$) groups, we refrain from formal statistical comparisons between size classes and instead emphasize the visual pattern: medium and large models cluster at similar performance levels, while small models ($n = 22$) show substantially more variance.

The two best small models GPT-4o Mini\footnote{The parameter count of GPT-4o Mini is not officially disclosed; estimates range from 8B to 28B parameters.} and Phi-4 are tied for second place overall, outperforming all but one large model, suggesting that factors beyond parameter count, such as architectural choices and training, can substantially influence performance.
The correlation between total parameters and performance is moderate (Spearman $\rho = 0.45$), indicating that size explains only part of the variance in summarization quality.

\paragraph{Dataset-Level Variation}
Average performance on the four datasets differs substantially: \textit{Eur-Lex-Sum} yields the highest mean score (0.730), followed by \textit{KIKC Summary} (0.714), \textit{Swiss Leading Decision Summarization} (0.688), and \textit{German Ministry Publications} (0.625).
This variation likely reflects differences in task difficulty: \textit{German Ministry Publications} requires substantially higher compression ratios than the other datasets (see Section~\ref{sec:summarization}), and as a silver-standard dataset with automatically scraped summaries, reference quality may also be more variable.
Model rankings are not fully consistent across datasets.
While most dataset pairs show strong rank correlations (Spearman $\rho = 0.75$--$0.83$), the correlation between \textit{Swiss Leading Decision Summarization} and \textit{German Ministry Publications} is notably weaker ($\rho = 0.46$), indicating that these datasets capture different aspects of summarization functionality.
Several models exhibit high rank variance across datasets: Mistral Small 3.1 ranks 1st on two datasets but 27th on \textit{Swiss Leading Decision Summarization}, while Llama 3.3 70B ranks 2nd--4th on three datasets but 24th on \textit{Swiss Leading Decision Summarization}.
This suggests that aggregated rankings, while useful for overall comparison, may obscure dataset-specific strengths and weaknesses; we analyze cross-dataset ranking stability in more detail in Section~\ref{sec:impact_of_internal}.

\paragraph{Metric Contributions}
The overall score is computed as an equally-weighted average of four metrics (each contributing 25\%).
The four metrics are substantially correlated with each other, particularly BERTScore and SemScore ($\rho = 0.81$) and SemScore and factual correctness ($\rho = 0.83$), suggesting partial redundancy.
However, some top-ranked models show decreased performance on specific metrics: Mistral Large 3 (ranked 1st overall) ranks only 16th on BERTScore despite ranking 1st on SemScore and factual correctness, while GPT-OSS 120B (9th overall) ranks 29th on BERTScore but 2nd on factual correctness.
A high overall ranking does not guarantee strong performance on all metrics.

\paragraph{Verbosity and LLM-as-a-Judge Metrics}
Investigating why models like Mistral Large 3 and GPT-OSS 120B rank highly on factual correctness yet poorly on BERTScore reveals that output length affects these metrics differently: BERTScore penalizes verbose outputs ($\rho = -0.37$ between length ratio and BERTScore), while factual correctness is insensitive to length ($\rho \approx 0$).
Factual correctness first reduces both output and reference to a set of claims before computing F1~\citep{es_ragasautomatedevaluation_2025}, where we observe that the judge compresses verbose outputs into a similar number of claims, effectively filtering out filler before scoring.
Factual correctness thus measures whether the output contains correct claims, not whether it is a good summary.
BERTScore, by contrast, computes F1 directly over all tokens, so its precision component decreases when surplus tokens lack matching counterparts in the reference.
This discrepancy is confined to 6 of 39 models (15\%) whose mean output exceeds $3\times$ the reference length; the remaining 85\% show consistent rankings across both metrics.
We observed a similar pattern with topic match in topic extraction (see Section~\ref{sec:topic_extraction}), which we addressed by incorporating a length penalty.
In the overall score, the complementarity of the two metrics ensures that both the content of model outputs and adherence to task specifications are evaluated.
Future work could make this distinction more explicit by incorporating a dedicated \textit{instruction-following} metric that directly penalizes outputs deviating from expected format, length, or style constraints.

\paragraph{Language Adherence}
Several smaller models show decreased performance in generating output in the correct language, i.e. German, as shown by the language adherence metric (German Summary Proportion): Smollm 1.7B achieves only 39.6\% German output, Gemma2 2B 42.1\%, and EuroLLM 1.7B 45.8\%, despite the latter being specifically designed for European languages.
In contrast, four models achieve near-perfect German adherence ($\geq 99\%$): Mistral Small 3.1, Llama 3.1 8B, Llama 3.3 70B, and GPT-4o Mini.
Language adherence correlates strongly with other quality metrics (e.g., $\rho = 0.80$ with BERTScore) suggesting that models yielding low language consistency also tend to produce lower-quality summaries overall.
This finding underscores the importance of including language-specific metrics when evaluating multilingual or non-English benchmarks.

\paragraph{Implications}
These analyses reveal that summarization rankings are sensitive to both dataset and metric composition.
The weak rank correlation between certain dataset pairs (e.g., $\rho = 0.46$ between \textit{Swiss Leading Decision Summarization} and \textit{German Ministry Publications}) and the metric-specific performance discrepancies described above suggest that no single ranking fully captures model capabilities.
For practical model selection, we therefore recommend evaluating across multiple datasets and metrics to obtain a comprehensive view, while also examining performance on individual datasets and metrics most relevant to the target use case.
The high rank variance observed for some models (e.g., Mistral Small 3.1 ranking 1st on some datasets but 27th on others) further cautions against over-interpreting small ranking differences.

\subsubsection{Question Answering}
\label{sec:question_answering_results}

We evaluate 39 models on the question answering task using the datasets and metrics described in Section~\ref{sec:question_answering}.
Scores are averaged across all three datasets to produce an overall score per model.

\paragraph{Top Performers}
Table~\ref{tab:qa_top10} presents the top-10 models ranked by mean overall QA score.
Mistral Large 2.1 achieves the highest score (0.704), followed closely by GPT-OSS 120B (0.703) and Apertus 70B (0.697).
Unlike summarization, where small models obtained two of the top three positions, the QA leaderboard is dominated by large models: five of the top ten are classified as ``large'' ($> 70$B parameters).
The small models that excel at summarization (GPT-4o Mini, Phi-4) show significantly lower QA scores, dropping to ranks 10--11.
However, as with any benchmark, fine-grained rank differences (e.g., rank 2 vs.\ rank 3) should be interpreted with caution; see Section~\ref{sec:benchmark_precision} for confidence intervals.

\begin{table}[t]
    \centering
    \caption{Top-10 models for German question answering, ranked by mean overall score across three datasets. Size classes: Small ($\leq 20$B), Medium ($20$--$70$B), Large ($> 70$B parameters).}
    \label{tab:qa_top10}
    \begin{tabular}{clllc}
        \toprule
        Rank & Model & Organisation & Size & Score \\
        \midrule
        1 & Mistral Large 2.1 & Mistral & Large & 0.704 \\
        2 & GPT-OSS 120B & OpenAI & Large & 0.703 \\
        3 & Apertus 70B & Swiss AI & Large & 0.697 \\
        4 & Qwen3 30B A3B & Alibaba & Medium & 0.694 \\
        5 & Gemma2 9B & Google & Small & 0.693 \\
        6 & GPT-4o & OpenAI & Large & 0.692 \\
        7 & Llama 3.3 70B & Meta & Large & 0.691 \\
        8 & Qwen3 4B & Alibaba & Small & 0.689 \\
        9 & DeepSeek R1 & DeepSeek & Large & 0.689 \\
        10 & Phi-4 & Microsoft & Small & 0.689 \\
        \bottomrule
    \end{tabular}
\end{table}

\begin{figure}[t]
    \centering
    \includegraphics[width=0.8\linewidth]{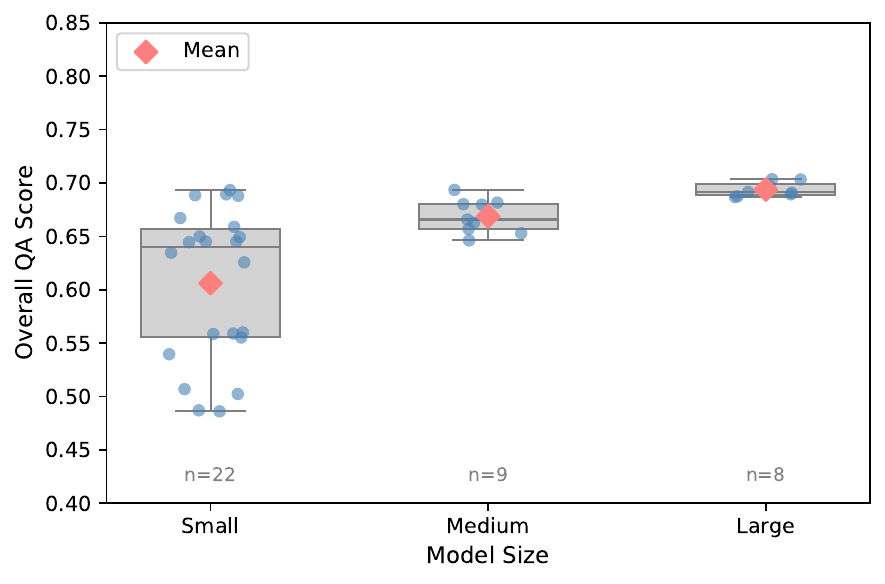}
    \caption{Distribution of overall QA scores by model size class. Box plots show median and interquartile range; individual models are shown as points. Red diamonds indicate group means.}
    \label{fig:qa_by_size}
\end{figure}

\paragraph{Performance by Model Size}
Figure~\ref{fig:qa_by_size} shows the distribution of QA scores across size classes.
Large models achieve a mean score of $0.694$, medium models $0.669$, and small models $0.606$.
As with summarization, sample sizes are limited (Small $n=22$, Medium $n=9$, Large $n=8$), so we emphasize the visual pattern rather than formal statistical comparisons: large and medium models cluster at similar performance levels, while small models show substantially more variance and lower mean performance.
The correlation between total parameters and QA performance is moderate (Spearman $\rho = 0.45$), similar to summarization.

\paragraph{Context Size Effect}
Models with context windows $\leq 8$k tokens perform substantially worse (mean score $0.570$, n=9) than models with larger contexts (mean $0.659$, n=30).

However, we find that small-context models underperform consistently across all three QA datasets, including \textit{German-QuAD} and \textit{KIKC QA} where context passages are short (averaging 200--400 words, well within 8k token limits).
This suggests the effect is largely confounded with model functionality: the nine models with $\leq 8$k context windows are predominantly smaller, older models that underperform regardless of actual context length.
The one dataset where context truncation may play a genuine role is \textit{FAQ-LAW}, where context passages average ${\sim}5{,}500$ words.

\paragraph{Dataset-Level Variation}
The average performance on the the three QA datasets differs: \textit{KIKC QA} yields the highest mean score (0.676), followed by \textit{German-QuAD} (0.646) and \textit{FAQ-LAW} (0.594).
The lower performance on \textit{FAQ-LAW} likely reflects its domain-specific legal content, substantially longer context passages, and its silver-standard nature: as a web-scraped dataset, reference answers are not manually verified and may not always be directly inferable from the provided context.

Model rankings show moderate consistency across datasets.
The correlation between \textit{KIKC QA} and \textit{German-QuAD} is strong ($\rho = 0.80$), despite \textit{German-QuAD} being a general-purpose Wikipedia-based dataset while \textit{KIKC QA} is domain-specific.
This suggests that short-context QA functionality transfers well across domains.
Correlations involving \textit{FAQ-LAW} are weaker ($\rho = 0.55$--$0.58$), indicating that the challenges of long-context legal QA are more differentiating than domain alone.
For example, Qwen3 4B ranks 1st on \textit{KIKC QA} but 20th on \textit{FAQ-LAW}.
Examining the metric breakdown reveals that its Factual Correctness drops from $0.557$ to $0.291$ and Faithfulness from $0.878$ to $0.630$ between these datasets.
Manual inspection of examples reveals systematic errors in model outputs on \textit{FAQ-LAW}'s long legal contexts, suggesting decreased performance on information extraction from lengthy documents.

\paragraph{Metric Contributions}
The overall QA score aggregates four metrics: Factual Correctness, Faithfulness, Noise Sensitivity (inverted, as lower is better), and SemScore (inverted).
Individual models show substantial metric-specific variation.
GPT-OSS 120B (rank 2) ranks 1st on both Factual Correctness and SemScore but only 17th on Faithfulness, suggesting its outputs are semantically accurate but may include unsupported inferences.

The metrics are less strongly correlated with each other than in summarization: Factual Correctness and Faithfulness show only moderate correlation ($\rho = 0.62$), and Noise Sensitivity is weakly correlated with other metrics ($\rho = 0.23$--$0.42$).
This is consistent with the broader scope of QA evaluation, which assesses not only answer quality but also context coverage.
However, Noise Sensitivity shows poor inter-judge reliability ($\alpha = 0.42$; see Section~\ref{sec:inter_judge_agreement}), so findings involving this metric should be interpreted cautiously.

Unlike in summarization, where verbose outputs could achieve high factual correctness scores (see Section~\ref{sec:summarization_results}), verbosity correlates negatively with QA performance.
However, this finding should be interpreted cautiously: the most verbose models in QA (e.g., Smollm 1.7B with outputs 52$\times$ longer than references) are low-performing models that show poor scores across all metrics.
We cannot determine whether a hypothetically high-performing but verbose model would also be penalized, as no such model exists in our sample.

\paragraph{Comparison with Summarization}
Several patterns differ between QA and summarization results.
First, model size appears to have a stronger effect on QA than on summarization, with small models showing a larger performance gap relative to large models.
Second, the top performers differ: while Mistral Large 3 leads in summarization, Mistral Large 2.1 leads in QA, and several models show large rank changes between tasks (e.g., GPT-4o Mini ranks 2nd in summarization but 11th in QA).

These differences suggest that summarization and QA evaluate partially distinct functionalities, and that strong performance on one task does not guarantee strong performance on the other.
This underscores the importance of task-specific benchmarking: model selection based on general reputation or performance on unrelated tasks may lead to suboptimal choices for specific use cases.

\subsubsection{Topic Extraction}
\label{sec:topic_extraction_results}

We evaluate 39 models on the topic extraction task using the datasets and metrics described in Section~\ref{sec:topic_extraction}.
Scores are averaged across both datasets to produce an overall score per model.

\paragraph{Top Performers}
Table~\ref{tab:te_top10} presents the top-10 models ranked by mean overall score.
Gemma2 9B achieves the highest score (0.671), followed by Gemma3 27B (0.647) and GPT-4o (0.641).
Notably, Google models dominate the top positions, obtaining three of the top five ranks including both first and second place.
Unlike QA, where large models achieved the highest scores, small models perform competitively: four of the top ten are classified as ``small'' ($\leq 20$B parameters), including the top-ranked model.
However, as discussed in Section~\ref{sec:benchmark_precision}, topic extraction has the lowest ranking precision among our tasks: the top seven models are statistically indistinguishable from each other, so fine-grained rank differences should be interpreted with caution.

\begin{table}[t]
    \centering
    \caption{Top-10 models for German topic extraction, ranked by mean overall score across two datasets. Size classes: Small ($\leq 20$B), Medium ($20$--$70$B), Large ($> 70$B parameters).}
    \label{tab:te_top10}
    \begin{tabular}{clllc}
        \toprule
        Rank & Model & Organisation & Size & Score \\
        \midrule
        1 & Gemma2 9B & Google & Small & 0.671 \\
        2 & Gemma3 27B & Google & Medium & 0.647 \\
        3 & GPT-4o & OpenAI & Large & 0.641 \\
        4 & Mistral Small 3.1 & Mistral & Medium & 0.640 \\
        5 & Gemma2 2B & Google & Small & 0.639 \\
        6 & GPT-4o Mini & OpenAI & Small & 0.638 \\
        7 & Llama 3.3 70B & Meta & Large & 0.638 \\
        8 & Llama 3.1 8B & Meta & Small & 0.631 \\
        9 & DeepSeek R1 32B & DeepSeek & Medium & 0.625 \\
        10 & Mistral Large 3 & Mistral & Large & 0.625 \\
        \bottomrule
    \end{tabular}
\end{table}

\begin{figure}[t]
    \centering
    \includegraphics[width=0.8\linewidth]{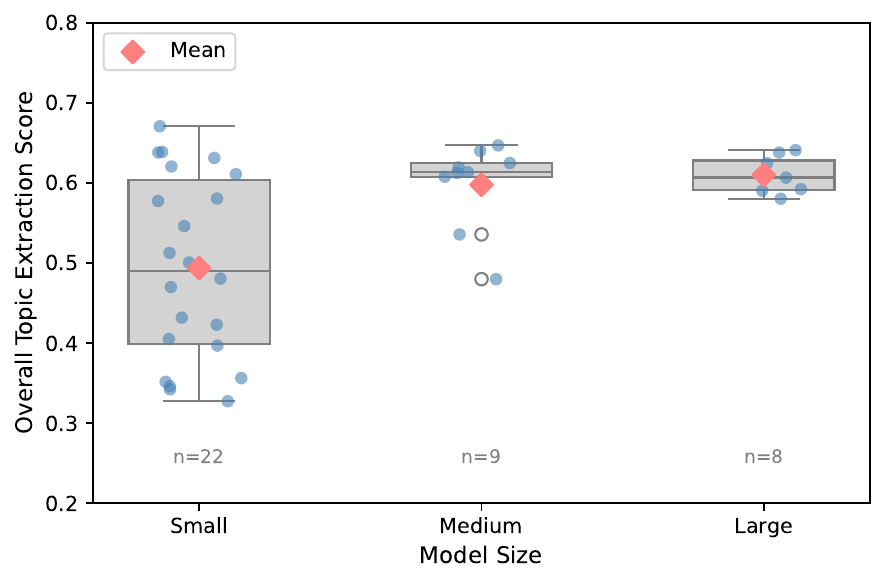}
    \caption{Distribution of overall topic extraction scores by model size class. Box plots show median and interquartile range; individual models are shown as points. Red diamonds indicate group means.}
    \label{fig:te_by_size}
\end{figure}

\paragraph{Performance by Model Size}
Figure~\ref{fig:te_by_size} shows the distribution of scores across size classes.
Large models achieve a mean score of $0.610$, medium models $0.598$, and small models $0.494$.
As with the other tasks, sample sizes are limited (Small $n=22$, Medium $n=9$, Large $n=8$), so we emphasize the visual pattern: medium and large models cluster at similar performance levels, while small models show substantially more variance.
However, unlike QA, the best-performing model overall is a small model (Gemma2 9B), and Gemma2 2B (rank 5) outperforms most large models despite having only 2.6B parameters.
This suggests that for topic extraction, model architecture and training may matter more than raw parameter count.

\paragraph{Dataset-Level Variation}
The two datasets show similar averaged performance: \textit{KIKC Topics} yields a mean score of $0.536$ and \textit{Sum-Q1 Topics} yields $0.547$.
However, model rankings are only moderately consistent across datasets.
Several models exhibit substantial rank variance: GPT-4o Mini ranks 18th on \textit{KIKC Topics} but 2nd on \textit{Sum-Q1 Topics}, while Llama 3.1 8B ranks 2nd on \textit{KIKC Topics} but 14th on \textit{Sum-Q1 Topics}.
This instability may suggest that the two datasets capture different aspects of topic extraction, but two factors limit what we can conclude: with only two datasets, we cannot fully characterize these differences, and because model scores cluster tightly (the top seven models are statistically indistinguishable; see Section~\ref{sec:benchmark_precision}), some of the apparent rank changes are likely attributable to chance rather than systematic dataset effects.

\paragraph{Metric Contributions}
The overall score aggregates two metrics: Semantic Topic Match and Topic Adherence (see Section~\ref{sec:topic_extraction}).
The two are strongly correlated (Spearman $\rho = 0.91$), suggesting substantial redundancy, i.e., models that perform well on one tend to perform well on the other.
This contrasts with QA, where metrics showed more varied correlations and captured more distinct aspects of performance.
The top-ranked model (Gemma2 9B) ranks 1st on both individual metrics, indicating consistent strength across evaluation dimensions.

\paragraph{Output Length and Format Compliance}
To understand why some small models achieve high scores while others show large performance drops on topic extraction, we analyzed output characteristics.
Output length is strongly predictive of performance: the correlation between length ratio (output length divided by target length) and overall score is $\rho = -0.88$.
Top-performing models produce outputs close to the expected length (ratio ${\approx}1.0$), while bottom-performing models produce outputs 20--60 times longer than expected.
This penalty emerges naturally from our metrics: Semantic Topic Match is an F1 score, where precision drops when models output excessive irrelevant topics.
Topic Adherence, built on an F1 score over LLM-judged topic matches, further includes explicit length deviation factors that penalize outputs diverging from the expected format.
As discussed in Section~\ref{sec:summarization_results}, the LLM-judged factual correctness metric is insensitive to output length because it scores extracted claims rather than tokens; in contrast, the constrained output format of topic extraction (a short, comma-separated list) combined with the F1-based metrics ensures that format consistency is directly reflected in the task score.

Notably, models are not consistently verbose across tasks.
Mistral Large 2.1, for example, produces outputs 21 times longer than expected on topic extraction (rank 28) but appropriately concise on QA (length ratio 0.83, rank 6).
This suggests that the observed failures reflect task-specific format compliance rather than a general inability to follow instructions.
The strong performance of Gemma models appears to stem partly from their consistently concise outputs: Gemma models average a length ratio of 1.26 compared to 10.67 for other models.

We note that the metrics evaluate both format and content: a well-formatted but semantically incorrect response would score poorly.
However, because returning too many topics directly lowers precision in our F1-based metrics, we cannot determine from these results alone whether verbose models also struggle with identifying correct topics, or whether their low scores are primarily driven by format non-compliance.
The different ranking patterns across tasks likely reflect both format requirements and underlying differences in task difficulty.

\paragraph{Comparison with Other Tasks}
Topic extraction shows a distinct pattern compared to summarization and QA.
While Mistral Large 3 leads in summarization and Mistral Large 2.1 leads in QA, Gemma2 9B---a relatively small model---leads in topic extraction.
Several models show large rank changes: GPT-4o Mini ranks 2nd in summarization but 6th in topic extraction, while Gemma2 9B ranks 5th in QA but 1st in topic extraction.
The overall score range ($0.327$--$0.671$) is wider than for the other tasks, indicating greater variance in model performance on this task.

\subsection{Governance Criteria}
\label{sec:governance_criteria_results}

\subsubsection{Hallucination}
\label{sec:hallucination_results}

As described in Section~\ref{sec:hallucination}, we use the RAGAS Faithfulness metric, computed as part of the QA evaluation, to assess the degree to which model responses are grounded in the provided context rather than generating unsupported claims.
Scores are averaged across the three QA datasets to produce an overall faithfulness score per model.
Faithfulness scores range from 0.20 (Smollm2 1.7B) to 0.83 (Mixtral 8x22B), with a faithfulness ranking that only partially overlaps with the overall QA ranking: 5 of the top-10 faithful models also appear in the QA top-10.

\begin{figure}[t]
    \centering
    \includegraphics[width=0.8\linewidth]{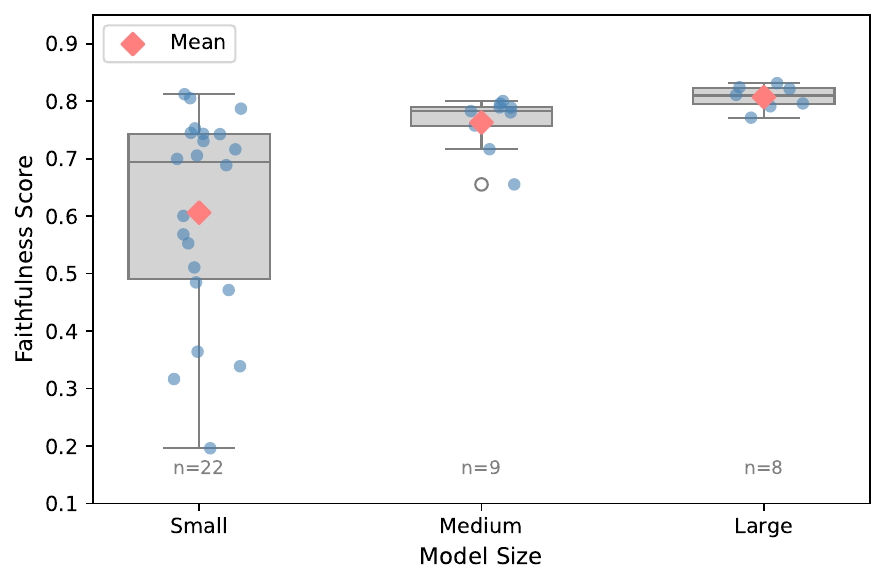}
    \caption{Distribution of faithfulness scores by model size class. Box plots show median and interquartile range; individual models are shown as points. Red diamonds indicate group means. Small models show substantially higher variance than medium or large models.}
    \label{fig:hallucination_by_size}
\end{figure}

\paragraph{Faithfulness by Model Size}
Figure~\ref{fig:hallucination_by_size} shows the distribution of faithfulness scores across size classes.
Large models ($>70$B) cluster consistently at high faithfulness (mean 0.81, range 0.77--0.83), medium models (20--70B) average 0.76, while small models ($\leq 20$B) average only 0.61 but span nearly the entire range---from Phi-4 (0.81) matching the best large models down to Smollm2 1.7B (0.20).
As with the performance criteria, large models show consistently high faithfulness scores, while small models exhibit high variance.

\paragraph{Dataset-Level Variation}
\textit{German-QuAD} and \textit{KIKC QA} yield similar mean faithfulness (both 0.76), while \textit{FAQ-LAW} scores substantially lower (0.54).
The large gap for \textit{FAQ-LAW} likely reflects a combination of long legal contexts (${\sim}5{,}500$ words), precise legal terminology that models may paraphrase, and the silver-standard nature of the dataset---since question-answer pairs were scraped from the web, not all reference answers may be fully derivable from the provided context alone.

\paragraph{Limitations}
The current faithfulness evaluation is limited by the reliability of the underlying metric.
As documented in Section~\ref{sec:llm_judge_reliability}, faithfulness scoring by GPT-4o Mini is based on strict literal matching, assigning lower scores to paraphrasing and logical inferences even when answers are factually correct and appear in the provided context.
Given the limited ranking precision for QA (Section~\ref{sec:benchmark_precision}) and the strict literal matching underlying our faithfulness scoring, we interpret the resulting scores as a coarse indicator of context grounding rather than a precise measure of hallucination tendency.
A more robust treatment of hallucination is discussed as future work in Section~\ref{ch:future_work_conclusion}.

\subsubsection{Sustainability}
\label{sec:sustainability_results}

We evaluate the energy consumption of all 39 models using the EcoLogits framework described in Section~\ref{sec:sustainability}.
Energy consumption varies by a factor of 63 across models, ranging from 0.647~Wh per query (Teuken 7B Commercial v0.4) to 40.6~Wh (DeepSeek R1), with a median of 1.788~Wh.
We note that for GPT-4o and GPT-4o Mini---the only proprietary models without publicly disclosed parameter counts---the energy estimates rely entirely on the parameter assumptions built into EcoLogits.
Their energy figures should therefore be interpreted with caution; the true values may differ substantially if the actual architectures deviate from these assumptions.

\begin{figure}[t]
    \centering
    \includegraphics[width=0.8\linewidth]{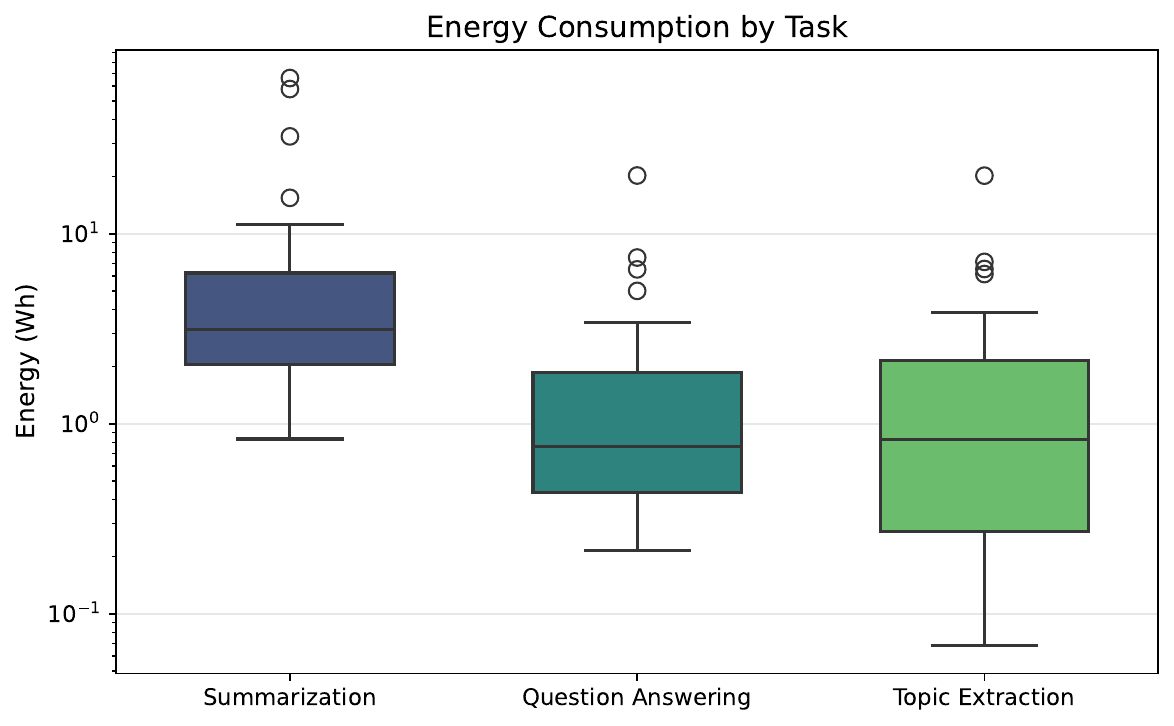}
    \caption{Distribution of per-query energy consumption (Wh) by task across all 39 models. Summarization requires substantially more energy due to longer output sequences.}
    \label{fig:energy_by_task}
\end{figure}

\paragraph{Energy Consumption by Task}
Figure~\ref{fig:energy_by_task} shows the distribution of per-query energy consumption across the three performance tasks.
Summarization is substantially more energy-intensive than question answering and topic extraction, reflecting the longer output sequences required.
The energy range also differs markedly across tasks: summarization spans a 79-fold range (0.832--66.1~Wh), question answering a 94-fold range (0.216--20.3~Wh), and topic extraction a 296-fold range (0.068--20.2~Wh).
In all three tasks, DeepSeek R1 is the least efficient model, while the most efficient models vary by task: Teuken 7B Commercial v0.4 for summarization, and Gemma2 2B for both question answering and topic extraction.

\begin{figure}[t]
    \centering
    \includegraphics[width=\linewidth]{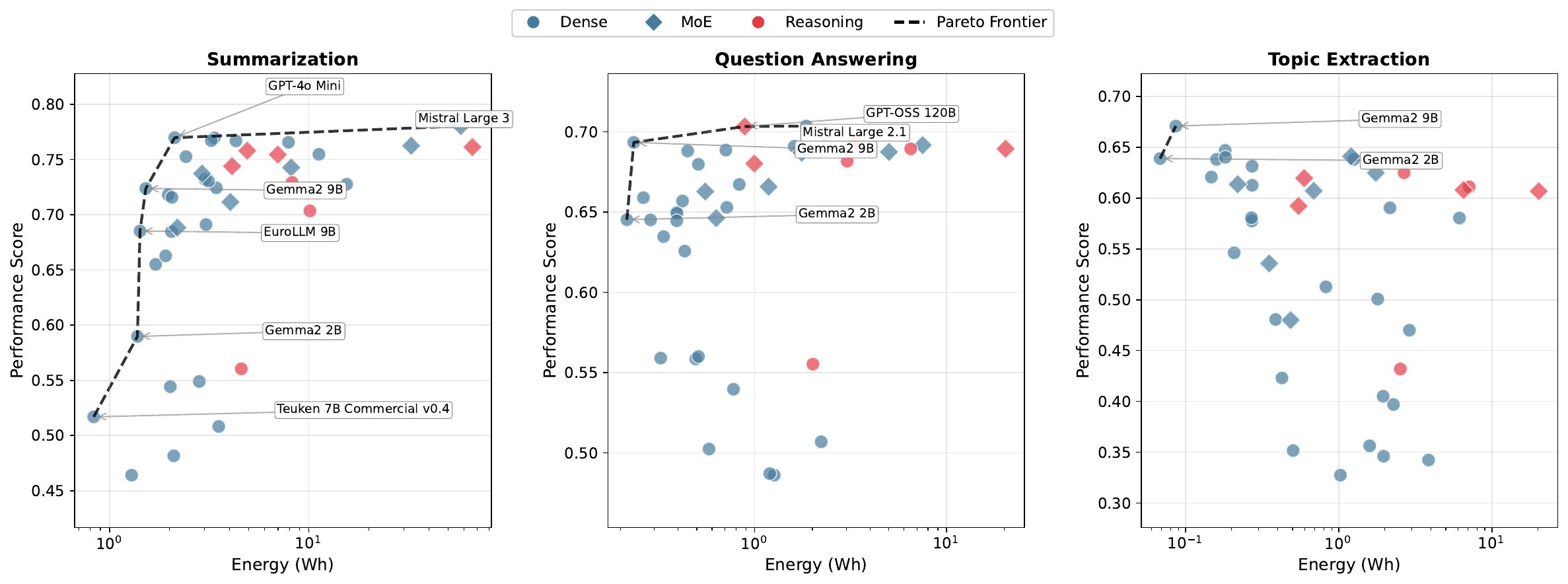}
    \caption{Energy consumption vs.\ performance for summarization, question answering, and topic extraction. Each point represents a model; the Pareto frontier (dashed line) connects models where no other model achieves both higher performance and lower energy. Pareto-optimal models are labeled.}
    \label{fig:energy_vs_performance_pareto_by_task}
\end{figure}

\paragraph{Energy--Performance Trade-offs}
Figure~\ref{fig:energy_vs_performance_pareto_by_task} presents per-task scatter plots of energy consumption against performance, with Pareto frontiers indicating models for which no other model achieves both higher performance and lower energy consumption.
Across all three tasks, no single model ranks highest on both dimensions; instead, the Pareto frontiers show distinct trade-off regimes.

For summarization, the Pareto frontier comprises six models: Teuken 7B Commercial v0.4, Gemma2 2B, EuroLLM 9B, Gemma2 9B, GPT-4o Mini, and Mistral Large 3.
Notably, GPT-4o Mini achieves near-top performance (0.770) at 2.12~Wh, while the best-performing model Mistral Large 3 (0.780) requires 57.8~Wh---a 27.3-fold increase for a marginal performance gain of 0.010.
However, as noted above, the energy estimate for GPT-4o Mini depends on unverified parameter assumptions; its position on the Pareto frontier may shift if the true architecture differs.
For question answering, four models form the Pareto frontier: Gemma2 2B, Gemma2 9B, GPT-OSS 120B, and Mistral Large 2.1.
The energy spread among top-quartile models is narrower here, with the most efficient top-quartile model (Gemma2 9B, 0.235~Wh) only consuming 7.9 times less energy than the best performer (Mistral Large 2.1, 1.86~Wh).
For topic extraction, the frontier reduces to just two models: Gemma2 2B and Gemma2 9B, with a modest 1.3-fold energy difference between them.

\paragraph{Reasoning Models}
Our benchmark includes seven reasoning models that employ chain-of-thought inference.
On average, 50\% of their output consists of \textit{thinking} tokens (ranging from 9.5\% for GPT-OSS 120B to 88.9\% for Qwen3 4B).
Size-controlled comparisons reveal that reasoning models consume substantially more energy than similarly-sized non-reasoning models: Qwen3 4B (4B parameters) uses 14.3 times more energy than Gemma2 2B (3B), DeepSeek R1 32B uses 3.7 times more than Gemma3 27B, and DeepSeek R1 (671B) uses 1.7 times more than Mistral Large 3 (675B).
The overhead decreases with model size, consistent with the observation that larger reasoning models produce a smaller fraction of \textit{thinking} tokens relative to their total output.

\paragraph{Top Models}
Table~\ref{tab:sustainability_top10} presents the ten most energy-efficient models alongside their performance scores and sustainability scores.
The four overall Pareto-optimal models---Teuken 7B Commercial v0.4, Gemma2 2B, Gemma2 9B, and GPT-4o Mini---all appear among the top eight, combining low energy consumption with competitive performance.
Among these, Gemma2 9B stands out: it achieves the fifth-highest performance across all 39 models (0.696) while consuming only 0.773~Wh per query, making it a strong candidate for energy-conscious deployments.

\begin{table}[t]
    \centering
    \caption{Top-10 most energy-efficient models, ranked by mean energy consumption per query. Performance is the mean overall score across all three tasks. The sustainability score $S_{\text{sus}}$ is computed as described in Section~\ref{sec:sustainability}.}
    \label{tab:sustainability_top10}
    \begin{tabular}{clccc}
        \toprule
        Rank & Model & Energy (Wh) & Performance & $S_{\text{sus}}$ \\
        \midrule
        1 & Teuken 7B Commercial v0.4 & 0.647 & 0.476 & 0.822 \\
        2 & Gemma2 2B & 0.701 & 0.625 & 0.813 \\
        3 & Gemma2 9B & 0.773 & 0.696 & 0.801 \\
        4 & EuroLLM 9B & 0.809 & 0.627 & 0.795 \\
        5 & Gemma3 4B & 0.996 & 0.666 & 0.769 \\
        6 & Mistral 7B v0.3 & 1.013 & 0.565 & 0.767 \\
        7 & Llama 3.2 3B & 1.082 & 0.618 & 0.758 \\
        8 & GPT-4o Mini & 1.128 & 0.699 & 0.753 \\
        9 & Mistral Nemo 12B & 1.188 & 0.614 & 0.746 \\
        10 & Smollm 1.7B & 1.224 & 0.426 & 0.742 \\
        \bottomrule
    \end{tabular}
\end{table}

\subsubsection{Transparency}
\label{sec:transparency_results}

In total, we assessed 39 models from 13 model providers on their transparency practices against the seven domains and 21~questions described in Section~\ref{sec:transparency}, resulting in 819 scored answers (see Appendix~\ref{app:transparency_questions} for the full question set).

\paragraph{Overall Scores}
Transparency scores (out of a maximum of 42) range from 18 (Nous Hermes 2 Mixtral 8x7B\footnote{Nous Hermes 2 Mixtral 8x7B is a DPO fine-tune of Mistral's Mixtral 8x7B released by a third party (Nous Research). Fine-tuned models may retain the base model's functionalities but not its documentation, raising the question of how transparency requirements should apply to derivative model releases.}, 42.9\%) to 38 (Apertus 70B, 90.5\%), with a mean of 26.9 (64.1\%).
Table~\ref{tab:transparency_top10} presents the top-10 models.
The highest score is achieved by Apertus 70B from Swiss~AI, which was explicitly designed with EU~AI~Act compliance in mind~\citep{apertus2025apertus}, followed by Apertus 8B and models from Google, Meta, Microsoft, and OpenAI.

\begin{table}[t]
    \centering
    \caption{Top-10 models by transparency score (max 42), evaluated against 21 EU~AI~Act documentation criteria. Eight models are tied at rank~3 with a score of 31; within this group the listing order is arbitrary.}
    \label{tab:transparency_top10}
    \begin{tabular}{clllc}
        \toprule
        Rank & Model & Organisation & Size & Score \\
        \midrule
        1 & Apertus 70B & Swiss AI & Large & 38 \\
        2 & Apertus 8B & Swiss AI & Small & 32 \\
        3 & Gemma2 2B & Google & Small & 31 \\
        3 & Gemma2 9B & Google & Small & 31 \\
        3 & Gemma3 27B & Google & Medium & 31 \\
        3 & GPT-OSS 20B & OpenAI & Medium & 31 \\
        3 & GPT-OSS 120B & OpenAI & Large & 31 \\
        3 & Llama 3.2 1B & Meta & Small & 31 \\
        3 & Llama 3.2 3B & Meta & Small & 31 \\
        3 & Phi-4 & Microsoft & Small & 31 \\
        \bottomrule
    \end{tabular}
\end{table}

\begin{figure}[h]
    \centering
    \includegraphics[width=\linewidth,height=0.6\textheight,keepaspectratio]{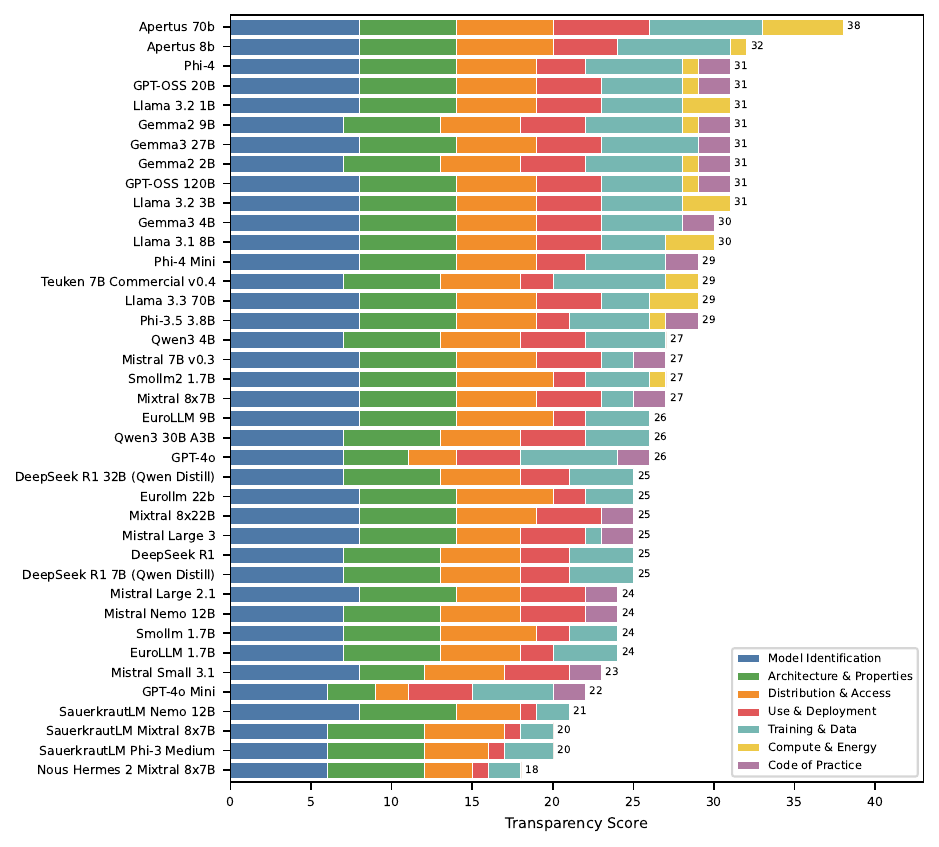}
    \caption{Transparency scores for all 39 models, sorted by domain and total score. Each segment represents the score in one of the seven Transparency Matrix documentation domains. The pronounced gap between well-documented domains (Model Identification, Architecture, Distribution \& Access) and poorly-documented ones (Use \& Deployment, Training \& Data, Compute \& Energy) is visible across nearly all models.}
    \label{fig:transparency_stacked}
\end{figure}

\paragraph{Domain-Level Analysis}
Figure~\ref{fig:transparency_stacked} shows the total score of each model decomposed by domain.
Transparency varies dramatically across domains: Architecture \& Properties is nearly universally documented (mean 97.0\%)\footnote{The few gaps in Architecture \& Properties stem from providers that do not disclose their models' parameter counts.} and so is Model Identification (93.6\%)\footnote{Model Identification falls short of 100\% primarily because several providers do not clearly state release dates for their models.}, while Distribution \& Access also reaches 81.2\%.
In contrast, Compute \& Energy remains the most opaque domain (mean 11.5\%), with 84.6\% of models providing no information on energy consumption and 79.5\% providing no energy measurement methodology.
Training \& Data is also insufficiently documented (mean 48.4\%): 51.3\% of models provide no information on measures against unsuitable training data, 38.5\% provide no bias mitigation measures, and only 7 of 39 models fully document their data curation process.
Use \& Deployment scores 54.3\% on average, dragged down primarily by the AI~Systems criterion (which asks providers to identify the AI-system categories their model is suited for): 97.4\% of models provide no such description, making this the single worst-documented criterion in the entire matrix.\footnote{Despite the GPAI Code of Practice being directed precisely at general-purpose models like LLMs, providers rarely enumerate the AI-system categories their model is suited for, instead marketing them as broadly general-purpose tools.}
This pattern mirrors findings from the Stanford Foundation Model Transparency Index~\cite{wan20252025}, which identifies training data and compute as the two most opaque areas across major AI companies.

\begin{figure}[t]
    \centering
    \includegraphics[width=\linewidth]{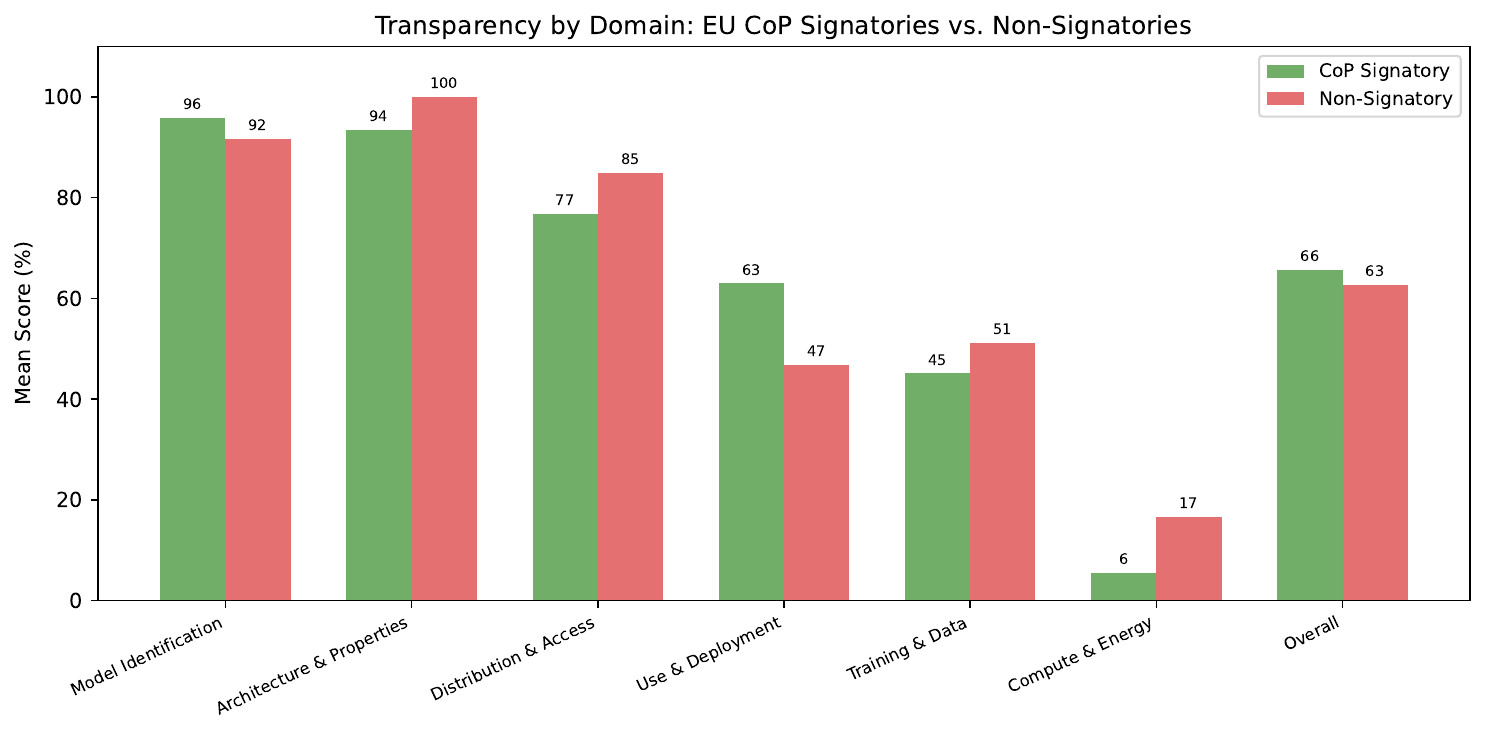}
    \caption{Mean transparency score (\%) by domain for EU~AI~Act Code of Practice signatories vs.\ non-signatories.}
    \label{fig:transparency_signatory_domains}
\end{figure}

\paragraph{EU Code of Practice Signatories}
Of the 39 models, 18 were developed by organisations that have signed the EU~AI~Act General-Purpose AI Code of Practice, including Google, Microsoft, Mistral and \mbox{OpenAI}, which are four of the 13 model providers in our sample, but account for 46\% of evaluated models due to their larger model portfolios.
Organisations such as Meta have decided against signing, while xAI has agreed to sign only the security chapter, explicitly not including the transparency and copyright chapters.
Signatories score slightly higher than non-signatories (mean 27.6 vs.\ 26.3), though the difference is marginal.
At the domain level (Figure~\ref{fig:transparency_signatory_domains}), signatories score notably higher on Use \& Deployment (63.0\% vs.\ 46.8\%).
This likely reflects the composition of the signatory group, which consists predominantly of commercial API providers (Google, Microsoft, Mistral, OpenAI) for whom Acceptable Use Policies and explicit intended-use documentation are operationally and legally necessary to govern downstream access; the non-signatory group is more heavily weighted toward open-weight releases by research labs and third-party fine-tuners, where such artefacts are less commercially essential.
However, non-signatories score higher on Distribution \& Access (84.9\% vs.\ 76.8\%), Architecture \& Properties (100\% vs.\ 93.5\%), Training \& Data (51.2\% vs.\ 45.1\%), and Compute \& Energy (16.7\% vs.\ 5.6\%), plausibly because the non-signatory group contains a higher share of open-weight model releases, where publishing model weights inherently requires documenting access methods and architecture.
In effect, the signatories' overall advantage is almost entirely driven by a single domain (Use \& Deployment); signing the Code of Practice does not result in greater transparency about upstream resources such as training data or computational costs.

\paragraph{Regional Differences}
US-based models score highest on average (mean 28.4, $n=18$), followed by European and Chinese models, which are essentially tied at 25.6 ($n=16$ and $n=5$ respectively).
However, given the small and unbalanced sample sizes, these differences should be interpreted with caution.
Nevertheless, despite the EU~AI~Act originating in Europe, European models do not demonstrate systematically higher transparency than US models on these criteria.

\paragraph{Large Commercial Providers vs.\ Smaller Providers}
Comparing large commercial AI companies ($n=27$) with smaller model providers, such as research labs and startups ($n=12$), reveals a modest difference in overall transparency (mean 27.6 vs.\ 25.3).
However, smaller providers show considerably higher variance (std 5.7 vs.\ 3.0), ranging from the lowest score in our benchmark (Nous Hermes 2 Mixtral 8x7B, 18/42) to the highest (Apertus 70B, 38/42).
This suggests that large companies converge around a similar level of documentation, while smaller organizations are more polarized: some primarily publish model weights with minimal documentation, while others like Swiss~AI invest heavily in transparency.
Part of this variance reflects a structural difference within the smaller-provider group: it includes both original model developers and third-party fine-tuners, who face fundamentally different documentation situations.
Fine-tuners such as VAGOSolutions or Nous Research typically document their own adaptations but do not replicate the base model provider's disclosures on training data or compute, resulting in systematically lower scores on upstream domains.

\subsubsection{Politics and Values}
\label{sec:politics_values_results}

In this section, we present the results of our \textit{politics and values} evaluation.
Recall that for the \textit{politics} component, we examine the extent to which LLMs can accurately reproduce the official positions of German political parties.
For the \textit{values} component, we evaluate model outputs by identifying stance directed toward values derived from the German \textit{Grundgesetz}, i.e. the German constitution.
In both experiments, we conduct analyses along three dimensions: \textit{organisation}, \textit{region}, and \textit{model size}.
\textit{Organisation} refers to the company responsible for developing the model, e.g. Mistral AI or DeepSeek.
\textit{Region} denotes the geographical location of the respective organisation.
We distinguish three regions: USA, China, and Europe.
Finally, we classify models into three parameter size categories \textit{small}, \textit{medium}, and \textit{large}, using the same category definitions as in our performance evaluation. 

\paragraph{Political Party Positions}

Beginning with our analysis on political party positions, we report the (aggregated) accuracy of correctly classifiying 4,788 positions across 64 political parties.

Table~\ref{tab:party_positions_top10} presents the top-10 highest-performing models, which span providers across all regions.
Half of the models originate from the USA, with OpenAI representing the most prominent provider.
The first rank is shared between DeepSeek's R1 and OpenAI's GPT-4o Mini, both yielding a mean accuracy of 0.671, suggesting the challenging nature of the task.
Among European models, Mistral attains the highest performance, with Mistral Small 3.1 obtaining the top score within this group.
While the leading models are predominantly developed by major organisations, namely DeepSeek, Meta, Mistral, and OpenAI, the fine-tuned model \textit{SauerkrautLM Mixtral 8x7B}, developed by VAGOSolutions, also achieves a competitive ranking.

\begin{table}[t]
    \centering
    \caption{Top-10 models for party-position classification, ranked by mean accuracy.}
    \label{tab:party_positions_top10}
    \begin{tabular}{cllllc}
        \toprule
        Rank & Model & Organisation & Region & Size & Score \\
        \midrule
        1 & DeepSeek R1 & DeepSeek & China & Large & 0.671 \\
        1 & GPT-4o Mini & OpenAI & USA & Small & 0.671 \\
        3 & GPT-OSS 120B  & OpenAI & USA & Large & 0.667 \\
        4 & Llama 3.3 70B & Meta & USA & Large & 0.663 \\
        5 & Mistral Small 3.1 & Mistral & Europe & Medium & 0.645\\
        6 & GPT-4o	& OpenAI	 & USA & Large & 0.624 \\
        7 & GPT-OSS 20B	 & OpenAI  & USA & Medium & 0.619\\
        8 & Mistral Large 3 & Mistral & Europe & Large & 0.610 \\
        9 & SauerkrautLM Mixtral 8x7B & VAGOSolutions  & Europe & Medium & 0.607 \\
        10 & DeepSeek R1 32B & DeepSeek  & China & Medium & 0.601 \\
        \bottomrule
    \end{tabular}
\end{table}

\begin{figure}[htbp]
    \centering
    \includegraphics[width=\linewidth]{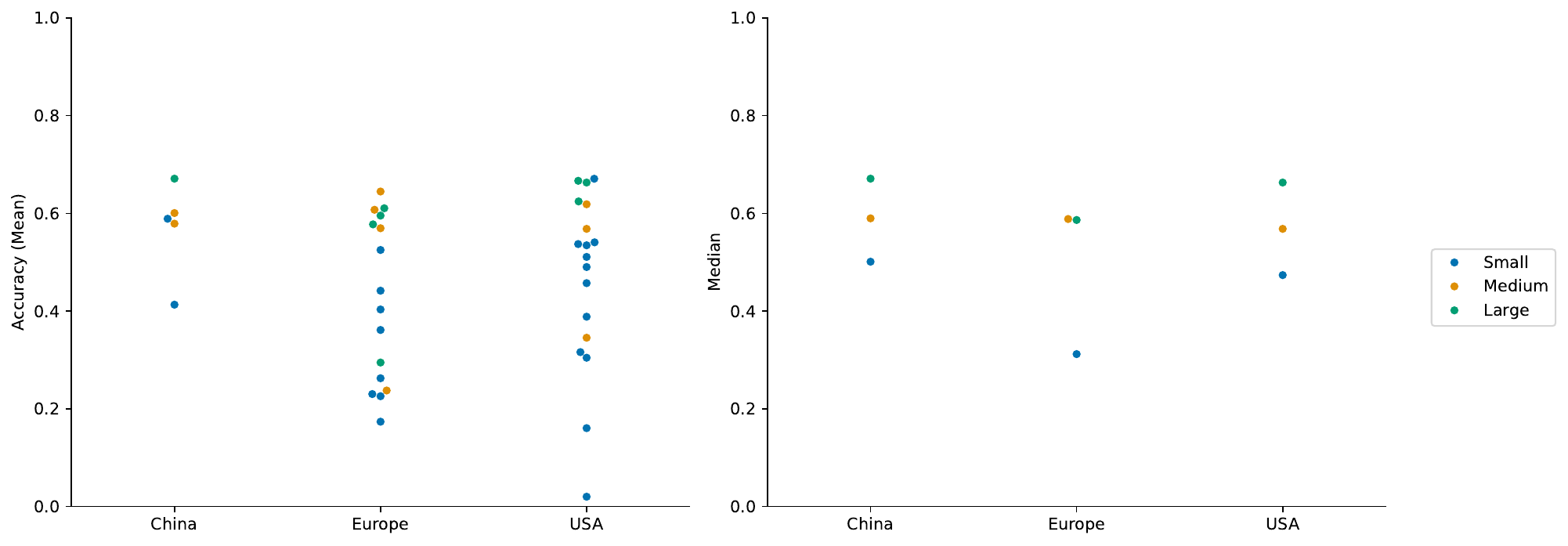}
    \caption{An analysis of classification performance on political party positions grouped by region and model size. The first plot shows accuracy for each model, while the second plot shows medians of model performance.}
    \label{fig:swarm_scatter_politics_region_by_size}
\end{figure}

The following analysis examines model results by region and parameter size.
The first plot of Figure~\ref{fig:swarm_scatter_politics_region_by_size} illustrates the distribution of classification accuracies across both dimensions, whereas the second plot presents the corresponding medians.
Chinese models exhibit the lowest dispersion (SD 0.095)
While this may partly reflect the comparatively small number of models, it is still notable that the majority of evaluated Chinese models appear to perform competitively.
European (0.168) and US models (0.181) display a comparable standard deviation.
With respect to model size, the results indicate a positive correlation between parameter count and classification accuracy.
Large models tend to achieve higher accuracies, whereas small models yield substantially lower scores.
This scaling effect is particularly pronounced among European models, which show a median accuracy of 0.312 for small models compared to $\approx0.587$ for both medium-sized and large models, thereby indicating a steeper scaling curve relative to other regions. 
However, parameter scaling alone does not fully explain the observed performance, as the large DeepSeek R1 and the small GPT-4o Mini achieve identical accuracy scores.
Furthermore, fine-tuning may represent a viable strategy for performance improvement, as demonstrated by the competitive results of SauerkrautLM Mixtral 8x7B.

\paragraph{Constitutional Values}

\begin{table}[t]
    \centering
    \caption{Models yielding particularly high or low stance scores as calculated by the LLM-as-a-Judge metric.}
    \label{tab:values_llmaj_stance_scores}
    \begin{tabular}{llllc}
        \toprule
        Model & Organisation & Region & Size & Score \\
        \midrule
        Llama 3.3 70B & Meta & USA & Large & 0.986 \\
        GPT-OSS 20B & OpenAI & USA & Medium & 0.979 \\
        Qwen3 30B A3B & Alibaba & China & Medium & 0.978\\
        GPT-OSS 120B & OpenAI & USA & Large & 0.976\\
        SauerkrautLM Mixtral 8x7B & VAGOSolutions & Europe & Medium & 0.969 \\
        Mixtral 8x7B & Mistral & Europe & Medium & 0.964 \\
        DeepSeek R1 & DeepSeek & China & Large & 0.964 \\
        Nous Hermes 2 Mixtral 8x7B & Nous Research & USA & Medium & 0.961 \\
        Apertus 8B & Swiss AI & Europe	 & Small & 0.960 \\
        Gemma3 4B & Google & USA & Small & 0.960 \\
        \midrule
        Llama 3.1 8B & Meta & USA & Small & 0.899 \\
        Qwen3 4B	 & Alibaba & China & Small & 0.894 \\
        SauerkrautLM Phi-3 Medium & VAGOSolutions & Europe & Small & 0.891 \\
        DeepSeek R1 32B & DeepSeek & China & Medium & 0.881 \\
        EuroLLM 1.7B & UTTER & Europe & Small & 0.814 \\
        DeepSeek R1 7B & DeepSeek	 & China & Small & 0.796 \\
        Phi-4 Mini & Microsoft & USA	 & Small & 0.790\\
        Llama 3.2 1B & Meta	 & USA & Small	 & 0.778\\
        Smollm2 1.7B	 & Huggingface & USA & Small & 0.730 \\
        Smollm 1.7B & Huggingface & USA & Small & 0.545 \\
        \bottomrule
    \end{tabular}
\end{table}

\begin{table}[t]
    \centering
    \caption{Models yielding particularly high or low stance scores as calculated by the GottBERT-based metric.}
    \label{tab:tab:values_gottbert_stance_scores}
    \begin{tabular}{llllc}
        \toprule
        Model & Organisation & Region & Size & Score \\
        \midrule
        Apertus 8B & Swiss AI & Europe & Small & 0.904 \\
        SauerkrautLM Mixtral 8x7B	& VAGOSolutions	 & Europe & Medium & 0.896 \\
        Nous Hermes 2 Mixtral 8x7B	& Nous Research	 & USA	 & Medium & 0.891 \\
        Mistral Small 3.1 & Mistral & Europe & Medium & 0.881 \\
        Mistral Large 2.1 & Mistral & Europe & Large & 0.880 \\
        Llama 3.3 70B & Meta	 & USA & Large & 0.871 \\
        GPT-4o Mini	& OpenAI & USA & Small & 0.868 \\
        Phi-4 & Microsoft & USA	 & Small & 0.842 \\
        Llama 3.2 3B	 & Meta & USA & Small & 0.836 \\
        Teuken 7B Commercial v0.4	& OpenGPTX & Europe & Small & 0.834 \\
        \midrule
        Smollm2 1.7B & Huggingface & USA & Small & 0.710 \\
        Gemma3 27B & Google &	USA & Medium & 0.691 \\
        GPT-OSS 20B & OpenAI & USA & Medium & 0.644 \\
        DeepSeek R1 32B & DeepSeek & China & Medium & 0.641 \\
        Smollm 1.7B & Huggingface & USA & Small & 0.639 \\
        GPT-OSS 120B & OpenAI & USA & Large & 0.621 \\
        EuroLLM 22B  & UTTER & Europe & Medium & 0.616 \\
        Qwen3 30B A3B & Alibaba & China & Medium &	0.606 \\
        DeepSeek R1 & DeepSeek & China & Large & 0.588 \\
        Mistral Large 3 & Mistral & Europe & Large & 0.536 \\
        \bottomrule
    \end{tabular}
\end{table}

Turning to the \textit{values} analysis, we again report results along the analytical dimensions \textit{organisation}, \textit{region}, and \textit{size}.
We present scores obtained from two distinct evaluation metrics: 1) an LLM-as-a-Judge approach and 2) a fine-tuned GottBERT classifier, which assess model-generated stance-bearing statements with respect to normative values drawn from the German constitution (see Section \ref{sec:politics_values} for details).

Table~\ref{tab:values_llmaj_stance_scores} presents the LLM-as-a-Judge stance scores aggregated across values, scenarios, and repeated measures (40 measurements in total) for the models at the upper and lower ends of the score distribution. 
Table~\ref{tab:tab:values_gottbert_stance_scores} reports the corresponding results obtained using the GottBERT metric. 
In contrast to the evaluations presented previously, however, these scores do not reflect performance rankings. 
Rather, they capture the tendency of models to generate statements interpretable as expressing positive, negative, or balanced stance.

Beginning with the LLM-as-a-Judge analysis, no model in the evaluated set exhibits distinct negative stance towards the values under consideration.
The lowest aggregated (and somewhat neutral) stance score is yielded by Huggingface's Smollm 1.7B (0.545). 
With respect to regional distribution, models are represented comparably across both the higher- and lower-scoring ends of the spectrum.
However, while models producing more positive-leaning stance vary in size, the majority of lower-scoring models are small models. 
Models associated with particularly positive stance include Llama 3.3 70B (0.986; Meta, USA), Qwen3 30B A3B (0.978; Alibaba, China), and SauerkrautLM Mixtral 8x7B (0.969; VAGOSolutions, Europe). 
Models attaining comparatively lower stance scores include the aforementioned Smollm 1.7B, DeepSeek R1 7B (0.796, DeepSeek, China) and EuroLLM 1.7B (0.814; UTTER, Europe).

Applying the GottBERT metric, Mistra Large 3 obtains the lowest stance score (0.536), thereby replicating the finding that no model generates a clearly negative stance on average. 
Notably, the GottBERT metric yields systematically lower scores overall compared to the LLM-as-a-Judge approach. 
The most-positive stance score is obtained by Apertus 8B (0.904; Swiss AI, Europe), followed by SauerkrautLM Mixtral 8x7B (0.896; VAGOSolutions, Europe). 
No Chinese model is represented among the models with the highest mean scores under this metric. 
Notable US models include Nous Hermes 2 Mixtral 8x7B (0.891; Nous Research) and Llama 3.3 70B (0.871; Meta). 
Lower GottBERT-based scores are obtained, among others, by Mistral Large 3, DeepSeek R1 (0.588; DeepSeek, China) and GPT-OSS 120B (0.621; OpenAI, USA). 
Contrasting with the LLM-as-a-Judge results, model sizes are more evenly distributed across the scoring range in this analysis.

\begin{figure}[htbp]
    \centering
    \includegraphics[width=\linewidth]{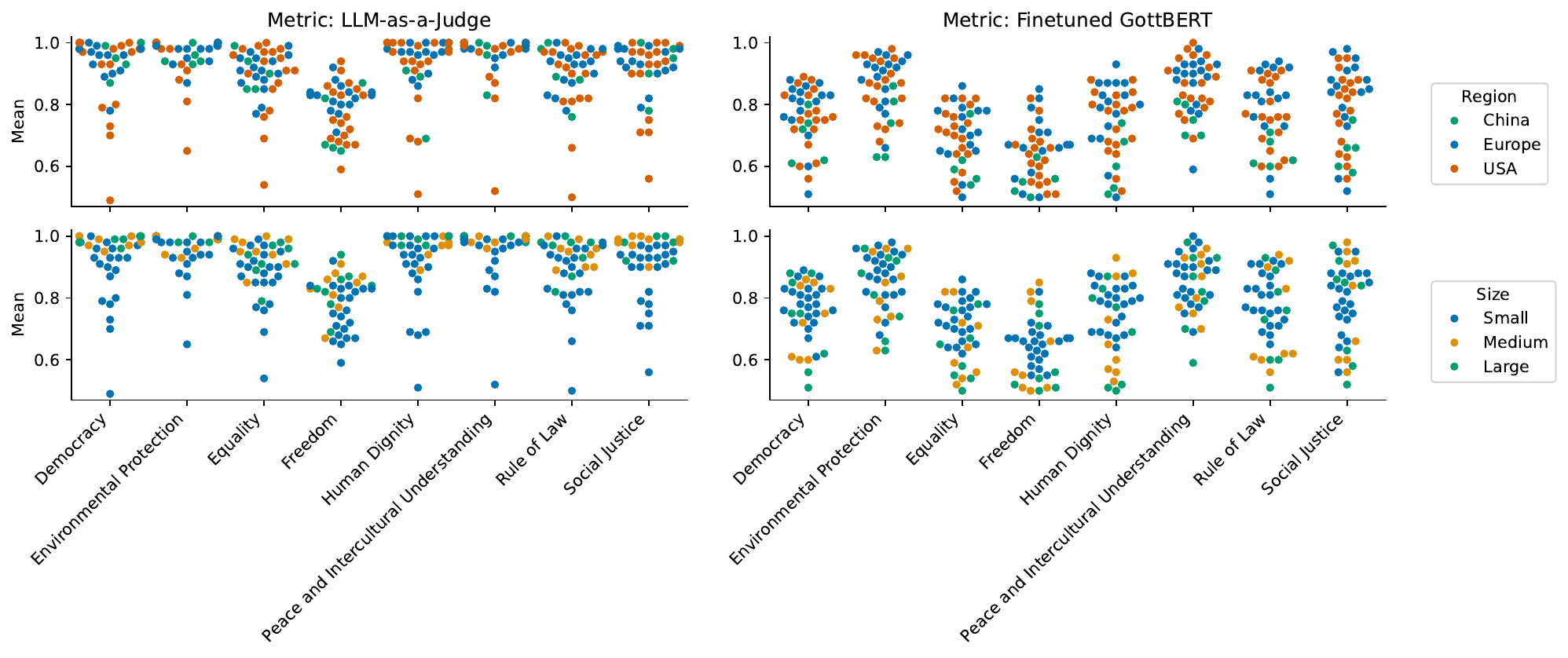}
    \caption{Mean of min-max normalized model results by value. Top plots show the region associated with each model, bottom plots depict the corresponding model size.}
    \label{fig:swarm_values}
\end{figure}

In the following, we analyse model performance at the level of individual values with respect to both region and model size.
Figure~\ref{fig:swarm_values} contains swarm plots of model-level mean scores for each value, where each mean represents the aggregation of the five scenario-specific measurements.
Consistent with the results reported in Table~\ref{tab:values_llmaj_stance_scores}, the LLM-as-a-Judge metric yields high scores across most values.
A notable exception is \textit{freedom}, for which mean scores range from 0.59 to 0.94 (median 0.82).
The results obtained with the fine-tuned GottBERT classifier exhibit a more heterogeneous pattern, with \textit{environmental protection} and \textit{peace and intercultural understanding} being examples of values obtaining consistently high mean scores across models (both median 0.87).
In contrast, values like \textit{equality} (median 0.71) and \textit{freedom} (median 0.65) are attain comparatively lower mean scores.
While region-related conclusions are difficult to draw from this analysis, the general pattern indicates that some values, e.g., \textit{freedom}, are associated with more diverse outputs, spanning both pro and contra arguments.
With respect to model size, the LLM-as-a-Judge metric indicates that small models tend to achieve lower mean scores (median 0.91) than medium-sized and large models (both median 0.98).
This pattern is less clearly discernible in the GottBERT-based evaluation, where the distribution of scores is more heterogeneous, thereby hindering a straightforward interpretation.

\begin{figure}[htbp]
    \centering
    \includegraphics[width=\linewidth]{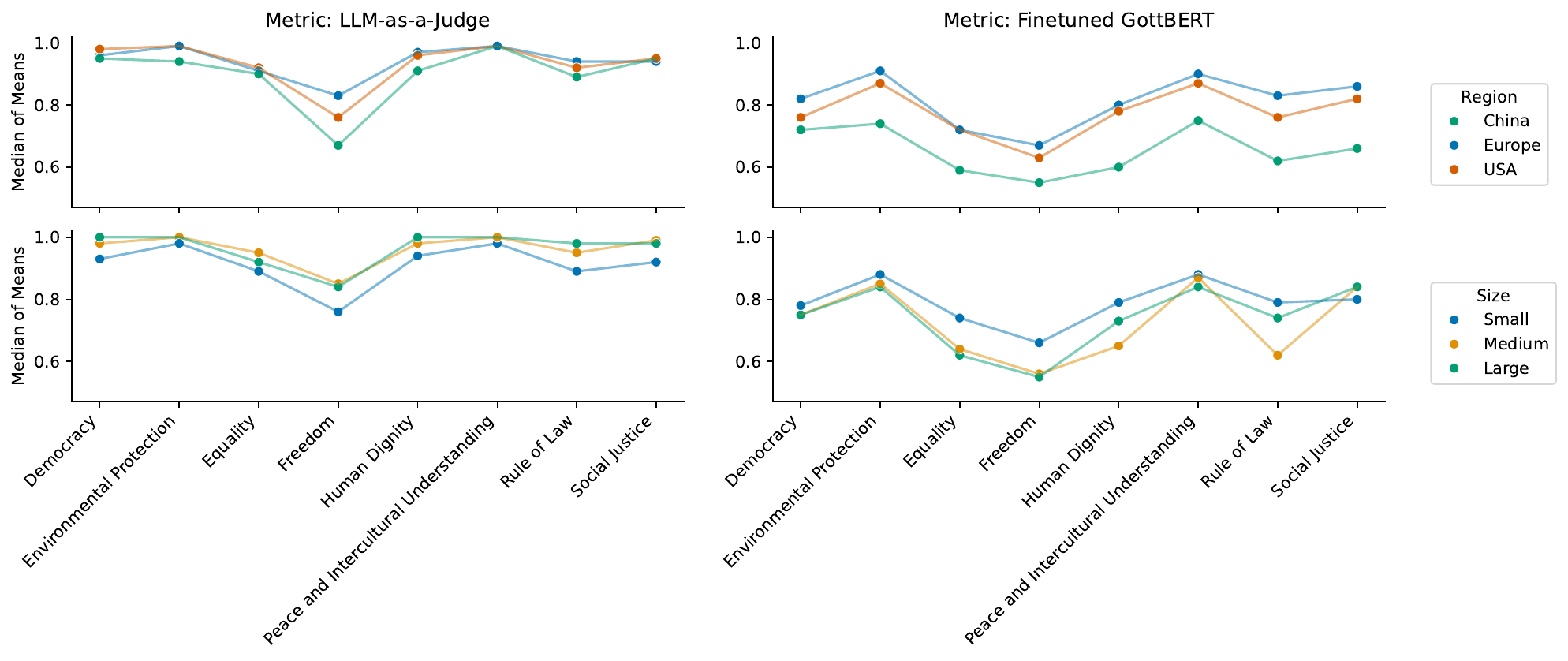}
    \caption{Median of means of min-max normalized model results by value. Top plots show medians by region, bottom plots show medians by size.}
    \label{fig:scatter_line_values_medians}
\end{figure}

Although the preceding analysis provides an overview of the distribution of model scores across values, it does not permit firm conclusions regarding systematic performance patterns.
To address this limitation, Figure \ref{fig:scatter_line_values_medians} presents the medians of means per value with respect to both region (top plots) and size (bottom plots).
Consistent with earlier findings, e.g., the lower overall scores for \textit{freedom}, the figure indicates region-specific performance differences.
Applying the GottBERT metric, European models attain the highest median of means for the majority of values, with US models ranking second, followed by Chinese models.
In line with the results shown in Figure \ref{fig:swarm_values}, the LLM-as-a-Judge metric yields generally high scores across all regions.
However, differences can be observed for \textit{freedom}, where European models attain the highest scores (median 0.83), whereas Chinese models obtain the lowest (median 0.67).
While requiring further proof, these findings support the idea that German (and European) values are captured during the training of European models.
Regarding model size, Figure \ref{fig:scatter_line_values_medians} confirms that small models achieve lower median scores than medium-sized and large models under the LLM-as-a-Judge evaluation.
In contrast, the GottBERT-based metric produces an inverse pattern, with small models yielding highest median of means for most values.

In summary, no model consistently generates content interpretable as expressing negative stance toward the evaluated values. 
Whether this reflects genuine alignment or a surface-level positivity bias in generation, remains an open question.
Nevertheless, the finding provides evidence that models rarely exhibit clearly negative stance with respect to constitutional values, irrespective of regional origin or parameter size.
Furthermore, we find that certain values elicit more heterogeneous responses than others, with \textit{freedom} constituting a particularly salient example.
We attribute this pattern to the contested nature of such concepts, which are likely to be represented through mixed or opposing perspectives in the training data.
For instance, while \textit{freedom} may be considered a fundamental value warranting protection, its exercise is subject to state restrictions in specific contexts, e.g., migration, which may be reflected in more ambivalent model outputs.
Finally, we observe that European models tend to achieve comparatively high stance scores, a pattern that may reflect regional differences in training data composition, pre-training paradigm, or alignment procedures. More research is needed, however, to provide evidence for this hypothesis.

%% file: sections/06_benchmark_evaluation.tex
\section{Benchmark Evaluation}
\label{ch:benchmark_evaluation}

In this section, we turn the lens on the MÖVE benchmark itself and examine five aspects of its methodological robustness.
First, we analyze the statistical precision of our rankings to determine how reliably the benchmark separates model scores (Section~\ref{sec:benchmark_precision}).
Second, we assess the reliability of our LLM-based judge metrics, both within a single judge and across different judge models (Section~\ref{sec:llm_judge_reliability}).
Third, we investigate whether the inclusion of our internally constructed datasets meaningfully affects model rankings compared to using public datasets alone (Section~\ref{sec:impact_of_internal}).
Fourth, we test the sensitivity of our rankings to prompt formulation (Section~\ref{sec:prompt_sensitivity}).
Finally, we validate our sustainability estimates by comparing them against hardware-level energy measurements (Section~\ref{sec:energy_validity}).

\subsection{Benchmark Precision Analysis}
\label{sec:benchmark_precision}
\begin{figure}[t]
    \centering
    \begin{subfigure}[t]{0.48\linewidth}
        \centering
        \includegraphics[width=\linewidth]{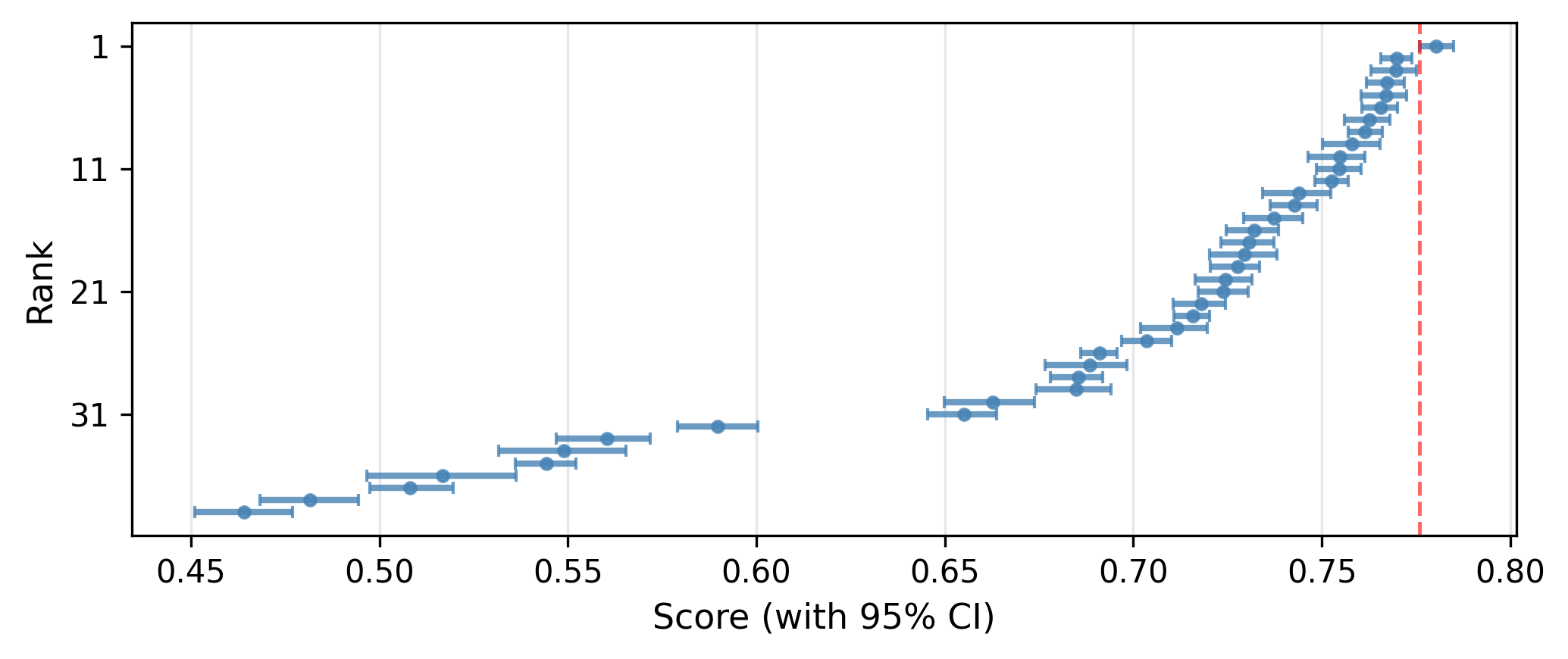}
        \caption{Summarization}
        \label{fig:forest_plot_summarization}
    \end{subfigure}
    \hfill
    \begin{subfigure}[t]{0.48\linewidth}
        \centering
        \includegraphics[width=\linewidth]{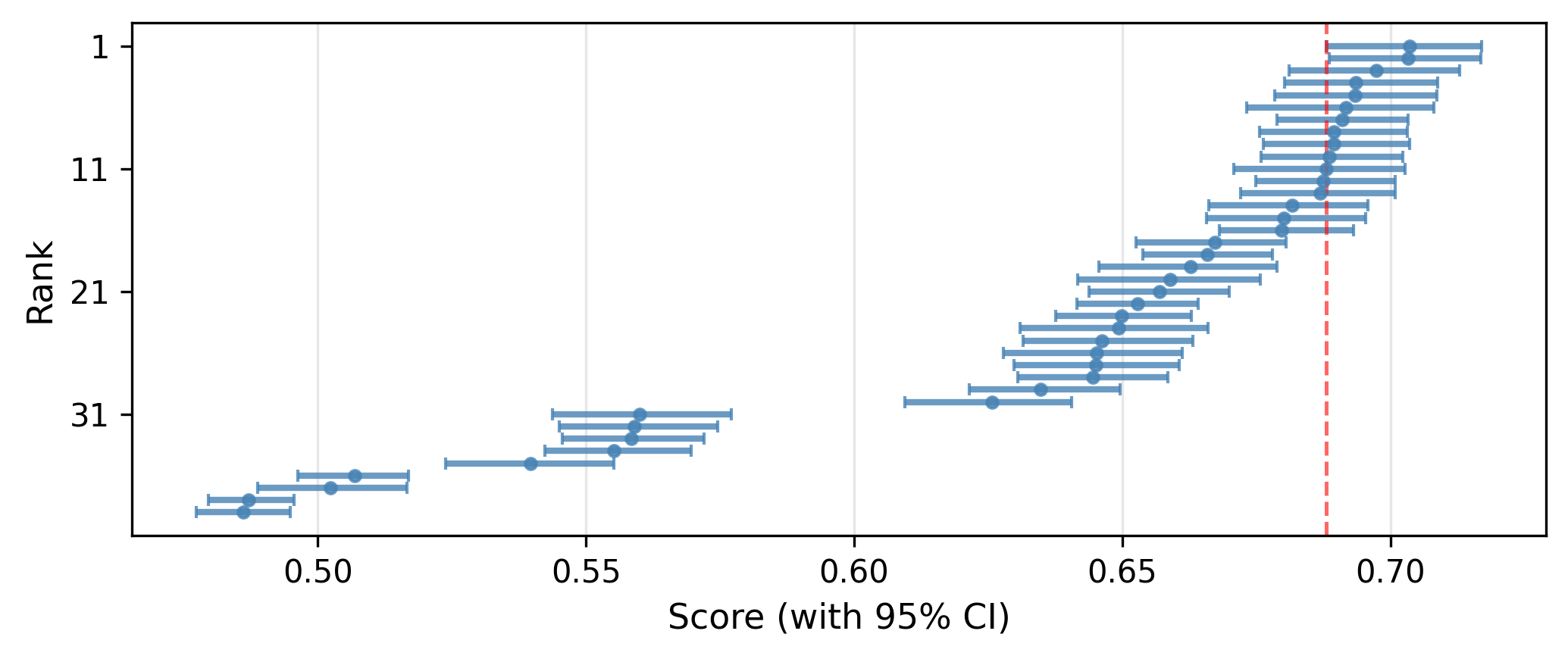}
        \caption{Question Answering}
        \label{fig:forest_plot_qa}
    \end{subfigure}

    \vspace{0.5em}

    \begin{subfigure}[t]{0.48\linewidth}
        \centering
        \includegraphics[width=\linewidth]{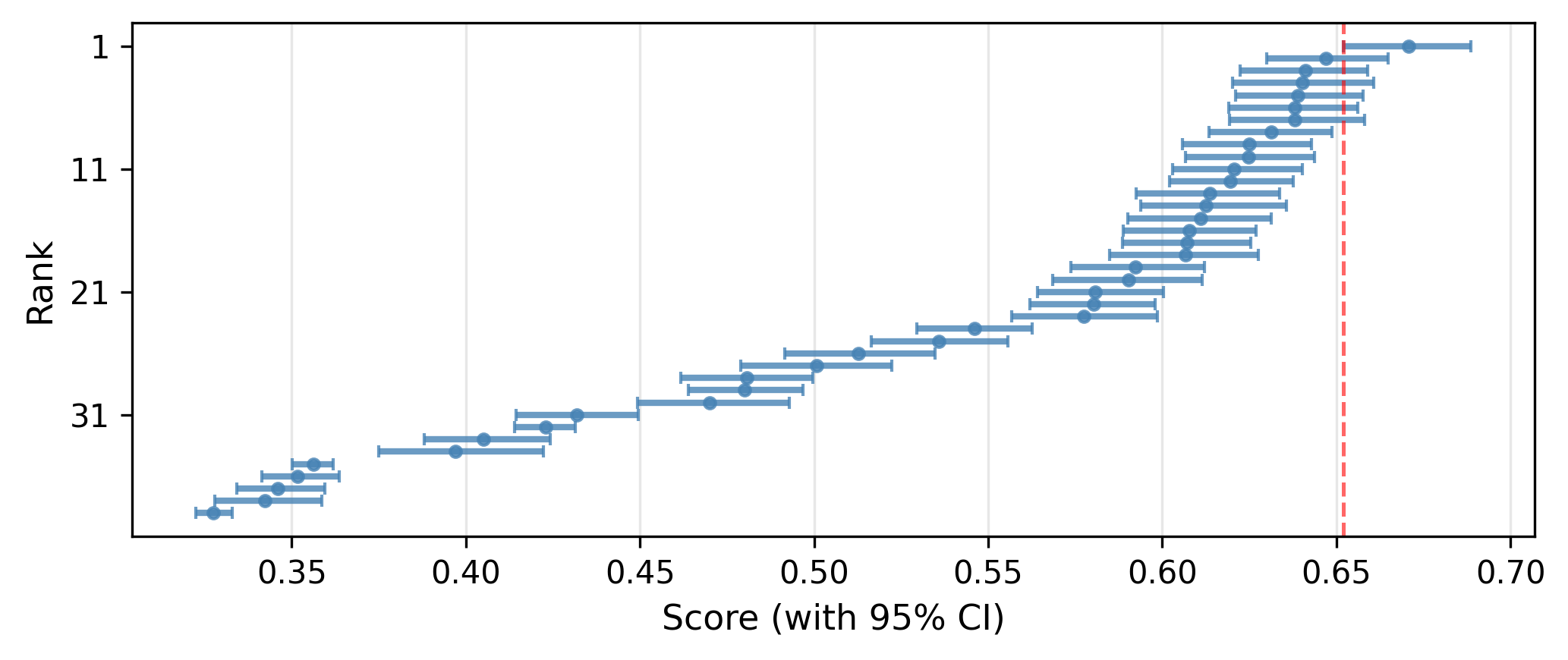}
        \caption{Topic Extraction}
        \label{fig:forest_plot_topic_extraction}
    \end{subfigure}
    \hfill
    \begin{subfigure}[t]{0.48\linewidth}
        \centering
        \includegraphics[width=\linewidth]{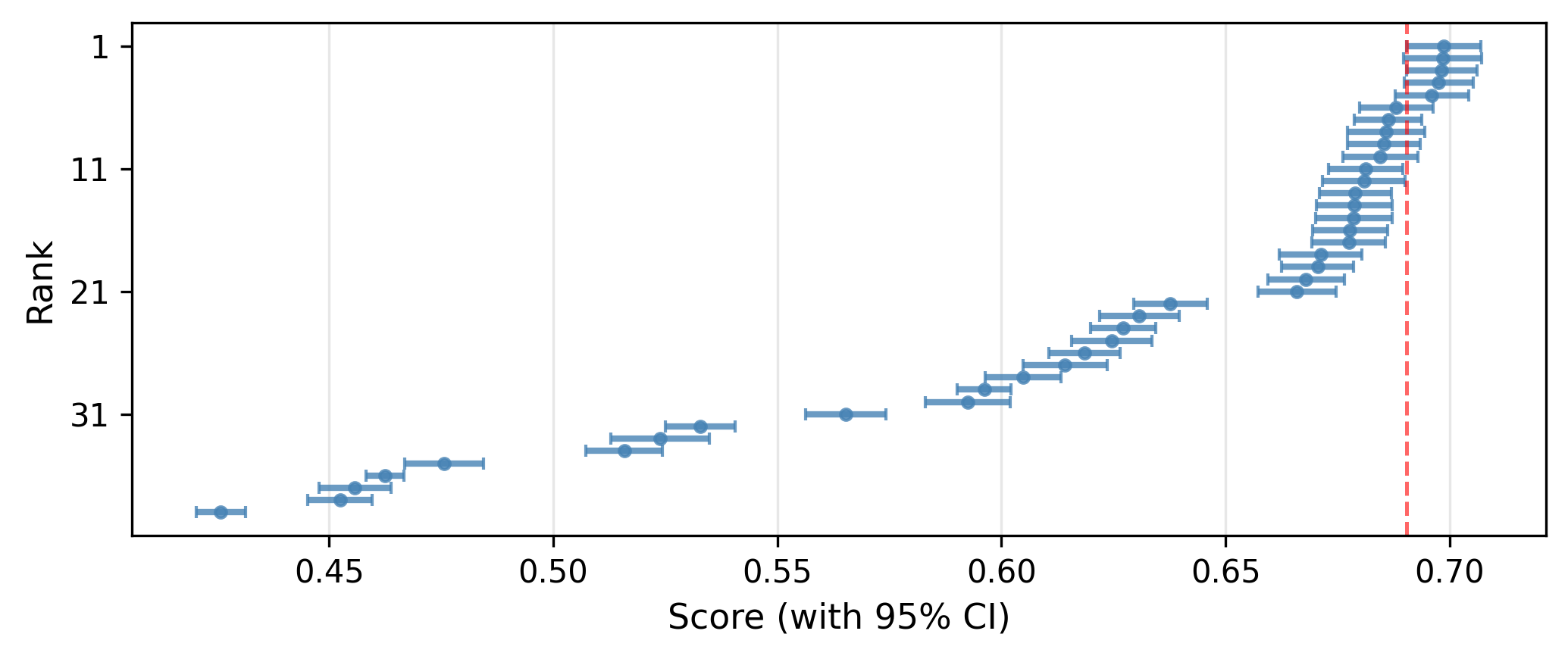}
        \caption{Performance Criteria Overall}
        \label{fig:forest_plot_combined}
    \end{subfigure}
    \caption{Forest plots showing model scores with 95\% bootstrap confidence intervals for each benchmark. Models are ranked by score (rank 1 at top). The red dashed line indicates the lower CI bound of the top-ranked model; models to the right of this line are statistically indistinguishable from the highest-ranked model.}
    \label{fig:forest_plots_individual}
\end{figure}

Benchmark leaderboards typically present model rankings without uncertainty estimates, leaving users unable to assess whether apparent differences reflect genuine functionality gaps.
We address this limitation by computing 95\% confidence intervals for each model's score, enabling more informed interpretation of rankings than is typically possible.
We therefore analyze the intrinsic statistical precision of each performance criterion in the \textit{MÖVE framework}, to assess how reliably it separates between models.
Using hierarchical bootstrap (resampling within datasets, then averaging across datasets), we compute 95\% confidence intervals for each model's score and derive three precision metrics: (i) CI width (both absolute and as a percentage of the score spread), (ii) the fraction of adjacent-ranked model pairs with non-overlapping CIs, and (iii) the size of the statistically indistinguishable ``tier'' around the top-ranked model.
Table~\ref{tab:benchmark_precision} summarizes these metrics across all three performance tasks, as well as for an aggregate ``Performance Criteria Overall'' score computed by averaging each model's task-specific scores.
Figure~\ref{fig:forest_plots_individual} visualizes the respective score distributions with confidence intervals.

\begin{table}[t]
    \centering
    \caption{Benchmark precision metrics for each task, computed over 39 models. \textit{Score Range} indicates the minimum and maximum model scores observed. \textit{CI Width (absolute)} is the median 95\% confidence interval width on the same 0--1 scale as the scores. \textit{CI Width (relative)} expresses the same value as a percentage of the score spread. \textit{Adjacent Disting.} is the percentage of consecutive-ranked model pairs with non-overlapping CIs. \textit{Top-1 Tier} is the number of models statistically indistinguishable from the top performer (including itself).}
    \label{tab:benchmark_precision}
    \begin{tabular}{lccccc}
        \toprule
        & & \multicolumn{2}{c}{CI Width} & & \\
        \cmidrule(lr){3-4}
        Task & Score Range & (absolute) & (relative) & Adjacent Disting. & Top-1 Tier \\
        \midrule
        Summarization      & 0.46--0.78 & 0.015 & 4.6\%  & 18.4\% & 1 \\
        Question Answering & 0.49--0.70 & 0.028 & 12.9\% & 5.3\%  & 16 \\
        Topic Extraction   & 0.33--0.67 & 0.037 & 10.8\% & 2.6\%  & 7 \\
        \midrule
        Perf.\ Criteria Overall & 0.43--0.70 & 0.016 & 6.0\% & 15.8\% & 10 \\
        \bottomrule
    \end{tabular}
\end{table}

\paragraph{Summarization}
The summarization benchmark exhibits the highest precision among the three tasks.
With a median CI width of only 0.015 (4.6\% of the score spread), model scores are estimated with high precision.
Consequently, 18.4\% of adjacent-ranked pairs (7 of 38) have non-overlapping confidence intervals, allowing statistically confident distinctions between neighboring ranks.
Most notably, the top-ranked model (Mistral-Large-3) stands alone in its tier: no other model has an overlapping CI, making it a statistically clear winner.

\paragraph{Question Answering}
The QA benchmark shows lower ranking resolution than summarization: CI widths are nearly three times larger relative to the score spread (12.9\%), and only 5.3\% of adjacent pairs (2 of 38) are statistically distinguishable.
The top-ranked model (Mistral-Large-2411) cannot be distinguished from 15 other models, yielding a 16-model top tier.
This indicates that many models have achieved comparable QA performance at current state-of-the-art levels, a finding that is itself informative for practitioners, who can select among top-tier models based on other criteria (e.g.\ sustainability, transparency, or value alignment) without sacrificing QA quality.

\paragraph{Topic Extraction}
Topic extraction shows the lowest ranking resolution, with only 1 of 38 adjacent pairs (2.6\%) statistically distinguishable and a 7-model top tier.
This reflects having only two datasets for this task, compared to four for summarization.
Nevertheless, the CI width of 0.037 (10.8\% of the score spread) remains small enough to reliably distinguish top-tier from lower-scoring models.

\paragraph{Combined Performance Score}
We also compute an aggregate ``Performance Criteria Overall'' score by averaging each model's scores across summarization, QA, and topic extraction.
The combined score shows intermediate precision (CI width 0.016, or 6.0\% of the spread; 15.8\% adjacent pairs distinguishable), benefiting from error averaging across tasks.
The top-ranked model (GPT-4o Mini) shares its tier with 9 other models, reflecting the aggregation of uncertainty across the constituent benchmarks.

\paragraph{Implications}
These results demonstrate that our benchmark provides meaningful precision across all tasks.
Even for QA and topic extraction, where relative CI widths are larger, the absolute uncertainty of 0.028--0.037 reliably separates high-scoring models from mid-range and low-scoring ones.
For summarization, rankings are meaningful down to individual positions; for QA and topic extraction, rankings should be interpreted as performance tiers: reporting that a model belongs to a ``top tier'' is more accurate than claiming a precise rank.
Confidence interval reporting remains uncommon in benchmark evaluations; by providing these estimates, we enable more principled model selection than is typically possible.
Expanding the dataset portfolio for QA and topic extraction would further improve ranking resolution.

\paragraph{Dataset Weighting}
Our current aggregation weights all datasets equally, regardless of sample size.
This ensures that our smaller gold-standard datasets retain comparable influence to both larger silver-standard and already existing datasets, thereby maintaining domain-specific coverage.
However, smaller datasets produce noisier estimates, which inflates overall CI widths.
We assessed a sample-size-proportional weighting approach, in which each dataset's share scales with its sample count.
This reduces CI widths by 56--68\% across tasks and improves ranking precision: for QA, the top-tier size shrinks from 16 to 4 models; for topic extraction, from 7 to 1.
The trade-off is that the evaluation signals from our gold- and silver-standard datasets become diluted, making the benchmark more similar to an evaluation based mainly on already existing datasets.
We retain equal weighting to preserve the influence of our domain-specific gold- and silver-standard datasets and to mitigate the risk of data contamination, where models might be inadvertently trained on data from already existing datasets.
We note that expanding our gold- and silver-standard datasets would achieve both higher statistical precision and stronger domain-specific representation.

\subsection{LLM Judge Reliability}
\label{sec:llm_judge_reliability}
Several metrics in our evaluation (e.g., Faithfulness, Noise Sensitivity, Factual Correctness, and Topic Match\footnote{Topic Match is the LLM-judged component of the Topic Adherence metric.}) rely on LLM-based judgments, which are non-deterministic\footnote{We use GPT-4o Mini as our judge model with temperature zero and a fixed seed, yet observe variance across runs. The OpenAI API does not guarantee deterministic outputs even under these settings.} and introduce variance beyond sampling uncertainty.
We investigate two aspects of this variance: within-judge stability (how consistent is a single judge across repeated runs?) and inter-judge agreement (do different judge models produce consistent evaluations?).

\subsubsection{Within-Judge Stability}
\label{sec:within_judge_stability}
To quantify within-judge variance, we sampled 200 instances per task and evaluated each LLM-based metric 10 times with independent calls to GPT-4o Mini, which serves as the judge model throughout our evaluation.

\begin{table}[t]
    \centering
    \captionsetup{width=\textwidth}
    \caption{Within-judge stability: 95\% CI width for mean scores across 10 repeated runs.}
    \label{tab:llm_judge_stability}
    \begin{tabular}{lc}
        \toprule
        Metric & CI Width \\
        \midrule
        Factual Correctness & 0.009 \\
        Faithfulness & 0.014 \\
        Noise Sensitivity & 0.021 \\
        Topic Match & 0.011 \\
        \bottomrule
    \end{tabular}
\end{table}

Table~\ref{tab:llm_judge_stability} reports the 95\% confidence intervals for each metric's mean score.
The CI widths range from 0.009 (Factual Correctness) to 0.021 (Noise Sensitivity).
Relative to the full score spread across models (see Section~\ref{sec:benchmark_precision}), these CI widths represent a small fraction of the range, sufficient to reliably distinguish high- from low-scoring models.
 However, these CI widths are comparable in magnitude to typical score differences between adjacently ranked models, reinforcing our earlier recommendation to interpret rankings as broad performance tiers rather than precise orderings.

\subsubsection{Inter-Judge Agreement}
\label{sec:inter_judge_agreement}

The preceding analysis quantifies variance from repeated calls to a single judge model.
A separate concern is whether different judge models would produce consistent evaluations.
If inter-judge agreement is low, benchmark results become contingent on an arbitrary choice of judge model, limiting their generalizability.

To assess inter-judge reliability, we selected three judge models from different providers: GPT-4o Mini (OpenAI), Llama 3.3 70B (Meta), and Mistral-Large-3 (Mistral AI).
For each LLM-based metric, we sampled 200 instances from the corresponding task datasets: 40 samples from each of the three judge models' outputs, plus 80 from other evaluated models.
Each judge independently evaluated all 200 samples, and the balanced sampling enables us to also test for self-scoring bias.
We report two complementary agreement metrics: Krippendorff's $\alpha$~\citep{krippendorff2013content}, which measures agreement on absolute scores, and the average pairwise Spearman correlation, which measures agreement on sample rankings.
For Krippendorff's $\alpha$, values above 0.67 are conventionally considered indicative of acceptable reliability, and values above 0.80 of good reliability.

\begin{table}[t]
    \centering
    \captionsetup{width=\textwidth}
    \caption{Inter-judge reliability for LLM-based metrics, computed over 200 samples evaluated by three judge models (GPT-4o Mini, Llama 3.3 70B, Mistral-Large-3). Krippendorff's $\alpha$ measures agreement on absolute scores; Spearman $\rho$ measures agreement on sample rankings. Values above 0.67 are conventionally considered acceptable reliability.}
    \label{tab:inter_judge_reliability}
    \begin{tabular}{lcc}
        \toprule
        Metric & Krippendorff's $\alpha$ & Avg.\ Spearman $\rho$ \\
        \midrule
        Factual Correctness (QA) & 0.679 & 0.684 \\
        Factual Correctness (Summ.) & 0.547 & 0.536 \\
        Topic Match & 0.549 & 0.574 \\
        Faithfulness & 0.510 & 0.550 \\
        Noise Sensitivity & 0.421 & 0.430 \\
        \bottomrule
    \end{tabular}
\end{table}

Table~\ref{tab:inter_judge_reliability} reports the inter-judge reliability for each metric.
Unlike the within-judge analysis, we report Factual Correctness separately for QA and summarization because we observed that task context substantially affects inter-judge agreement.
Factual Correctness applied to QA samples is the only metric that achieves acceptable reliability ($\alpha = 0.68$), while the same metric applied to summarization samples falls well below the threshold ($\alpha = 0.55$).
The remaining metrics show lower agreement, with Noise Sensitivity exhibiting particularly poor reliability ($\alpha = 0.42$).

Notably, the Spearman rank correlations are nearly identical to the score-based metrics (Krippendorff's $\alpha$), indicating that the low agreement is not merely due to calibration differences between judges.
If judges simply operated on different scales but produced consistent relative rankings, we would observe low $\alpha$ but high Spearman $\rho$.
Instead, the similar values indicate substantive inconsistencies in judge scores, which is a more fundamental form of score divergence that directly affects model rankings.

Interestingly, the same metric can yield different reliability depending on task: Factual Correctness achieves higher agreement on QA samples ($\alpha = 0.68$) than on summarization ($\alpha = 0.55$).
QA outputs are shorter and more constrained, making factual accuracy easier to verify than for summaries, where judges have more latitude in what counts as an accurate representation of the source.

\paragraph{Comparison with Within-Judge Variance}
Notably, the low inter-judge agreement cannot be attributed to the stochastic nature of LLM outputs.
To compare variance sources on the same scale, we compute the within-judge standard deviation from the 10-run stability experiment above (average SD of scores for a single sample across the 10 runs) and compare it to the between-judge standard deviation (average SD across the three judges for each sample).
Within-judge SD ranges from 0.043 (Faithfulness) to 0.054 (Noise Sensitivity), while between-judge SD ranges from 0.100 (Faithfulness) to 0.221 (Noise Sensitivity).
The between-judge variance is thus 2--4 times larger than the within-judge variance, depending on the metric.
This suggests that the inter-judge divergence is systematic rather than stochastic, i.e., different judge models produce consistently divergent scores when evaluating the same outputs, and this effect exceeds the random variation observed when re-running a single judge.

This interpretation is further supported by examining score distributions across judges.
For instance, on the Faithfulness metric, Mistral-Large-3 assigns a mean score of 0.898 while GPT-4o Mini assigns 0.823, a consistent 0.075-point calibration difference that persists across samples.
Similar systematic biases appear across all metrics, suggesting that judge models differ not only in their noise characteristics but in their underlying scoring criteria.

\paragraph{Qualitative Analysis of Patterns of Divergence}
Motivated by these findings, we conducted a qualitative analysis of high-divergent samples to understand the underlying causes.
The patterns are striking: for Faithfulness, GPT-4o Mini is the lowest scorer on 90\% of samples, while for Topic Match it scores lowest on 76\% of samples with Llama 3.3 70B scoring highest on 66\%.
Inspection of individual cases reveals cross-judge variation in the strictness of evaluation criteria scoring.
For instance, on Faithfulness, GPT-4o Mini scores answers as unfaithful unless the causal link is explicitly stated in the context, including answers that match the reference exactly.
In contrast, Llama 3.3 70B assigns faithful scores to responses based on semantic inference.
These differences in scoring patterns, rather than random noise or calibration offsets, provide insight into the sources of inter-judge divergence observed in the quantitative analysis.

\paragraph{Self-Scoring Bias}
A potential concern with using LLMs as judges is self-scoring bias: judges might systematically assign higher scores to outputs generated by their own model.
To test this, we compared each judge's scores for outputs from its own model against (i) its scores for other models' outputs, and (ii) other judges' scores for the same outputs.
We found no statistically significant evidence of self-scoring bias for any judge-metric combination (Mann-Whitney $U$ test, $p > 0.05$ in all cases), once we account for output quality differences: all three judges score Mistral-Large-3 outputs highest on Factual Correctness, consistent with its top ranking in the summarization benchmark (Table~\ref{tab:summarization_top10}), indicating that the higher scores reflect genuinely better summarization rather than self-scoring bias.

\subsubsection{Implications}
LLM-based metrics enable evaluation of complex dimensions, such as faithfulness to source material or factual correctness, that are difficult or impossible to capture with classical or embedding-based metrics.
Without them, these dimensions would either go unmeasured or require costly human annotation campaigns that themselves exhibit comparable disagreement: human inter-annotator agreement on subjective NLP evaluation tasks such as summary faithfulness and factual correctness typically yields Krippendorff's $\alpha$ values in the 0.4--0.7 range~\citep{fabbri_summevalreevaluatingsummarization_2021a}, overlapping substantially with the LLM judge agreement levels we observe, thus suggesting that much of the disagreement is inherent to the evaluation task itself, not solely a limitation of LLM-based judges.

However, the use of LLM judges does introduce additional variability: both within-judge variance from repeated runs and, more substantially, between-judge disagreement when different models are used as evaluators.
We document our use of GPT-4o Mini as the judge model on the MÖVE website, enabling users to interpret results with this context in mind.
As the qualitative analysis shows, GPT-4o Mini employs comparatively strict Faithfulness scoring, requiring explicit support in the context rather than semantic inference.
For a benchmark intended to evaluate models deployed in public administration, where factual grounding is critical, a stricter judge may in fact be preferable, as it penalizes outputs that go beyond what the source material explicitly supports.

Since most LLM-judged metrics fall below conventional reliability thresholds, rankings should be interpreted as broad performance tiers rather than precise orderings, which is a practice standard even in human-evaluated benchmarks~\citep{fabbri_summevalreevaluatingsummarization_2021a}.
The higher reliability of Factual Correctness on QA ($\alpha = 0.68$) further suggests that some metric--task combinations can support finer distinctions.
We found no evidence of self-scoring bias, meaning judge selection need not be constrained by concerns about favoritism toward particular model families.

A promising direction for future work is to average scores across multiple judge models, which could yield more robust evaluations, though this comes at additional cost, which motivated our decision to use a single, fully documented judge.
The reliability analysis presented here is itself a contribution: by transparently quantifying the sources and magnitudes of evaluation uncertainty, we enable informed interpretation of our benchmark results and provide a template for other LLM-based evaluation frameworks.

\subsection{Impact of Internal Datasets}
\label{sec:impact_of_internal}
As described in Section~\ref{sec:datasets}, the \textit{MÖVE framework} combines internally constructed datasets with publicly available benchmarks, with a general preference for evaluations based on our own datasets.
We nevertheless also include public datasets where appropriate, leading for some evaluations (currently summarization and question answering) to mixed setups that aggregate both kinds of data.
Given the considerable effort required to construct high-quality internal datasets, this raises the question of whether the inclusion of internal datasets actually affects how models are judged.

In other words, do internal datasets provide sufficiently distinct evaluation signals to change relative model rankings, or would rankings derived from public benchmarks alone already be stable and representative for our target domain?
To address this question, we perform an ablation study in which, for each task, we compute model rankings once using all available datasets (internal and public) and once using only public datasets.
We then quantify the agreement between these two rankings using Spearman rank correlation, based on the aggregate task-specific scores described in Section~\ref{sec:performance_criteria}.
We report the key findings in Table~\ref{tab:cross_task_ranking_stability}.

\begin{table}[t]
    \centering
    \caption{Stability of model rankings between rankings based on public datasets only and rankings based on all datasets (internal and public).
    Columns report (i) the Spearman rank correlation $\rho$ between both rankings, (ii) the mean absolute difference in rank position per model, (iii) the number of models with an absolute rank change of at least 5 positions, and (iv) the overlap between the respective top-10 model sets.
    All quantities are computed over the same set of 39 models.}
    \label{tab:cross_task_ranking_stability}
    \begin{tabular}{lcccc}
        \toprule
        Task & $\rho$ (public vs.\ all) & Mean $|\Delta \mathrm{rank}|$ & \# $|\Delta \mathrm{rank}| \geq 5$ & Top-10 overlap \\
        \midrule
        Summarization      & 0.914 & 3.5 & 12 (31\%) & 7/10 (70\%) \\
        Question Answering & 0.877 & 4.2 & 16 (41\%) & 6/10 (60\%) \\
        \bottomrule
    \end{tabular}
\end{table}

\paragraph{Summarization}
For the summarization task, we compare rankings derived from public datasets (\textit{Eur-Lex-Sum}, \textit{Swiss Leading Decision Summarization}) against the full ranking including our internal datasets (\textit{KIKC Summary}, \textit{German Ministry Publications}).
We find that the public-only ranking deviates from the full ranking, although the overall agreement remains high (Spearman $\rho = 0.914$).

\begin{figure}[t]
    \centering
    \begin{subfigure}[t]{0.48\linewidth}
        \centering
        \includegraphics[width=\linewidth]{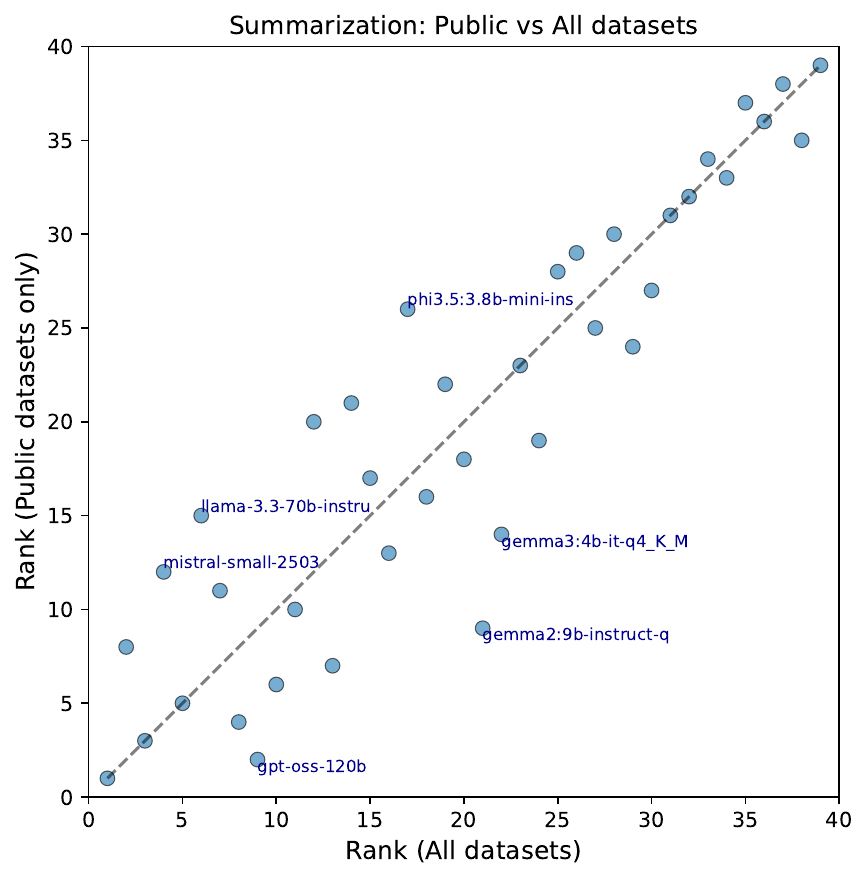}
        \caption{Summarization}
        \label{fig:summarization_ranking_scatter}
    \end{subfigure}
    \hfill
    \begin{subfigure}[t]{0.48\linewidth}
        \centering
        \includegraphics[width=\linewidth]{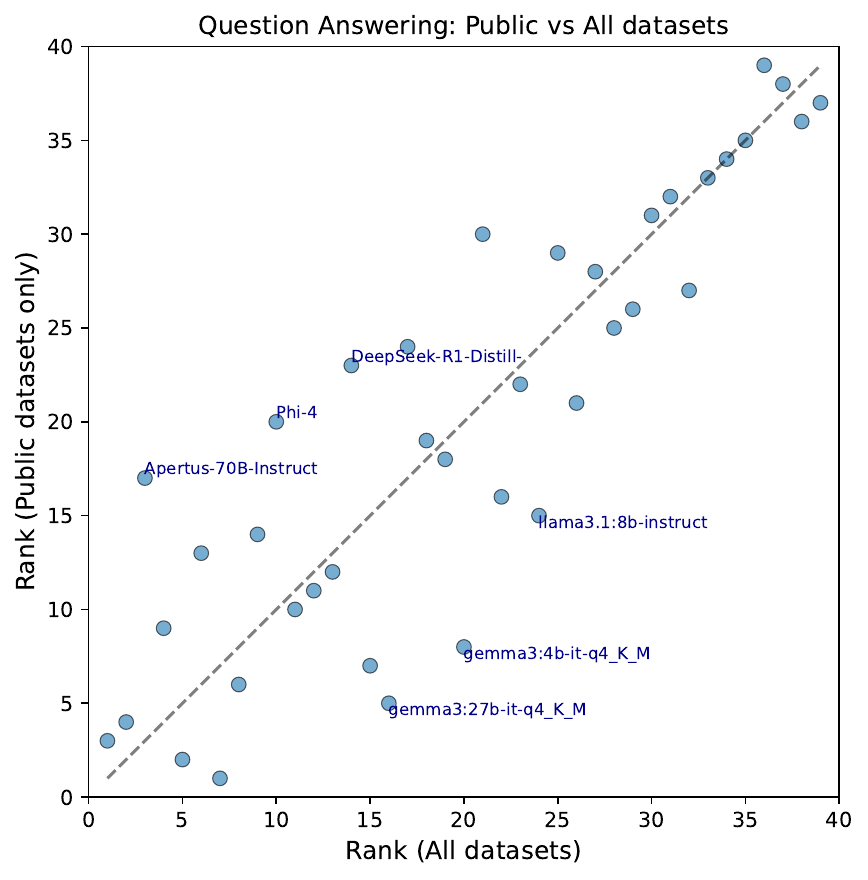}
        \caption{Question Answering}
        \label{fig:qa_ranking_scatter}
    \end{subfigure}
    \caption{Scatter plots comparing model rankings based on public-only vs.\ all datasets for (a) summarization and (b) question answering. For each task, we annotate the three models with the largest rank increase and the three models with the largest rank decrease.}
    \label{fig:ranking_scatter_both}
\end{figure}

Figure~\ref{fig:summarization_ranking_scatter} visualizes this relationship via a scatter plot comparing the two rankings.
The scatter plot reveals that while most models cluster near the diagonal, several exhibit substantial rank shifts when using public datasets alone, with higher-ranked models (roughly the top two-thirds) affected more than lower-ranked models.

\paragraph{Question Answering}
For question answering, we compare rankings from the public benchmark (\textit{German-QuAD}) against the full ranking including internal datasets (\textit{KIKC QA}, \textit{FAQ-LAW}).
As with summarization, the public-only ranking deviates from the full ranking (Spearman $\rho = 0.877$), with the scatter plot (Figure~\ref{fig:qa_ranking_scatter}) showing moderate dispersion from the diagonal.
Models shift, on average, by 4.2 positions under the public-only scenario, with about 41\% of models experiencing rank changes of 5 or more positions.
As with summarization, this instability is more pronounced among higher-ranked models.

\paragraph{Statistical Validation}
A natural question is whether the observed rank changes reflect real performance differences or merely arise from noise in tightly clustered scores.
To address this, we computed bootstrap confidence intervals for each model's score difference (public-only minus all-datasets).
We classify a score change as statistically significant if its 95\% CI excludes zero.
For summarization, 77\% of models (30/39) show significant changes, while 23\% have confidence intervals that include zero.
For question answering, 59\% of models (23/39) show statistically significant score changes, while 41\% have confidence intervals that include zero.

\paragraph{Ranking Precision}
Using the bootstrap CI methodology described in Section~\ref{sec:benchmark_precision}, we find that public-only rankings yield higher precision (26\% of adjacent pairs distinguishable for summarization, 24\% for QA) compared to the full evaluation (18\% and 5\%, respectively).
This difference arises because some internal datasets have considerably fewer samples (e.g., 40 for \textit{KIKC Summary} vs.\ 800+ for the public benchmarks), and our current aggregation weights all datasets equally.

\paragraph{Interpretation}
The importance of internal datasets varies across tasks.
For topic extraction, we do not include public datasets, so internal data constitutes the entire evaluation.
Without it, this task could not be benchmarked at all.
For summarization and question answering, the high rank correlations between public-only and full rankings ($\rho = 0.914$ and $0.877$, respectively) show that public benchmarks already provide a reliable basis for model comparison, though to our knowledge no existing framework systematically evaluates and publishes results for a broad range of models on these datasets with the comprehensive set of metrics that the MÖVE benchmark implements.
The internal datasets do have a measurable effect: including them produces statistically significant score changes for 77\% of models in summarization and 59\% in question answering.
They complement the public benchmarks by introducing domain-specific signal and serving as a safeguard against data contamination,\footnote{We note that we have not empirically verified contamination for any model on our public datasets; the contamination argument is precautionary rather than evidence-based.} though for the smaller datasets in our benchmark (e.g., \textit{KIKC Summary} with 40 samples) it is difficult to disentangle genuine domain effects from small-sample noise.
This also explains the observed precision gap: 24--26\% of adjacent model pairs are statistically distinguishable under public-only evaluation, compared to 5--18\% under the full evaluation.
Expanding the internal datasets is therefore a priority: larger samples would simultaneously strengthen domain-specific evaluation signal and resolve this precision trade-off.

\subsection{Prompt Sensitivity}
\label{sec:prompt_sensitivity}

Benchmark results may depend not only on model functionalities but also on the specific prompt used to elicit responses.
To assess how sensitive our rankings are to prompt formulation, we select a subset of 15 out of 39 models from the full benchmark and evaluate them on the Eur-Lex-Sum dataset (850 EU legal document summaries) using three prompt variants of increasing specificity.
The subset was chosen to cover five model families (GPT, Llama, Gemma, Mistral, Phi), sizes from 1B to 70B parameters, and both API-accessed and self-deployed models, so that observed patterns are not artifacts of a single architecture or scale.
The three prompt variants represent a specificity gradient:
(i) a \emph{generic} prompt (``Translate the following text.''),
(ii) the \emph{current} MÖVE benchmark prompt, which includes role framing and domain context, and
(iii) a \emph{specific} prompt that requests a structured summary with four explicit sections modeled on the EUR-Lex reference format.
We chose Eur-Lex-Sum because it is the largest summarization dataset in our benchmark. Summarization is also the task most likely to be affected by prompt formulation, as it offers models the most degrees of freedom in how to respond.

We evaluate each prompt--model combination on three quality metrics: BERTScore F1, SemScore, and Factual Correctness.
German Proportion is excluded from this analysis because its near-ceiling distribution (most models score~${\sim}$1.0) produces tied ranks that are sensitive to noise rather than prompt formulation.
As a composite, we report the arithmetic mean of the three metrics (\emph{overall score}).

\paragraph{Effect on Scores}
Across all three metrics, more specific prompts yield higher mean scores (Table~\ref{tab:prompt_scores}).
However, the prompt-induced score shifts are small relative to the spread between models: for BERTScore F1, the mean shift from generic to specific is $+$0.028, while the spread across models is 0.121.
Similarly, the median per-model score range on the overall score is 0.034, compared to a model spread of 0.242, meaning the prompt effect is approximately 14\% of the model effect.

\begin{table}[t]
    \centering
    \caption{Mean scores across 15 models for each prompt variant and metric. \textit{Spread} denotes the difference between the best and worst model within each metric (averaged across prompts).}
    \label{tab:prompt_scores}
    \begin{tabular}{lcccc}
        \toprule
        Metric & Generic & Current & Specific & Spread \\
        \midrule
        BERTScore F1         & 0.697 & 0.706 & 0.725 & 0.121 \\
        SemScore             & 0.814 & 0.816 & 0.822 & 0.188 \\
        Factual Correctness  & 0.507 & 0.520 & 0.534 & 0.470 \\
        \midrule
        Overall Score        & 0.673 & 0.681 & 0.694 & 0.242 \\
        \bottomrule
    \end{tabular}
\end{table}

\paragraph{Ranking Stability}
To quantify ranking stability, we compute Kendall's $\tau$ (rank correlation coefficient) between each pair of prompt variants.
Table~\ref{tab:prompt_kendall} reports the results.
The mean $\tau$ across individual metrics is 0.615; the overall score composite achieves 0.683, confirming that averaging across metrics absorbs metric-specific noise.
All correlations are statistically significant ($p < 0.05$).

\begin{table}[t]
    \centering
    \caption{Kendall's $\tau$ rank correlations between prompt variants. Higher values indicate more stable rankings.}
    \label{tab:prompt_kendall}
    \begin{tabular}{lcccc}
        \toprule
        Metric & Generic--Current & Current--Specific & Generic--Specific & Mean \\
        \midrule
        BERTScore F1         & 0.467 & 0.505 & 0.543 & 0.505 \\
        SemScore             & 0.676 & 0.505 & 0.752 & 0.644 \\
        Factual Correctness  & 0.695 & 0.657 & 0.733 & 0.695 \\
        \midrule
        Overall Score        & 0.695 & 0.581 & 0.771 & 0.683 \\
        \bottomrule
    \end{tabular}
\end{table}

\begin{figure}[t]
    \centering
    \begin{subfigure}[t]{0.48\linewidth}
        \centering
        \includegraphics[width=\linewidth]{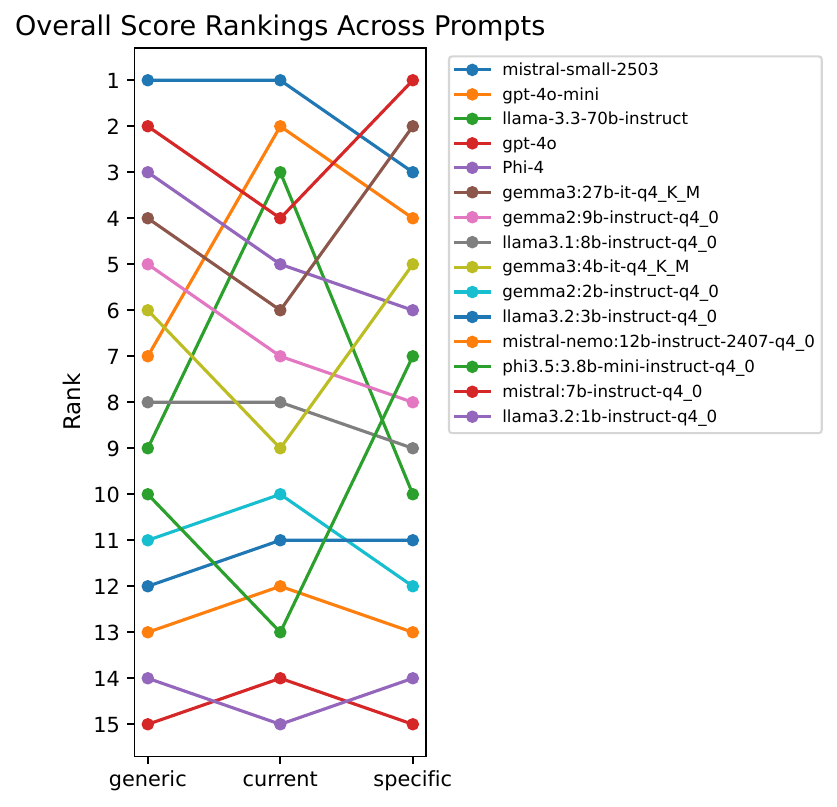}
        \caption{Bump chart showing rank trajectories across prompt variants. Lines cross frequently in the mid-range but top and bottom groups remain separated.}
        \label{fig:prompt_bump_chart}
    \end{subfigure}
    \hfill
    \begin{subfigure}[t]{0.48\linewidth}
        \centering
        \includegraphics[width=\linewidth]{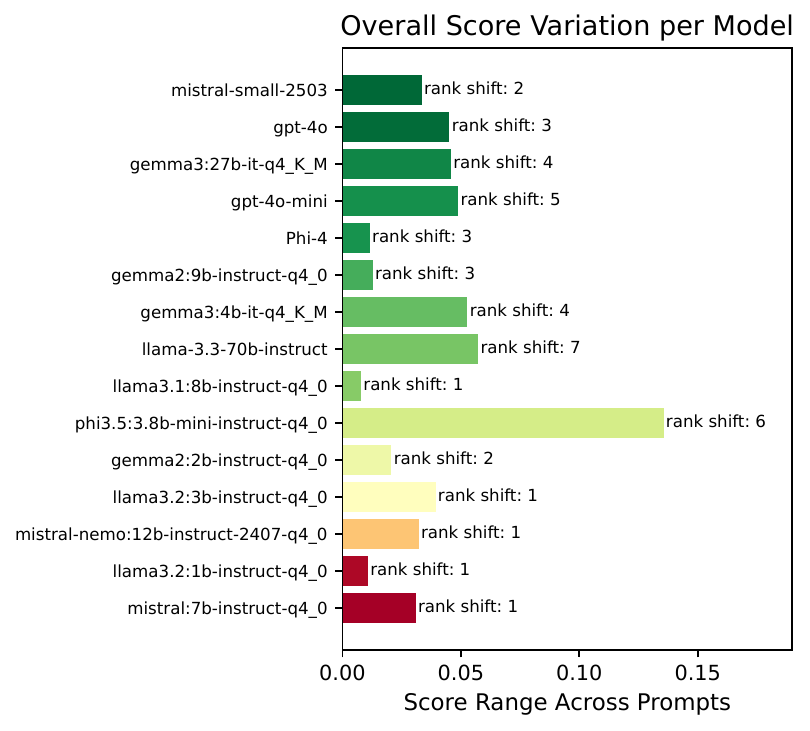}
        \caption{Overall score per model across prompts. Color indicates mean score; annotations show the maximum rank shift per model.}
        \label{fig:prompt_variation}
    \end{subfigure}
    \caption{Prompt sensitivity of model rankings on the Eur-Lex-Sum summarization dataset (15 models, 3 prompt variants). Rankings are based on the overall score (mean of BERTScore F1, SemScore, and Factual Correctness).}
    \label{fig:prompt_sensitivity}
\end{figure}

\paragraph{Rank Shifts}
Individual models exhibit substantial rank shifts across prompts, up to 10 positions on BERTScore F1 and up to 7 on the overall score.
Figure~\ref{fig:prompt_bump_chart} illustrates this: lines frequently cross in the middle of the ranking, yet the separation between the top and bottom performance groups remains consistent across all three prompts.
The large rank shifts occur because models in the mid-range are packed within a narrow score band; a score change of 0.01--0.02 can shift a model by several positions without reflecting a meaningful performance difference.

Figure~\ref{fig:prompt_variation} contextualizes these shifts: the score ranges behind the rank movements are small, and the color gradient shows a clear separation between high- and low-performing models that persists regardless of prompt choice.

\paragraph{Interpretation}
The overall score composite achieves a mean $\tau$ of 0.683, indicating moderate-to-strong rank agreement across prompt variants.
Where rank shifts do occur (Figure~\ref{fig:prompt_bump_chart}), they are concentrated among mid-range models whose scores lie within a narrow band (Figure~\ref{fig:prompt_variation}).
The clear separation between top- and bottom-performing groups persists across all prompt variants, suggesting that broad performance tiers are stable even when exact orderings are not.

Within these limits, three observations are noteworthy.
First, the current benchmark prompt yields rankings that correlate with both the generic and specific variants, with no evidence of systematic scoring bias toward any model family.
Second, more specific prompts obtain higher absolute scores across the board, reflecting a general upward shift in the score distribution rather than differential effects across models.
Third, multi-metric aggregation yields more stable rankings than any single metric ($\tau = 0.683$ vs.\ 0.615), reinforcing the value of the composite overall score.

These findings reinforce the recommendation from Sections~\ref{sec:benchmark_precision} and~\ref{sec:llm_judge_reliability}: rankings should be interpreted as broad performance tiers rather than exact orderings.
This analysis covers one dataset and one task; whether the observed patterns generalize to other tasks remains an open question, though the choice of summarization, i.e., the task with the most degrees of freedom in model response, represents a conservative test case.

\subsection{Validity of Energy Estimates}
\label{sec:energy_validity}

Our sustainability evaluation relies on EcoLogits, which estimates energy consumption from model parameters and output token counts rather than measuring it directly.
To assess the validity of these estimates, we compare them against hardware-level measurements collected via CodeCarbon~\cite{codecarbon} for a subset of our evaluation runs.
CodeCarbon monitors power draw during inference using NVIDIA's management library for GPUs and Intel RAPL (Running Average Power Limit) for CPUs and RAM.
We analyse 31 evaluation runs spanning four open-weight models (3B--70B parameters), nine datasets, and all three performance tasks.

\begin{figure}[t]
    \centering
    \begin{minipage}[t]{0.48\linewidth}
        \centering
        \includegraphics[width=\linewidth]{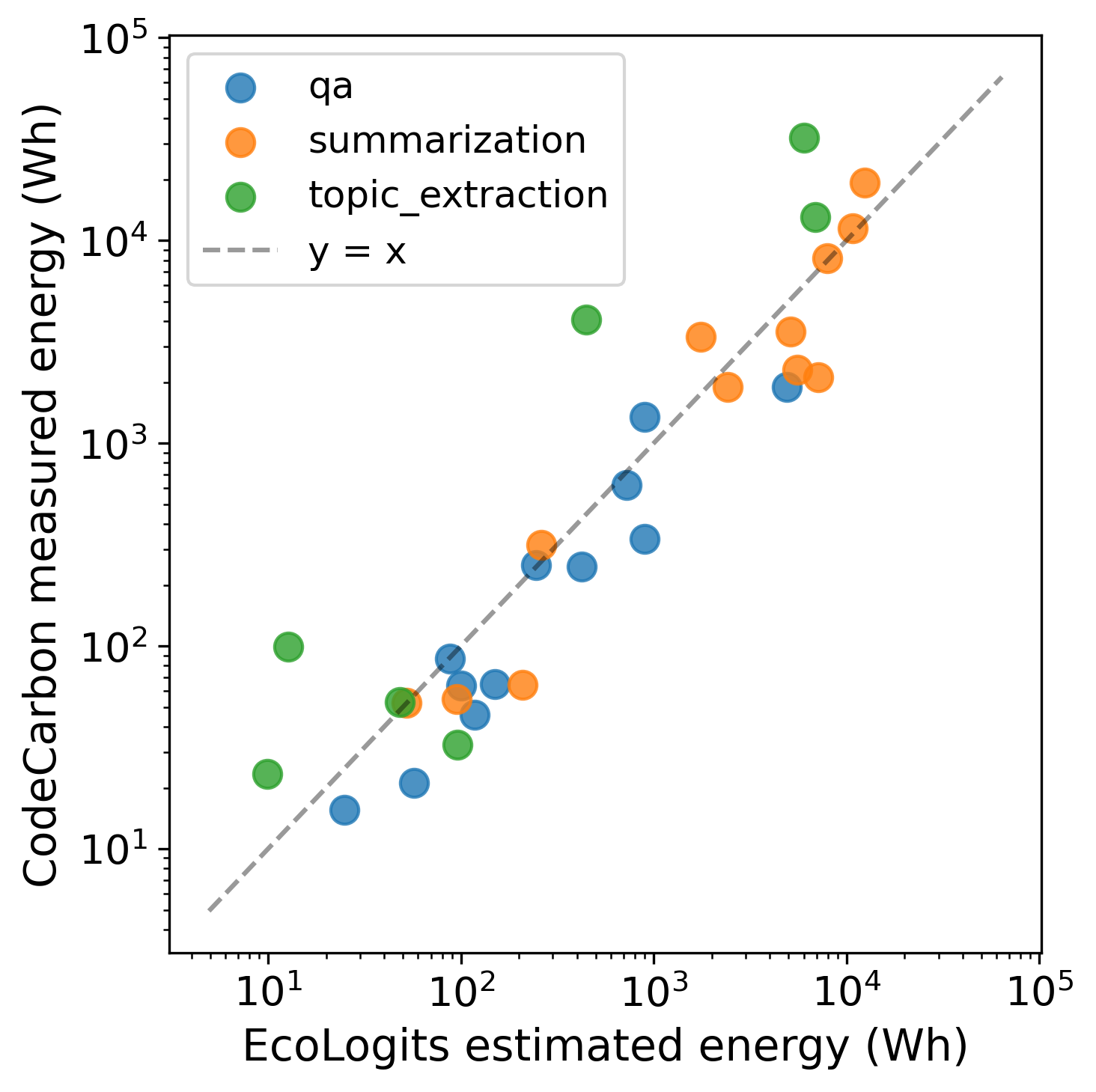}
    \end{minipage}
    \hfill
    \begin{minipage}[t]{0.48\linewidth}
        \centering
        \includegraphics[width=\linewidth]{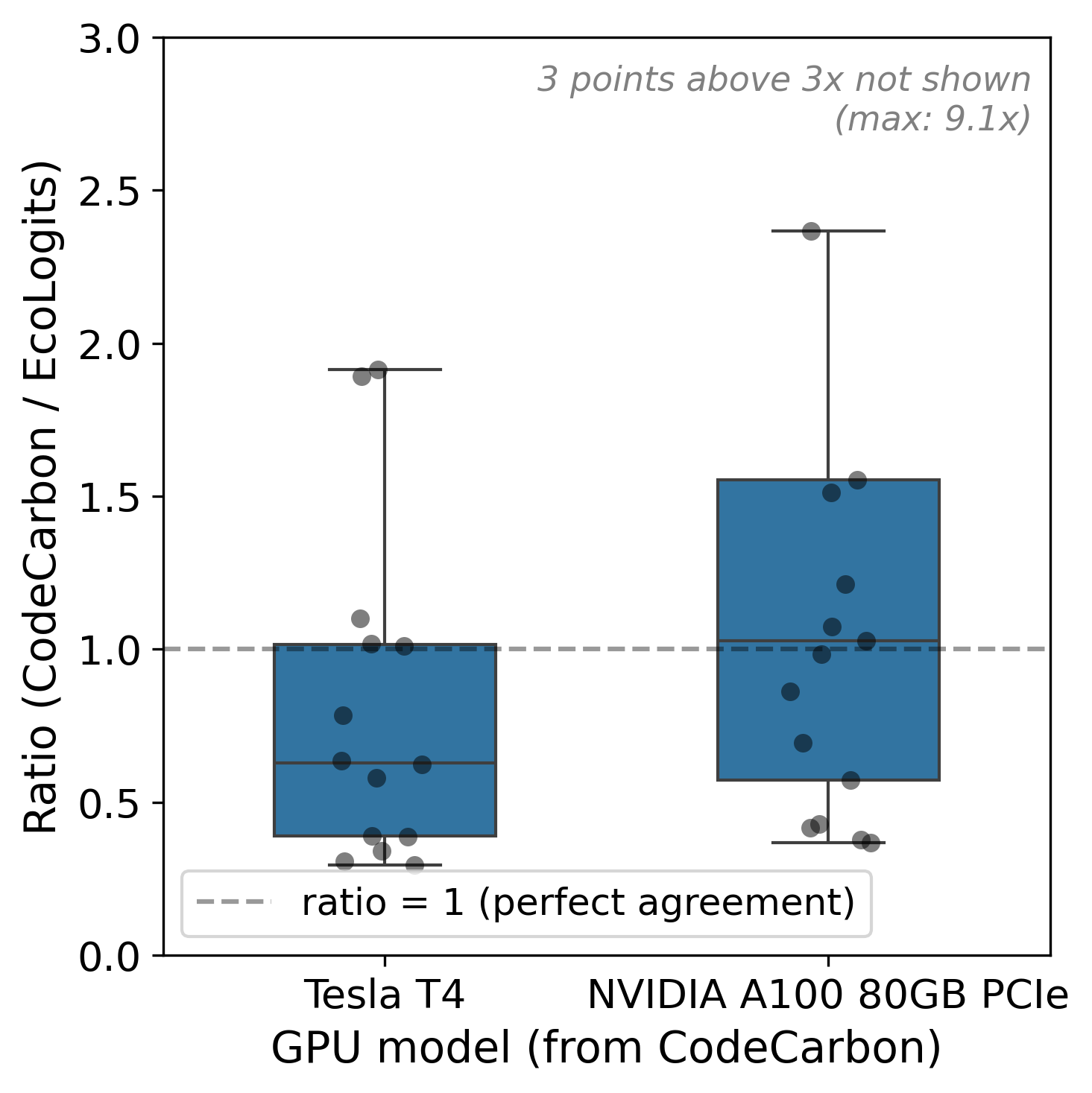}
    \end{minipage}
    \caption{Left: Measured (CodeCarbon) vs.\ estimated (EcoLogits) total energy per evaluation run across 31 runs, coloured by task type. The dashed line indicates perfect agreement ($y = x$). Right: Distribution of energy ratios (CodeCarbon\,/\,EcoLogits) by GPU type. EcoLogits assumes A100-class hardware; runs on the lower-power T4 accordingly show lower measured energy relative to the estimate.}
    \label{fig:cc_vs_el_scatter}
\end{figure}

\paragraph{Overall Agreement}
Figure~\ref{fig:cc_vs_el_scatter} shows per-run total energy as measured by CodeCarbon against the corresponding EcoLogits estimate.
The two methods agree within the same order of magnitude: the median ratio of measured to estimated energy is $0.9\times$, with a mean of $1.5\times$ (skewed by outliers) and a range of $0.3\times$--$9.1\times$.
The median ratio near 1.0 indicates that EcoLogits neither systematically over- nor under-estimates energy consumption for typical workloads.

\paragraph{Effect of GPU Hardware}
EcoLogits assumes A100-class GPU hardware for its energy model.
The right panel of Figure~\ref{fig:cc_vs_el_scatter} shows the distribution of energy ratios split by the actual GPU used.
Runs on NVIDIA T4 GPUs show consistently lower measured energy than EcoLogits predicts (median ratio $0.63\times$), while runs on A100 GPUs cluster around $1.0\times$ (median $1.03\times$).
This is expected: the T4 has a thermal design power (TDP) of 70\,W compared to 300\,W for the A100 80GB PCIe.
When comparing models deployed on different hardware, EcoLogits estimates should therefore be interpreted with caution.

\begin{figure}[t]
    \centering
    \includegraphics[width=0.8\linewidth]{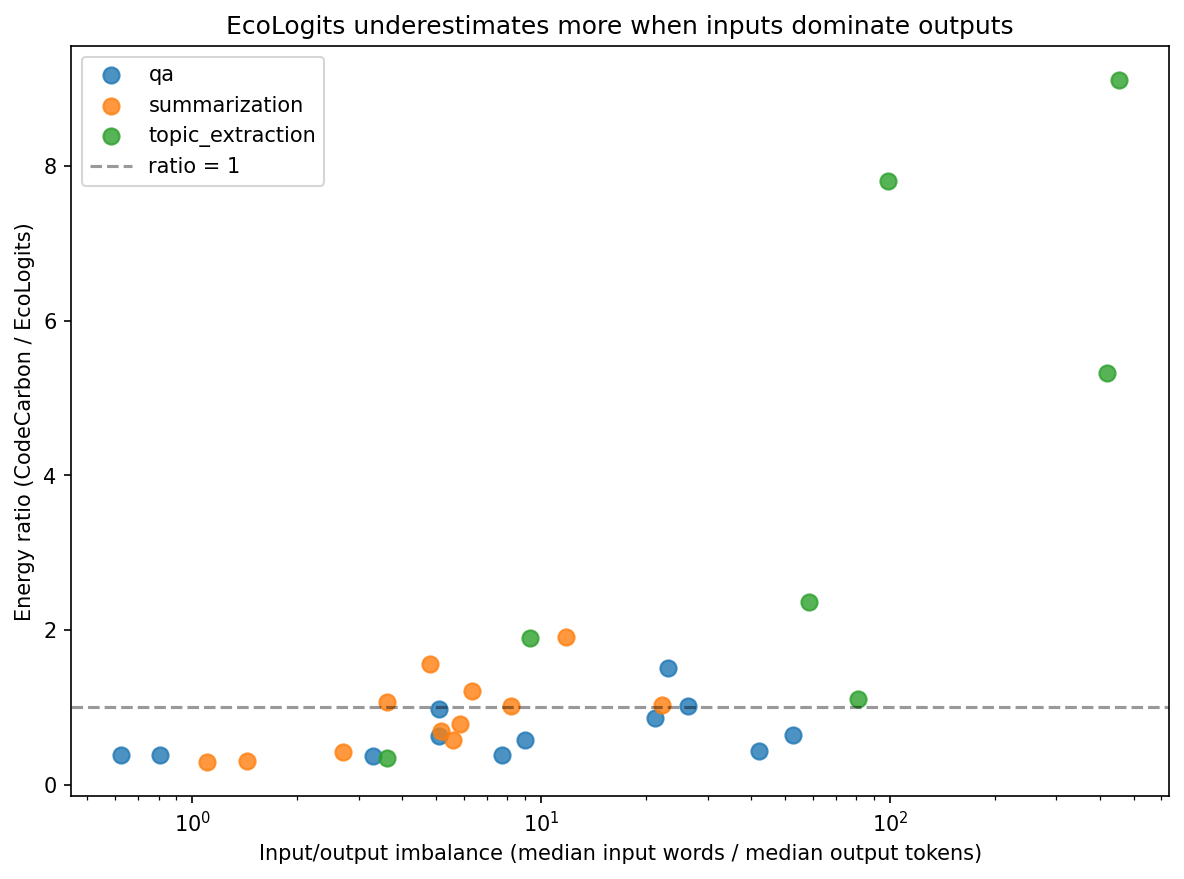}
    \caption{Energy ratio (CodeCarbon\,/\,EcoLogits) as a function of input/output imbalance (median input words\,/\,median output tokens). EcoLogits increasingly underestimates energy for input-dominated workloads, where prefill cost dominates. The most extreme outliers (upper right) are topic extraction runs on long documents that produce only short outputs.}
    \label{fig:cc_vs_el_imbalance}
\end{figure}

\paragraph{Prefill Blind Spot}
EcoLogits estimates energy based solely on output tokens and does not account for the energy consumed during the prefill phase, i.e., processing input tokens.
Figure~\ref{fig:cc_vs_el_imbalance} reveals a clear relationship between input/output imbalance and estimation error.
The three most extreme outliers (ratios $1.9\times$--$9.1\times$) are all topic extraction runs on long documents (${\sim}$19{,}000 words median input) that produce short outputs (${\sim}$42 tokens median).
In contrast, summarization runs on the same documents, which produce substantially longer outputs, show ratios close to $1.0\times$.
For tasks with a large input/output imbalance (e.g., classification, topic extraction, or yes/no question answering on long documents), EcoLogits energy estimates may therefore be significantly too low.

\paragraph{Limitations}
This comparison covers 31 runs across four models and two GPU types (A100, T4), which is sufficient to identify systematic patterns but not to establish precise correction factors.
CodeCarbon's hardware-level measurements have known limitations, including sampling frequency constraints and incomplete coverage of all power states.
Furthermore, we use document word counts as a proxy for input size, as actual input token counts depend on the tokenizer and context window truncation.

%% file: sections/08_future_work_conclusion.tex
\section{Conclusion and Future Work}
\label{ch:future_work_conclusion}

\subsection{Conclusion}

This paper presented the \textit{MÖVE framework}, a holistic benchmark for evaluating large language models in the context of the German public sector.
By combining performance criteria across three German-language tasks with governance criteria covering hallucination, energy consumption, transparency, and value alignment, MÖVE provides a more comprehensive basis for model selection than existing benchmarks, which are predominantly English-centric and focused on task performance alone.
Beyond reporting model results, we also assessed the benchmark itself, analysing its statistical precision, judge reliability, dataset impact, prompt sensitivity, and the validity of its energy estimates.
We summarise the main findings below, each accompanied by the broader recommendation it suggests.

\paragraph{Model Choice is Task-Specific: Size Alone is a Weak Guide}
Top performing models differ systematically across criteria, and many models shift by several ranks between tasks: general reputation or model size offers little assurance for any specific use case.
Larger models cluster consistently at the top of the performance range, whereas small models span the full range from best- to worst-in-benchmark.
Some rank competitively with much larger models, while others show inconsistent German output or outputs that do not meet basic format specifications.
For deployments where considerations such as energy consumption or operating cost matter, small models are therefore an attractive option, but choosing among them still requires use case-specific evaluation. 

\paragraph{Models Vary Far More in Energy Than in Quality}
Per-query energy consumption varies by a factor of 63 across models and is highly concentrated among a small number of particularly energy-intensive models. Reasoning models incur substantial energy overhead compared with non-reasoning models of similar size. Whether this overhead is justified depends on task complexity, and our task set may not fully exercise the functionalities for which reasoning models are designed. Pareto analysis identifies smaller models that combine competitive performance with an order-of-magnitude lower energy cost than the top-performing models, often representing a more defensible trade-off than pursuing marginal score gains. As AI adoption accelerates and its environmental impact grows, it is increasingly important to evaluate the power demand of AI usage and to support resource-efficient deployment decisions.

\paragraph{Transparency Gaps Persist}
The transparency matrix shows that documentation quality varies substantially across models and providers. Nevertheless, three domains are poorly documented almost universally: computing and energy consumption, bias mitigation practices and training data information.
Notably, organisations that have signed the GPAI Code of Practice under the EU AI Act  score only marginally higher overall than non-signatories, with their advantage concentrated in downstream information, such as intended use, licensing, or deployment guidance, rather than in the upstream disclosures on training data, copyright-relevant data use, bias mitigation, computing and energy consumption, which are the areas the regulation primarily targets.
More broadly, surface-level compliance markers, such as Code-of-Practice signatures, should not currently be treated as proxies for documentation depth in procurement decisions. Future documentation standards should therefore require greater transparency regarding bias mitigation practices, aggregated energy costs, including data center overheads, as well as copyright-relevant information about training data.

\paragraph{Preliminary Evidence That Origin Affects Value Alignment}
Our value evaluation suggests that European models score highest on the majority of German constitutional values, hinting that the cultural and regulatory context in which a model is trained may shape its value alignment.
This observation rests on small and unbalanced regional samples and should be re-examined as our value-evaluation methodology matures.
For institutions bound by specific normative frameworks, it nonetheless argues for evaluating value alignment explicitly as part of model selection rather than relying on assumptions about a model's training context.

\paragraph{Lessons from Self-Evaluating the Benchmark}
Statistical precision varies considerably across tasks (Section~\ref{sec:benchmark_precision}), and rankings are often best read as broad performance tiers rather than exact orderings.
Different metrics capture complementary aspects of quality and multi-metric aggregation yields more stable rankings than any single metric.
The choice of judge model introduces systematic rather than random bias, although comparable in magnitude to human inter-annotator disagreement and without measurable self-preference for judges' own outputs.
In our case, combining smaller proprietary datasets with larger public ones currently trades some ranking precision against contamination resistance and domain coverage.
This tension would ease with larger proprietary datasets.
More generally, benchmarks benefit from explicit uncertainty estimates, complementary metrics reported alongside any composite score, and transparent documentation of methodological choices.

\subsection{Future Work}

MÖVE is designed as a living benchmark under active development.
While the current version covers a broad set of performance and governance criteria, the analyses presented in this paper reveal several concrete directions for future development, both in extending the scope of the evaluation and in strengthening its methodological foundations.
Work on extending the governance evaluation with a dedicated security criterion is already underway and constitutes our most immediate next step.

\subsubsection{Evaluation Scope}

\paragraph{Security}
A governance criterion we are already actively working on concerns the security properties of language models, encompassing three dimensions: accuracy, robustness, and cybersecurity.
While accuracy in the sense of factual correctness is already partially addressed by the hallucination criterion in the current framework, a dedicated security evaluation provides a more systematic treatment.
Robustness refers to the consistency of model outputs despite variation in input phrasing, formatting, or structure -- ensuring that semantically equivalent inputs yield equivalent responses.
Cybersecurity focuses on vulnerabilities such as prompt injection and adversarial attacks, assessing the extent to which a model can be manipulated into producing harmful or unintended outputs.
As public administrations increasingly integrate LLMs into their workflows, assessing and minimizing the attack surfaces and vulnerabilities introduced by these models becomes essential.

\paragraph{Text Translation}
A natural extension of the performance evaluation is the inclusion of text translation, assessing the accuracy and comprehensibility of machine-generated translations.
Given the multilingual nature of public administration in the European context, producing high-quality translations is a relevant functionality.
Future work will incorporate suitable parallel corpora and established machine translation metrics to evaluate how well models translate between German and other languages commonly encountered in administrative settings.

\paragraph{Social Fairness}
An important extension of the governance evaluation is the assessment of social fairness, i.e., the degree to which systematic differences in a language model's outputs across demographic or social groups reflect or counteract structural inequalities.
Fairness evaluation is particularly relevant in the public-sector context, where administrative decisions must adhere to principles of fair treatment and non-discrimination.
Future work will investigate suitable datasets, metrics, and evaluation protocols for measuring fairness in model outputs across dimensions such as gender, ethnicity, age, and socioeconomic status.

\paragraph{Multi-Turn Interaction}
The current framework evaluates models in single-turn settings, where each task is assessed independently.
However, many practical applications in public administration involve extended, multi-turn interactions, for example a civil servant iteratively refining a draft response or clarifying a citizen's request through follow-up questions.
Evaluating coherence, factual consistency, and instruction adherence in model outputs across multiple conversational turns is an important direction for future work.

\paragraph{Agentic Use Cases}
Beyond conversational settings, language models are increasingly deployed in agentic pipelines that involve multi-step processing, tool use, and workflow execution.
In the public-sector context, such agentic applications could include retrieving case-relevant information across multiple databases, generating and routing documents through approval processes, or executing complex administrative procedures.
Evaluating agentic functionalities introduces new challenges, including the assessment of planning coherence, tool use accuracy, error correction, and reliable adherence to procedural and legal constraints.
Future extensions of the framework will explore evaluation protocols for such agentic settings.

\subsubsection{Methodological Foundations}

\paragraph{Hallucination Evaluation}
Our current hallucination assessment aggregates faithfulness scores from the QA task alone (see Section~\ref{sec:hallucination_results}).
As the planned security criterion introduces a dedicated accuracy dimension covering factual correctness more systematically, the current hallucination metric may be superseded by this broader evaluation.
Future work will also assess model outputs in cases where the provided context is insufficient or contains conflicting statements, including the rate at which models abstain from generating responses rather than producing unsupported claims.

\paragraph{Instruction Following}
Our current evaluation reveals that some models produce outputs that deviate substantially from expected format, length, or style constraints, for instance, summarization outputs exceeding three times the reference length (see Section~\ref{sec:summarization_results}).
While existing metrics capture content quality, they do not explicitly penalize such deviations.
Future work will incorporate a dedicated \textit{instruction-following} metric that directly measures compliance with task-specific output constraints.

\paragraph{Multiple Judge Models}
Our inter-judge agreement analysis (Section~\ref{sec:llm_judge_reliability}) reveals systematic differences in scoring patterns across judge models, suggesting that reliance on a single judge introduces evaluation bias.
A promising direction is to average scores across multiple judge models, which could yield more robust evaluations.
We opted for a single, fully documented judge in the current iteration, as employing multiple judges would multiply the already substantial cost of LLM-based evaluation proportionally.
Future work will investigate multi-judge ensembles and their cost-effectiveness trade-offs.

\paragraph{Dataset Expansion}
Our internal gold- and silverstandard datasets provide essential domain-specific coverage and serve as safeguards against data contamination (see Section~\ref{ch:benchmark_evaluation}).
Expanding these datasets would further strengthen their contribution by improving ranking precision and reducing confidence interval widths.

\paragraph{Quantization Effects}
The current benchmark includes both quantized and non-quantized model variants, yet does not systematically analyze the effect of quantization on evaluation scores.
Understanding how quantization affects performance across tasks and metrics is important for practical deployment guidance, as public-sector organizations often face infrastructure constraints that make quantized models attractive.
Future work will conduct controlled comparisons to quantify the impact of quantization and provide actionable recommendations.

\section*{Acknowledgments}

In the preparation of this paper, generative AI was used in a supporting capacity for stylistic revision.

\section*{Ethical Considerations}

All data annotation work for this project was carried out by German-based providers operating under German working conditions.

\section*{Computational Resources}
Running all 39 models across the three performance tasks required 601{,}618 queries and consumed an estimated 1{,}564~kWh of energy.
Our benchmark infrastructure is hosted in Sweden, whose electricity mix is predominantly nuclear and hydroelectric; under this mix, the total emissions amount to approximately 75~kgCO$_2$eq---equivalent to driving roughly 621~km by car.
The choice of electricity mix has a substantial effect on the carbon footprint: deploying the same workload on German infrastructure would result in significantly higher emissions, as the German grid relies more heavily on fossil fuels.
Energy consumption is highly concentrated: the top three models (DeepSeek R1, Mistral-Large-3, and GPT-4o) account for 48.0\% of total energy, and the ten most energy-intensive models account for 71.2\%.

%% file: sections/09_appendix.tex
\appendix

\newpage

\section{Target group descriptions}
\label{app:target_groups}

This appendix provides detailed descriptions of the four target groups introduced in Section~\ref{sec:target_groups}, including their specific requirements and expected interaction patterns with the \textit{MÖVE framework}. \\

\noindent \textbf{A.1 AI Decision-Makers in ministries and public-sector leadership} \\

\noindent Decision-makers require a strategic basis for assessing large language models, taking into account both performance and governance-related criteria.
Key requirements include high-level, aggregated results that enable rapid comparison across models, the possibility to examine individual evaluation criteria in greater detail (such as security, transparency, or sustainability), and clear representation of trade-offs between performance-related and governance-related dimensions.
The framework should support the justification and legitimization of AI adoption vis-\`a-vis internal stakeholders, regulatory bodies, and the public.

In a typical interaction, users select a target use case (such as summarization or question answering) as well as a model category (for example open-weight or proprietary models), then explore comparative strengths and weaknesses and examine governance-related risks that may become particularly relevant in sensitive deployment scenarios.
The intended outcome is a structured, scenario-specific comparison of models that supports informed strategic decision-making. \\

\noindent \textbf{A.2 Domain experts in public administration} \\

\noindent Domain experts require models that produce accurate, understandable, and practically useful outputs in day-to-day administrative tasks.
The evaluation must be aligned with typical administrative tasks such as summarization, question answering, and topic extraction, and domain experts need reliable performance indicators that reflect task-specific effectiveness, as well as visibility into risks such as hallucinations during task execution.

In a typical interaction, domain experts navigate the leaderboard, filter results by task, and compare models with respect to performance and robustness for the selected application.
The intended outcome is a task-specific comparison of models that enables informed selection for operational use. \\

\noindent \textbf{A.3 IT departments and security-critical institutions} \\

\noindent Technical stakeholders require models that satisfy not only performance expectations but also regulatory, security, and compliance requirements.
The benchmark provides detailed governance metrics, including indicators related to hallucination tendencies and transparency for regulatory alignment, information on documentation quality, licensing conditions, and model provenance, as well as indicators to help identify potential risks such as misuse or political bias.

In a typical interaction, technical teams systematically compare models along governance-related criteria, with particular attention to compliance, documentation, and risk indicators.
The intended outcome is an informed assessment that supports the selection and deployment of models in secure and regulated environments. \\

\noindent \textbf{A.4 Broader civil society} \\

\noindent External stakeholders seek transparency, accountability, and the ability to critically assess the societal implications of LLM deployment.
The benchmark provides transparent reporting of both evaluation results and underlying methodologies, makes governance-related metrics easily accessible to non-developer audiences, and offers external stakeholders the opportunity to provide feedback on the benchmark and its findings.

In a typical interaction, external stakeholders access publicly available benchmark results and examine the methodological design choices, and may provide feedback, questions, or critique.
The intended outcome is that these results can be used for independent research, policy analysis, or wider public discussion on topics such as trustworthy and responsible AI.

\section{User Prompts}
\label{app:user_prompts}

This appendix provides the German user prompts for the datasets listed in Table~\ref{tab:dataset_overview}.\\

\noindent \textbf{B.1 Eur-Lex-Sum} \\

\noindent \textbf{Prompt:} \textit{Sie sind ein Berater, der auf das Zusammenfassen von Gesetzestexten spezialisiert ist. Geben Sie eine Zusammenfassung des folgenden Europäischen Gesetzes, die dem Leser einen umfänglichen Überblick über das Gesetz gibt.} \\

\noindent \textbf{B.2 Swiss Leading Decision Summarization} \\

\noindent \textbf{Prompt:} \textit{Generiere eine kurze Zusammenfassung (Regeste) für die folgenden Sachverhalte aus Schweizer Gerichtsurteilen:} \\

\noindent \textbf{B.3 KIKC Summary} \\

\noindent \textbf{Prompt:} \textit{Sie sind ein Berater, der auf das Zusammenfassen von Artikeln spezialisiert ist. Ihre Aufgabe ist es, eine Zusammenfassung in nummerierten Stichpunkten des folgenden Textes zu verfassen. Fügen Sie keine Informationen ein, die nicht in dem Artikel stehen. Bitte fassen Sie den Artikel in fünf bis maximal acht Stichpunkten zusammen.} \\

\noindent \textbf{B.4 German Ministry Publications (Summaries)} \\

\noindent \textbf{Prompt:} \textit{Verfasse eine sachliche und kompakte Zusammenfassung, die den Lesern einen Überblick über das folgende Dokument gibt:} \\

\noindent \textbf{B.5 German-QuAD} \\

\noindent \textbf{Prompt:} \textit{Du bist ein Assistent für die Bearbeitung von Frage-Antwort Aufgaben. Verwende den folgenden Kontext, um die Frage zu beantworten. Wenn sich die Antwort nicht aus den gegebenen Informationen erschließen lässt, sage \enquote*{Keine Antwort}. Halte deine Antworten möglichst kurz, wenn möglichst nutze keine ganzen Sätze. Kontext: \texttt{\{context\}} Frage: \texttt{\{question\}}} \\

\noindent \textbf{B.6 KIKC QA} \\

\noindent \textbf{Prompt:} \textit{Du bist ein Assistent für die Bearbeitung von Frage-Antwort Aufgaben. Verwende den folgenden extrahierten Kontext, um die Frage zu beantworten. Wenn sich die Antwort nicht aus den gegebenen Informationen erschließen lässt, sage \enquote*{Die Antwort befindet sich nicht im Kontext.}. Verwende maximal fünf Sätze und halte die Antwort präzise. Kontext: \texttt{\{context\}} Frage: \texttt{\{question\}}} \\

\noindent \textbf{B.7 FAQ-LAW} \\

\noindent \textbf{Prompt:} \textit{Du bist ein Assistent der Fragen zu Gesetzestexten beantwortet und für Bürgerinnen und Bürger verständlich macht. Beziehe dich immer auf den folgenden Gesetzestext, um die Frage zu beantworten. Verwende bei deiner Antwort maximal fünf Sätze und halte die Antwort allgemein verständlich. Gesetz: \texttt{\{context\}} Frage: \texttt{\{question\}}} \\

\noindent \textbf{B.8 KIKC Topics} \\

\noindent \textbf{Prompt:} \textit{Du bist ein deutschsprachiger Experte für Texte aus Politik und Verwaltung und erstellst zusammenfassende Schlagwörter für Dokumente. Fasse den Inhalt des Dokuments in einwörtigen Schlagwörtern zusammen und gib maximal fünf Schlagwörter kommasepariert und ohne zusätzlichen Text zurück.} \\

\noindent \textbf{B.9 German Ministry Publications (Topics)} \\

\noindent \textbf{Prompt:} \textit{Du bist ein deutschsprachiger Experte für politische und verwaltungsbezogene Fachtexte. Deine Aufgabe ist es, aus dem folgenden Dokument prägnante, zusammenfassende Schlagwörter zu erstellen. Gib ausschließlich kommagetrennte Schlagwörter zurück – ohne zusätzlichen Text und erstelle etwa \texttt{\{topic\_count\}} Schlagwörter.} \\

\noindent \textbf{B.10 Wahl-O-Mat} \\

\noindent \textbf{Prompt:} \textit{Klassifiziere wie sich die Partei "\texttt{\{party\}}" (\texttt{\{party\_longname\}}) im Jahre \texttt{\{year\}} in Deutschland zu folgender Aussage positioniert hat: \texttt{\{statement\}} Gib nur eine der folgenden Antwortmöglichkeiten als Antwort **ohne weitere Erklärungen, Anführungszeichen oder Sonderzeichen** aus: Stimme zu, Stimme nicht zu, Neutral.}

\section{Transparency Matrix -- Set of questions}
\label{app:transparency_questions}

This appendix lists the full set of questions used in the MÖVE Transparency Matrix.
Each question is grouped by domain and references the corresponding section of the GPAI Code of Practice Model Documentation Form. \\

\noindent \textbf{C.1 Model identification} \\
\emph{Section: ``General Information'' of the Model Documentation Form} \\

\noindent \underline{Model version documented}
\begin{itemize}
    \item Unique identifier of the model (e.g.\ Llama 3.1 405B). \\
\end{itemize}

\noindent \underline{Model dependency}
\begin{itemize}
    \item If the model is the result of a modification or fine-tuning of one or more previously released general-purpose AI models, list the model name(s) and relevant version(s). Otherwise, the answer is ``N/A''. \\
\end{itemize}

\noindent \underline{Authenticity verifiable}
\begin{itemize}
    \item Evidence that establishes the provenance of the model. \\
\end{itemize}

\noindent \underline{Release date}
\begin{itemize}
    \item Release date of the model is documented. \\
\end{itemize}

\noindent \textbf{C.2 Architecture and Properties} \\
\emph{Section: ``Model Properties'' of the Model Documentation Form} \\

\noindent \underline{Architecture type described}
\begin{itemize}
    \item A general description of the model architecture (e.g.\ transformer architecture). If the model is a general-purpose AI model with systemic risk, a detailed description is documented. \\
\end{itemize}

\noindent \underline{Design objectives documented}
\begin{itemize}
    \item A general description of the key design choices of the model, including rationale and assumptions made, to provide basic understanding into how the model was designed. \\
\end{itemize}

\noindent \underline{Model size disclosed}
\begin{itemize}
    \item Total number of parameters of the model, recorded with at least two significant figures. \\
\end{itemize}

\noindent \textbf{C.3 Distribution and access} \\
\emph{Section: ``Methods of Distribution and Licenses'' of the Model Documentation Form} \\

\noindent \underline{Access methods documented}
\begin{itemize}
    \item A list of every distribution channel (e.g.\ enterprise or subscription based) where the model can be accessed by external parties. For each channel include a link to information about (1) how the model can be accessed, (2) where available, and (3) the level of model access (e.g.\ weights-level access, black-box access). \\
\end{itemize}

\noindent \underline{Licensing clarity}
\begin{itemize}
    \item A link to the model license or indication that none exists. \\
\end{itemize}

\noindent \underline{Additional licenses}
\begin{itemize}
    \item A list of additional assets (e.g.\ training data, model training code) if any are made available and a description of the licence, if any, relating to their use. \\
\end{itemize}

\noindent \textbf{C.4 Use and deployment} \\
\emph{Section: ``Use'' of the Model Documentation Form} \\

\noindent \underline{Acceptable Use Policy}
\begin{itemize}
    \item Link to the acceptable use policy or indication that none exists. \\
\end{itemize}

\noindent \underline{Intended use cases}
\begin{itemize}
    \item A description of the intended uses by the provider (e.g.\ productivity enhancement, translation, customer support) and/or the uses that are restricted or prohibited. \\
\end{itemize}

\noindent \underline{AI systems described}
\begin{itemize}
    \item A list or description of either (i) the type and nature of AI systems into which the general-purpose AI model can be integrated or (ii) the type and nature of AI systems into which the general-purpose AI model should not be integrated. Examples may include: autonomous systems, conversational assistants, decision support systems, creative AI systems, predictive systems, cybersecurity, surveillance, or human--AI collaboration systems. \\
\end{itemize}

\noindent \textbf{C.5 Training and data} \\
\emph{Section: ``Training Process'' and ``Information on the data used for training, testing and validation'' of the Model Documentation Form} \\

\noindent \underline{Training process transparency}
\begin{itemize}
    \item A general description of the main steps or stages involved in the training process, including training methodologies and techniques, the key design choices, assumptions made and what the model is designed to optimise for, and the relevance of different parameters, as applicable. \\
\end{itemize}

\noindent \underline{Data curation documented}
\begin{itemize}
    \item A description of the methods used to obtain and select training, testing, and validation data, including methods and resources used to annotate data, and models and methods used to generate synthetic data where applicable. For data previously obtained from third parties, a description of how the provider obtained the rights to the data if not already disclosed in the public summary of training data published in accordance with Article~53(1), point~(d). \\
\end{itemize}

\noindent \underline{Measures against unsuitable data}
\begin{itemize}
    \item A description of any methods implemented in data acquisition or processing, if any, to detect the presence of unsuitable data sources considering the model's intended uses, including but not limited to illegal content, child sexual abuse material (CSAM), non-consensual intimate imagery (NCII), and personal data leading to its unlawful processing. \\
\end{itemize}

\noindent \underline{Bias mitigation measures}
\begin{itemize}
    \item A description of any methods implemented in data acquisition or processing, if any, to address the prevalence of identifiable biases in the training data. \\
\end{itemize}

\noindent \textbf{C.6 Computational resources and energy consumption} \\
\emph{Section: ``Computational Resources'' and ``Energy Consumption'' of the Model Documentation Form} \\

\noindent \underline{Training time (compute)}
\begin{itemize}
    \item Measured or estimated amount of computation used for training, reported in computational operations and recorded with at least two significant figures (e.g.\ $2.4 \times 10^{25}$ floating point operations). \\
\end{itemize}

\noindent \underline{Energy consumption (training)}
\begin{itemize}
    \item Measured or estimated amount of energy used for training, reported in Megawatt-hours and recorded with at least two significant figures (e.g.\ $1.0 \times 10^{2}$~MWh). If the amount of energy used for training cannot be estimated due to the lack of critical information from a compute or hardware provider, enter ``N/A''.\\
\end{itemize}

\noindent \underline{Energy measurement methodology}
\begin{itemize}
    \item In the absence of a delegated act adopted in accordance with Article~53(5) AI Act to detail measurement and calculation methodologies, describe the methodology used to measure or estimate the amount of energy used for training. Where the energy consumption of the model is unknown, the energy consumption may be estimated based on information about computational resources used. If the amount of energy used for training cannot be estimated due to a lack of critical information from a compute or hardware provider, the provider should disclose the type of information they lack. \\
\end{itemize}

\noindent \textbf{C.7 Alignment with the EU Transparency Code of Practice} \\
\emph{MÖVE Team Addition} \\

\noindent \underline{Code of Practice signed}
\begin{itemize}
    \item Confirmation whether the AI Act Code of Practice ``Transparency'' has been signed by the model provider.
\end{itemize}